\documentclass[10pt,twocolumn]{article}
\usepackage{preprint}
\colortheme{green}

\usepackage{booktabs}
\usepackage{graphicx}
\usepackage{capt-of}

\usepackage{algorithm}
\usepackage{algpseudocode}
\usepackage{float}

\usepackage{titletoc}

\runningtitle{Mumcu and Yilmaz}
\setstatus{DORMS}

\ptitle{{DORMS: Image Segmentation via Depth-Ordered Region Merge Trees}}

\pauthor{Furkan Mumcu}{1}{furkan@usf.edu}
\pauthor{Yasin Yilmaz}{1}{}
\paffiliations{$^{1}$Department of Electrical Engineering, University of South Florida}

\pabstract{%
We introduce DORMS (Depth-Ordered Region Merging for Segmentation), a purely algorithmic approach to image segmentation from monocular depth maps, requiring no neural networks or learned parameters in the segmentation process. An off-the-shelf monocular depth estimator provides the input depth maps. Our method constructs depth-ordered merge trees through a deterministic union-find sweep, combining component area, depth persistence, and spatial contact constraints to guide region merging. A threshold search determines the segmentation granularity according to a specified target region count, while global depth detrending improves the separation of local geometric structures. Extensive experiments on COCO Panoptic, Cityscapes, and ADE20K demonstrate consistent improvements over classical segmentation baselines and competitive performance against zero-shot neural segmentation models. Notably, our method outperforms all evaluated zero-shot models on Cityscapes, while remaining competitive with several approaches on COCO and ADE20K. These results highlight the potential of purely algorithmic geometric reasoning for image segmentation without learned segmentation models.

\par\vspace{0.6em}

Project Page: \url{https://github.com/furkanmumcu/DORMS}

}

\begin{document}

\makepreprinttitle

\section{Introduction}
\label{sec:introduction}

Image segmentation is a fundamental problem in computer vision and has been extensively studied for decades. Early approaches relied on algorithmic techniques, including graph-based segmentation, clustering, and superpixels, to group pixels based on low-level image properties \cite{felzenszwalb2004efficient,achanta2012slic,arbelaez2011contour}. The development of deep learning shifted the field toward learned representations, enabling significant advances in semantic, instance, and panoptic segmentation \cite{long2015fully,he2017mask,kirillov2019panoptic}. More recently, transformer-based architectures and large foundation models have further advanced segmentation capabilities \cite{cheng2022masked,kirillov2023segment,ravi2024sam2}. Despite these advances, modern segmentation methods predominantly rely on learned visual representations and often require substantial training data or computational resources.

\begin{figure*}[t]
\centering
\begin{minipage}[c]{0.174\textwidth}
    \centering
    \includegraphics[width=\linewidth]{figures/introduction/input_rgb.png}\\[-1pt]
    {\scriptsize Input RGB}
\end{minipage}%
\hfill$\rightarrow$\hfill%
\begin{minipage}[c]{0.174\textwidth}
    \centering
    \includegraphics[width=\linewidth]{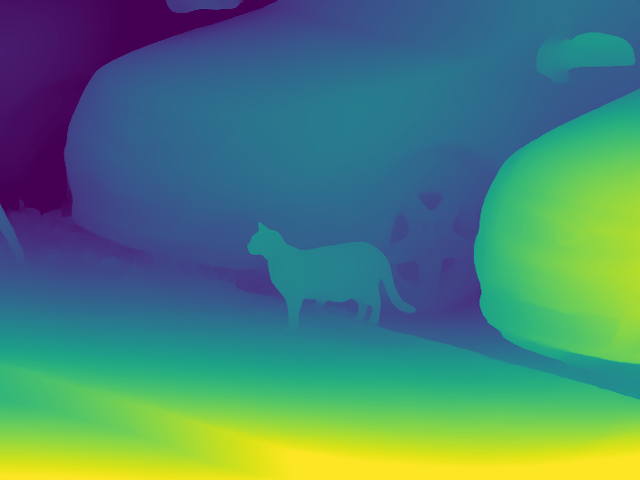}\\[-1pt]
    {\scriptsize Estimated depth}
\end{minipage}%
\hfill$\rightarrow$\hfill%
\begin{minipage}[c]{0.174\textwidth}
    \centering
    \includegraphics[width=\linewidth]{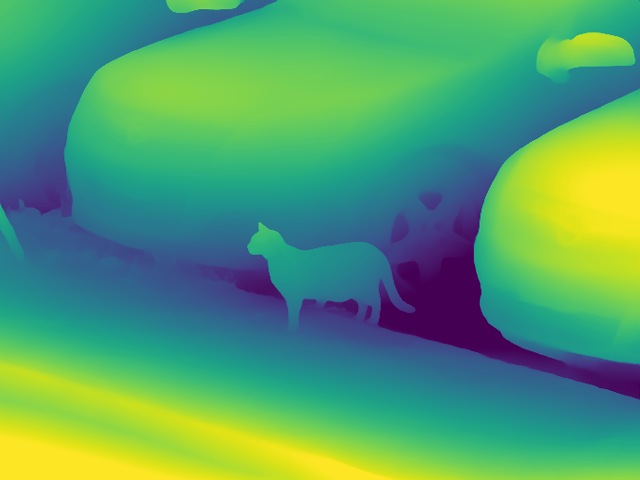}\\[-1pt]
    {\scriptsize Detrended depth}
\end{minipage}%
\hfill$\rightarrow$\hfill%
\begin{minipage}[c]{0.173\textwidth}
    \centering
    \fbox{\parbox[c][0.85in][c]{0.88\linewidth}{\centering
        \scriptsize\textbf{Depth-ordered}\\[-1pt]
        \scriptsize\textbf{region merging}\\[3pt]
        \tiny Area $\times$ persistence\\[1pt]
        \tiny Gated spatial contact\\[1pt]
        \tiny Repeated merge tests\\[1pt]
        \tiny Target count $k$
    }}\\[1pt]
    {\scriptsize Our algorithm}
\end{minipage}%
\hfill$\rightarrow$\hfill%
\begin{minipage}[c]{0.174\textwidth}
    \centering
    \includegraphics[width=\linewidth]{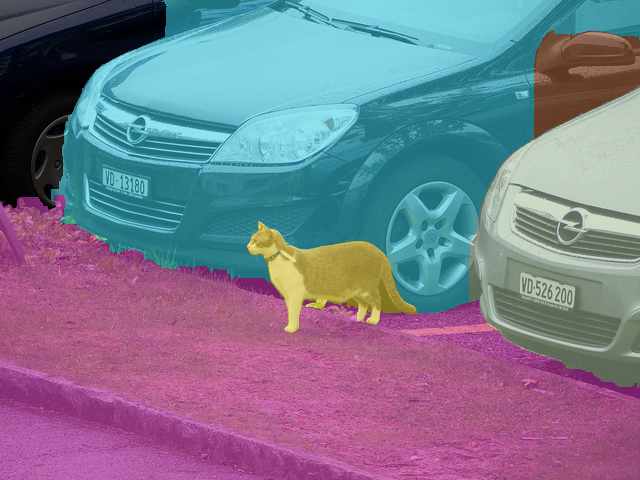}\\[-1pt]
    {\scriptsize Predicted segmentation}
\end{minipage}
\caption{Overview of the proposed depth-based segmentation pipeline on a real image. An off-the-shelf monocular depth estimator provides the input depth map, which is globally detrended before depth-ordered region merging. Our algorithm repeatedly evaluates merges using component area, depth persistence, and gated accumulated spatial contact, and selects a threshold according to a target region count. The final panel shows our actual class-agnostic segmentation prediction overlaid on the RGB image, not the ground truth.}
\label{fig:method_overview}
\end{figure*}

In this work, we investigate a complementary direction: can meaningful image segmentation be achieved directly from monocular depth maps without using neural networks for segmentation? Depth provides geometric information about scene structure, including relative distances, object boundaries, and spatial relationships \cite{dimiccoli2009hierarchical,yang2024depth}. Our motivation is to investigate how much of this structure can be recovered through algorithmic reasoning alone, without learning segmentation-specific representations or relying on semantic supervision. Beyond this investigation, we aim to lay the groundwork for lightweight and practical segmentation methods that can leverage depth maps obtained from different estimators or sensing systems. While the input depth can be obtained from a pretrained monocular depth estimator, the segmentation process itself uses no neural networks or learned parameters.

Building upon classical merge-tree and topological segmentation principles, we propose \textit{DORMS}, a novel algorithmic framework for image segmentation from monocular depth maps. DORMS integrates depth-ordered region merging, area--persistence-based scoring, gated spatial-contact constraints, and global depth detrending into a unified pipeline. Specifically, we first remove the dominant linear depth trend to emphasize local geometric variations. We then process pixels in depth order using a deterministic union--find procedure, evaluating candidate merges based on component area and depth persistence while using a gated neck-width constraint to prevent premature merging across narrow connections. A bounded threshold search selects a segmentation according to a specified target region count, followed by small-region cleanup to produce the final partition. An overview of DORMS is illustrated in Fig.~\ref{fig:method_overview}.

We evaluate our approach on COCO Panoptic, Cityscapes, and ADE20K \cite{lin2014microsoft,kirillov2019panoptic,cordts2016cityscapes,zhou2017scene}, comparing against classical segmentation algorithms and neural segmentation models under a unified class-agnostic evaluation protocol. Using Hungarian-matched mean Intersection-over-Union (mIoU), our method outperforms all evaluated classical baselines across the three datasets and all tested zero-shot segmentation models on Cityscapes. Stronger zero-shot methods achieve higher mIoU on COCO and ADE20K, while Segmentation Covering reveals additional differences in region quality and granularity. We further demonstrate that global depth detrending consistently improves segmentation performance across datasets, highlighting its importance for depth-based geometric reasoning.

Our main contributions are summarized as follows:
\begin{itemize}
    \item We introduce DORMS, an algorithmic image segmentation framework operating solely on depth maps, combining depth-ordered merge trees, area--persistence scoring, and gated spatial-contact constraints without learned segmentation parameters.

    \item We incorporate global depth detrending and demonstrate its effectiveness across diverse scenes, including those without a dominant planar surface, revealing the importance of local depth structure for geometric segmentation.

    \item We conduct an extensive class-agnostic evaluation across three segmentation benchmarks against classical and neural approaches, demonstrating both the capabilities and limitations of image segmentation based on depth geometry alone.
\end{itemize}

\section{Related Work}
\label{sec:related_work}

\textbf{Classical and Hierarchical Image Segmentation.}
Classical image segmentation methods group pixels according to low-level visual and geometric properties. Felzenszwalb and Huttenlocher~\cite{felzenszwalb2004efficient} proposed an efficient graph-based segmentation algorithm based on local evidence for region boundaries, while SLIC~\cite{achanta2012slic} introduced spatially coherent superpixels through iterative clustering. Hierarchical approaches further organize regions across multiple segmentation scales, including contour-based hierarchies~\cite{arbelaez2011contour} and merge-tree representations~\cite{liu2016image}. In particular, Liu et al.~\cite{liu2016image} constructed a hierarchy of superpixels using learned boundary-merging decisions. Our method similarly builds regions through hierarchical merging, but operates directly on depth-ordered pixels rather than an initial superpixel partition, and determines merges using geometric criteria without a learned classifier.

\textbf{Topological Segmentation and Persistence-Based Clustering.}
Component trees and contour trees provide established representations of the evolution of connected components in scalar fields~\cite{jones1997component,carr2003computing}. Persistent homology extends this perspective by characterizing the significance of components through their lifetimes across filtration levels. These principles have been applied to image segmentation through persistence-guided region merging~\cite{letscher2007image}, topological image analysis~\cite{assaf2018persistent}, and persistence-based clustering~\cite{chazal2013persistence}. Our approach builds upon these foundations by applying a depth-ordered component sweep and evaluating candidate merges through the product of component area and depth persistence. We further introduce a gated spatial-contact constraint that delays merging between sufficiently large components until adequate contact has accumulated. Together, these mechanisms define our depth-based region-merging procedure rather than relying on persistence alone.

\textbf{Depth-Based Geometric Segmentation.}
Depth information has long been used to support image segmentation through geometric relationships, occlusion cues, and surface structure. Dimiccoli and Salembier~\cite{dimiccoli2009hierarchical} incorporated depth ordering inferred from monocular image cues into hierarchical region representations. Other methods have explored joint color--depth clustering and region merging~\cite{hasnat2016joint}, as well as depth-image segmentation based on geometric surface properties~\cite{antic2021depth}. More recently, monocular depth estimation models such as Depth Anything V2~\cite{yang2024depth} have enabled the extraction of dense depth maps from single images. Our work investigates a complementary segmentation formulation that uses such depth maps as its sole input, without RGB features or semantic information during segmentation. Rather than targeting specific surface primitives, we combine global depth detrending with depth-ordered geometric merging to obtain general class-agnostic image partitions.

\textbf{Learning-Based and Foundation-Model Segmentation.}
Deep learning has substantially advanced semantic, instance, and panoptic segmentation through learned visual representations, including transformer-based architectures such as Mask2Former~\cite{cheng2022masked}. Foundation models such as SAM~\cite{kirillov2023segment} and SAM2~\cite{ravi2024sam2} further enable promptable segmentation using large-scale pretrained networks. Other approaches, including GroupViT~\cite{xu2022groupvit} and CLIPSeg~\cite{lueddecke2022image}, leverage vision--language representations, while DiffCut~\cite{couairon2024diffcut} performs algorithmic graph partitioning over pretrained diffusion features. Although these methods differ in their supervision and segmentation mechanisms, they rely on learned representations to produce image regions. In contrast, our segmentation algorithm uses only the supplied depth values and explicitly defined geometric operations. A pretrained model may provide the initial depth estimate, but no neural network or learned parameters are involved in the subsequent segmentation process.

\section{Proposed Method}
\label{sec:method}

We propose DORMS (Depth-Ordered Region Merging for Segmentation), an algorithmic framework that segments monocular depth maps through global detrending and depth-ordered merge trees. A deterministic pixel sweep evaluates region merges using component area, depth persistence, and spatial contact, while a bounded threshold search controls segmentation granularity, followed by conditional small-region cleanup. The segmentation process requires no learned parameters. Detailed pseudocode and an algorithmic breakdown are provided in the supplementary material, and the implementation code is publicly available in our project page.

\subsection{Problem Formulation and Depth Detrending}
\label{sec:depth_detrending}

Let $\Omega \subset \mathbb{Z}^2$ denote the image domain containing $N=|\Omega|$ pixels, and let $G=(\Omega,\mathcal{E})$ be its 4-connected adjacency graph. Given a depth map $d:\Omega\rightarrow\mathbb{R}$, our objective is to partition $\Omega$ into non-overlapping regions using only depth values. We adopt the convention that larger values correspond to points closer to the camera, consistent with the relative depth representation used by our monocular depth estimator.

Before constructing the segmentation, we remove the dominant linear spatial trend from the depth map. Specifically, we fit a plane to the depth values using ordinary least squares:
\begin{equation}
\label{eq:depth_plane}
(\hat a,\hat b,\hat c)
=
\arg\min_{a,b,c}
\sum_{(x,y)\in\Omega}
\left[d(x,y)-(ax+by+c)\right]^2.
\end{equation}
The detrended depth map is then obtained as
\begin{equation}
\label{eq:depth_detrending}
\tilde d(x,y)
=
\operatorname{Norm}_{[0,1]}
\left(d(x,y)-(\hat a x+\hat b y+\hat c)\right),
\end{equation}
where $\operatorname{Norm}_{[0,1]}$ denotes min--max normalization.

This operation removes the global spatial gradient rather than necessarily identifying a physical ground plane. Such gradients can influence the depth ordering and persistence of connected components, even when they do not correspond to distinct objects. Detrending reduces this influence and emphasizes local depth variations that can provide more informative cues for region separation. The resulting map $\tilde d$ is used throughout the subsequent segmentation process.

\subsection{Depth-Ordered Merge-Tree Construction}
\label{sec:merge_tree}

We construct a depth-ordered hierarchy of connected components through a deterministic pixel sweep. Let $\pi$ denote the permutation of pixels sorted from nearest to farthest according to $\tilde d$:
\begin{equation}
\label{eq:depth_order}
\tilde d(\pi(1)) \geq \tilde d(\pi(2))
\geq \cdots \geq \tilde d(\pi(N)).
\end{equation}
Ties are resolved using a fixed pixel index to ensure deterministic processing.

Pixels are activated sequentially in this order, while a disjoint-set (union--find) structure maintains the active connected components. When a pixel is activated, three cases are considered. If none of its 4-connected neighbors is active, the pixel establishes a new component. If its active neighbors belong to a single component, the pixel is absorbed into that component. Otherwise, the pixel connects multiple distinct components and generates candidate merge evaluations.

For each component $C$, we maintain its current area $A(C)$ and birth depth $b(C)$, corresponding to the depth value at which the component was created. Component areas are updated as pixels are added and other components are absorbed. When multiple components meet, the component with the earliest birth in the sweep is designated as the \emph{elder}, with ties resolved by the smallest root index. Each other component is treated as a \emph{junior}, producing a candidate merge evaluation with the elder. The newly activated pixel always joins the elder, regardless of whether the candidate merges are accepted.

\begin{figure*}[t]
\centering
\setlength{\tabcolsep}{0.8pt}
\begin{tabular}{@{}ccccccc@{}}
{\scriptsize RGB} & {\scriptsize 10\%} & {\scriptsize 25\%} & {\scriptsize 50\%} & {\scriptsize 75\%} & {\scriptsize 90\%} & {\scriptsize 100\%} \\
\includegraphics[width=0.136\textwidth]{figures/merge_sweep/context_rgb.png} &
\includegraphics[width=0.136\textwidth]{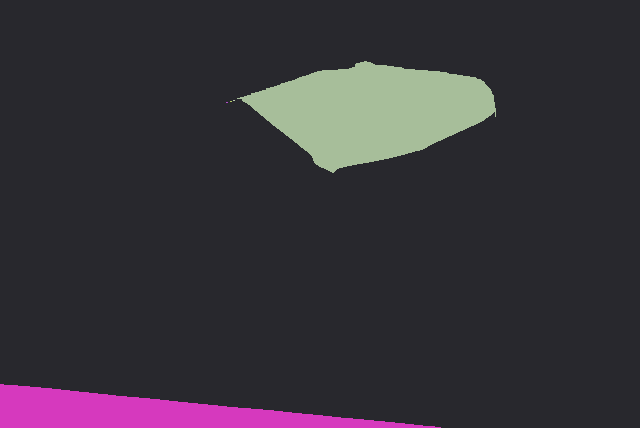} &
\includegraphics[width=0.136\textwidth]{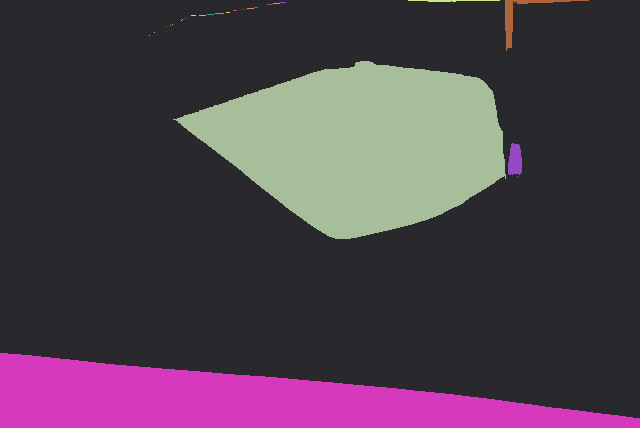} &
\includegraphics[width=0.136\textwidth]{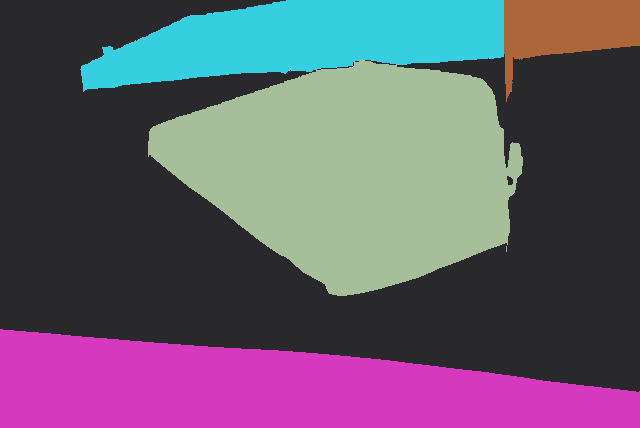} &
\includegraphics[width=0.136\textwidth]{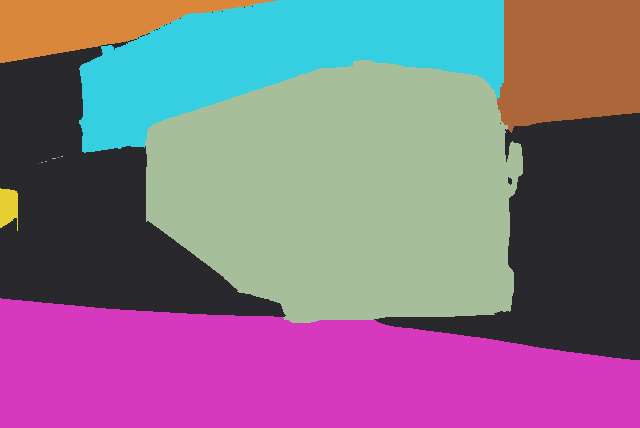} &
\includegraphics[width=0.136\textwidth]{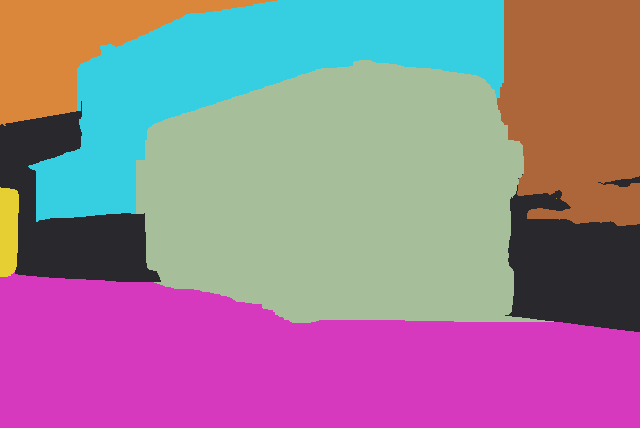} &
\includegraphics[width=0.136\textwidth]{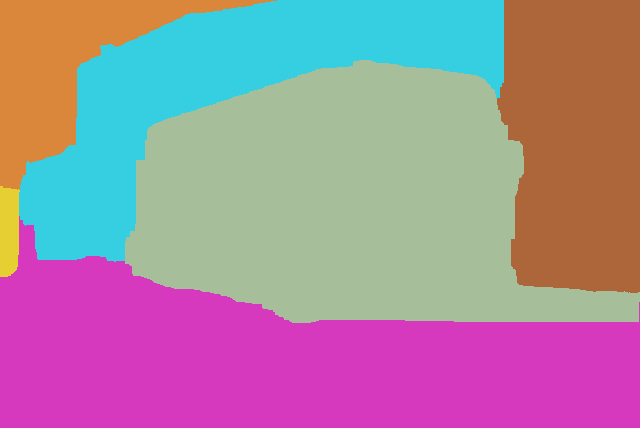}
\end{tabular}
\caption{Depth-ordered pixel activation and component evolution on a COCO image. The left panel shows the input RGB image; the remaining panels show the active regions after processing 10\%, 25\%, 50\%, 75\%, 90\%, and 100\% of pixels in descending detrended-depth order. Dark pixels have not yet been activated; colors distinguish components within each snapshot. Components can form and merge as the sweep progresses, so their number need not change monotonically.}
\label{fig:depth_sweep}
\end{figure*}

Candidate merges are evaluated during the sweep rather than automatically
accepted. Importantly, rejecting a merge does not necessarily prevent the
corresponding components from merging later: a pair rejected by the
spatial-contact constraint is re-evaluated at subsequent bridging pixels and may
be admitted once sufficient contact has accumulated, whereas a pair rejected by
the area--persistence criterion remains rejected for the remainder of the sweep
(Sec.~\ref{sec:merge_decisions}). Consequently, the component hierarchy depends
on the sequence of merge decisions, rather than being a fixed tree pruned only
after construction. Figure~\ref{fig:depth_sweep} illustrates the evolution of connected components during the depth-ordered sweep.

\subsection{Geometric Merge Decisions}
\label{sec:merge_decisions}

For a candidate evaluation $e$ between an elder component $C_e$ and a junior component $C_j$, occurring at the currently activated pixel $p$, we define two geometric quantities. The first is the junior component's current area $A(C_j)$. The second is its depth persistence,
\begin{equation}
\label{eq:persistence}
P(e)=b(C_j)-\tilde d(p),
\end{equation}
which measures the depth range over which the junior component has existed independently.

We combine these quantities into an area--persistence score:
\begin{equation}
\label{eq:merge_score}
\varphi(e)=A(C_j)\,P(e).
\end{equation}
The area term measures the spatial extent of the component, while persistence measures its separation across depth levels. Their product favors preserving components that are both spatially substantial and persistent in depth. Given a threshold $\tau$, a candidate merge is permitted by this criterion only when $\varphi(e)<\tau$. Thus, higher scores indicate stronger evidence for keeping two components separate.

\textbf{Gated spatial-contact constraint.}
Area and persistence alone do not characterize how two components are spatially connected. In particular, distinct regions may touch through a narrow connection and be merged despite having otherwise meaningful boundaries. To address this, we introduce an additional constraint based on accumulated contact between components.

When the neck-width mechanism is enabled, we maintain a contact counter $c(C_e,C_j)$ for each encountered component pair. The counter increases at every bridging-pixel encounter, regardless of component size or whether a previous merge was rejected. Unlike a geometric width measured at the initial contact, which is necessarily one pixel in the sweep, this counter captures contact accumulated as processing continues.

The constraint is applied only when both components are sufficiently large:
\begin{equation}
\label{eq:area_gate}
A(C_e)\geq gN,
\qquad
A(C_j)\geq gN,
\end{equation}
where $g$ is an area-gate fraction. 
Both areas are current values and include any components that either the elder
or the junior has previously absorbed.
When both conditions hold, merging is prevented until the accumulated contact reaches a minimum threshold $m$. Otherwise, the candidate proceeds directly to the area--persistence criterion. The area gate therefore controls whether the contact constraint is enforced, not whether contact is accumulated.

The complete merge decision is
\begin{equation}
\label{eq:merge_decision}
\operatorname{Admit}(e)=
\begin{cases}
0, &
\begin{aligned}
&m>1,\; A(C_e)\geq gN,\\[-2pt]
&A(C_j)\geq gN,\; c(C_e,C_j)<m,
\end{aligned}\\[4pt]
\mathbb{I}\!\left[\varphi(e)<\tau\right],
& \text{otherwise},
\end{cases}
\end{equation}
where $\mathbb{I}[\cdot]$ denotes the indicator function and the contact counter is updated before evaluating the constraint. The spatial-contact constraint is checked before computing the area--persistence score, and setting $m=1$ disables it entirely. Rejected pairs remain eligible for evaluation at subsequent bridging pixels.

\textbf{Repeated Evaluations and Merge Decisions.}
A component pair may be encountered multiple times during the depth-ordered sweep, but the outcome of a rejected merge depends on which constraint prevents it. A rejection by the spatial-contact constraint is temporary: as contact accumulates, the pair may eventually satisfy the contact threshold and proceed to the area--persistence criterion. In contrast, rejection by the area--persistence criterion is permanent. The score $\varphi(e)=A(C_j)\,P(e)$ is monotonically non-decreasing because the junior component's area $A(C_j)$ cannot decrease, while its persistence $P(e)=b(C_j)-\tilde d(p)$ cannot decrease as pixels are processed in descending depth order. Moreover, the score is independent of the elder component's identity. Therefore, once $\varphi(e)\geq\tau$, the junior component cannot satisfy the merge criterion at any subsequent encounter, even with a different elder. Because the score cannot decrease while a pair is deferred, delaying the test can only make the area--persistence criterion harder to satisfy, never easier. Repeated evaluations thus allow a contact-deferred pair to reach the area--persistence criterion, but cannot reverse a rejection by it.

\subsection{Target Region Count and Threshold Search}
\label{sec:threshold_search}

The merge threshold $\tau$ determines which candidate merges are accepted and therefore influences the number of regions in the resulting partition. Rather than choosing a fixed threshold across all images, we specify a target region count $k$ and search for a threshold that produces an appropriate partition.

Let $\mathcal{S}_{\tau}$ denote the partition obtained by repeating the depth-ordered sweep with threshold $\tau$, and let
\begin{equation}
\label{eq:region_count}
R(\tau)=|\mathcal{S}_{\tau}|
\end{equation}
be its number of regions before cleanup.

The function $R(\tau)$ is not guaranteed to be monotonic. Accepting or rejecting an earlier merge can change the components encountered later in the sweep, potentially altering subsequent merge decisions. Consequently, conventional binary search based on monotonicity is not generally applicable.

We instead use bounded adaptive subdivision of the threshold range, evaluating candidate thresholds by replaying the sweep and collecting the region counts they produce. The search interval is defined by
\begin{equation}
\label{eq:threshold_bound}
\tau_{\min}=10^{-6},
\qquad
\tau_{\max}=\max(10^6,20s_{\max}),
\end{equation}
where $s_{\max}$ denotes the maximum merge score used to determine the image-specific upper bound.

The search is bounded by a maximum number of sweep replays, a minimum bracket resolution, and a region-count cap. During adaptive subdivision, a bracket is pruned when both endpoint region counts exceed the cap, preventing unnecessary exploration of counts far above the target. The reported evaluations use a maximum replay budget of 2,000 and a region-count cap of $\max(24,3k_{\max})$, where $k_{\max}$ is the largest target in the evaluated target-count grid. The search therefore examines a bounded subset of thresholds rather than exhaustively enumerating all possible outcomes.

Let $\mathcal{R}$ denote the set of region counts discovered during the search. The selected count is
\begin{equation}
\label{eq:region_selection}
R^{*}=
\begin{cases}
\min\{R\in\mathcal{R}:R\geq k\},
& \exists R\in\mathcal{R}:R\geq k,\\[3pt]
\min\displaystyle\arg\min_{R\in\mathcal{R}}|R-k|,
& \text{otherwise}.
\end{cases}
\end{equation}
Thus, the search prefers the smallest discovered region count that meets or exceeds $k$. If no such count is found, it selects the closest discovered count, resolving equal-distance ties in favor of the smaller count. Each discovered count is associated with a witness threshold $\tau$ that produced it, and the final partition is obtained by replaying the sweep at the selected witness threshold $\tau^*$, rather than an arbitrary bracket midpoint. This is important because $R(\tau)$ is non-monotonic, and a bracket midpoint need not reproduce the selected region count. Since achievable counts may contain gaps and the search is bounded, the selected count need not equal the target exactly, and global optimality over all thresholds is not guaranteed.

\subsection{Small-Region Cleanup and Final Segmentation}
\label{sec:region_cleanup}

The neck-width constraint can leave small residual components whose boundaries are too short to accumulate the required contact for merging. To address this, we apply a conditional cleanup step to remove such regions.

We define a minimum region area threshold
\begin{equation}
\label{eq:cleanup_threshold}
A_{\min}=\max(A_{\mathrm{min,abs}},\rho N),
\end{equation}
where $A_{\mathrm{min,abs}}$ is an absolute minimum region size and $\rho$ is a minimum area fraction. 
In all reported experiments $\rho N$ is the binding term; $A_{\mathrm{min,abs}}$
acts only as a floor for degenerate inputs, guaranteeing a positive threshold
when $\rho N < A_{\mathrm{min,abs}}$.
Cleanup is performed only when the neck-width constraint is enabled ($m>1$) and the integer-valued area threshold exceeds one pixel. In particular, no cleanup is applied when $m=1$.

When cleanup is enabled, each pixel belonging to a region smaller than the
threshold is reassigned the label of its nearest surviving pixel, located by a
Euclidean distance transform. Reassignment is therefore performed per pixel
rather than per region, so a removed region may be absorbed by more than one
neighbouring region. Region labels are then renumbered consecutively from zero,
which is what makes the reported region count well defined. This produces the
final segmentation $\mathcal{S}_{\mathrm{final}}$.

The number of regions reported in our experiments is measured after the applicable cleanup step. Consequently, the final region count can differ from the count selected by the threshold search, including cases where cleanup reduces it below the specified target $k$.

The complete pipeline is therefore
\begin{equation}
\label{eq:full_pipeline}
d
\xrightarrow{\mathrm{Detrend}}
\tilde d
\xrightarrow{\mathrm{Merge\;and\;Select}}
\mathcal{S}_{\tau^{*}}
\xrightarrow{\mathrm{Conditional\;Cleanup}}
\mathcal{S}_{\mathrm{final}}.
\end{equation}
All stages operate directly on depth values and geometric relationships, without neural networks or learned parameters in the segmentation algorithm.

\section{Experimental Setup}
\label{sec:experimental_setup}

\subsection{Datasets and Depth Inputs}
\label{sec:datasets}

We evaluate DORMS on three image segmentation benchmarks: COCO Panoptic~\cite{lin2014microsoft,kirillov2019panoptic}, Cityscapes~\cite{cordts2016cityscapes}, and ADE20K~\cite{zhou2017scene}. We use the COCO val2017 split containing 5,000 images, the Cityscapes fine validation split containing 500 images, and the ADE20K validation split containing 2,000 images. These datasets cover diverse environments, from everyday objects and indoor scenes to complex urban environments.

For DORMS, depth maps are obtained using pretrained Depth Anything V2~\cite{yang2024depth}, while all subsequent segmentation operations rely solely on depth values without RGB features or semantic labels. We also evaluate Cityscapes stereo disparity and conduct ablations with different depth-estimator backbones. For ADE20K, panoptic annotations are constructed from semantic and instance masks. Further details on annotation construction, depth preprocessing, and backbone ablations are provided in the supplementary material.

\subsection{Comparison Methods}
\label{sec:baselines}

We compare against classical segmentation algorithms and neural segmentation models under a unified class-agnostic evaluation protocol. Classical baselines include Felzenszwalb's graph-based segmentation~\cite{felzenszwalb2004efficient} and SLIC~\cite{achanta2012slic}, evaluated using RGB and depth inputs where applicable. Neural comparisons include the zero-shot segmentation methods SAM~\cite{kirillov2023segment}, SAM2~\cite{ravi2024sam2}, GroupViT~\cite{xu2022groupvit}, CLIPSeg~\cite{lueddecke2022image}, and DiffCut~\cite{couairon2024diffcut}. We additionally include COCO-supervised Mask2Former~\cite{cheng2022masked} as a fully supervised reference on COCO.

For vocabulary-conditioned methods, GroupViT and CLIPSeg receive the corresponding dataset's category vocabulary as text input. This provides dataset-specific semantic information unavailable to the other zero-shot and classical baselines, which should be considered when interpreting their results. Overlapping SAM and SAM2 proposals are converted into complete, non-overlapping partitions, with uncovered pixels retained as an additional region. All methods are evaluated using the same ground-truth annotations and scoring implementation rather than relying on metrics reported in their original publications. Additional baseline configurations and partition-conversion details are provided in the supplementary material.

\begin{table*}[t]
\centering
\makebox[\textwidth][c]{%
\begin{minipage}[t]{0.755\textwidth}
\centering

\captionof{table}{Class-agnostic segmentation comparison on COCO Panoptic, Cityscapes, and ADE20K. We report Hungarian-matched mIoU (H-mIoU), Segmentation Covering (Cov.), and the average number of predicted regions (\#R). All methods are evaluated using the same scoring protocol. Bold indicates the highest H-mIoU on each dataset. Mask2Former is a supervised reference; other neural baselines are evaluated in a zero-shot setting.}
\label{tab:segmentation_comparison}

\setlength{\tabcolsep}{2.5pt}
\scriptsize
\resizebox{\linewidth}{!}{%
\begin{tabular}{lccccccccc}
\toprule
& \multicolumn{3}{c}{COCO Panoptic}
& \multicolumn{3}{c}{Cityscapes}
& \multicolumn{3}{c}{ADE20K} \\
\cmidrule(lr){2-4}
\cmidrule(lr){5-7}
\cmidrule(lr){8-10}

Method
& \#R & H-mIoU & Cov.
& \#R & H-mIoU & Cov.
& \#R & H-mIoU & Cov. \\
\midrule

\multicolumn{10}{l}{\textit{Classical methods}} \\

Felzenszwalb (RGB)
& 14.04 & 0.1349 & 0.4638
& 35.21 & 0.0917 & 0.4562
& 14.11 & 0.1413 & 0.4425 \\

Felzenszwalb (depth)
& 11.10 & 0.1516 & 0.4901
& 70.16 & 0.0522 & 0.4598
& 10.55 & 0.1336 & 0.4348 \\

SLIC (RGB)
& 3.49 & 0.1537 & 0.3931
& 20.09 & 0.0732 & 0.2350
& 6.77 & 0.1528 & 0.3684 \\

SLIC (depth/disparity)
& 5.16 & 0.1712 & 0.3823
& 25.91 & 0.0735 & 0.1871
& 4.75 & 0.1366 & 0.3460 \\

\midrule
\multicolumn{10}{l}{\textit{Zero-shot neural methods}} \\

SAM (16 points)
& 27.17 & 0.1608 & 0.5765
& 26.53 & 0.1234 & 0.4536
& 26.95 & 0.1946 & 0.5558 \\

SAM (32 points)
& 47.41 & 0.1168 & 0.5813
& 52.98 & 0.1050 & 0.4766
& 47.43 & 0.1472 & 0.5635 \\

SAM2 (16 points)
& 10.11 & 0.2783 & 0.6002
& 22.77 & 0.1365 & 0.4063
& 8.66 & 0.2615 & 0.5357 \\

SAM2 (32 points)
& 14.59 & 0.2648 & 0.6107
& 41.35 & 0.1464 & 0.4356
& 11.92 & 0.2696 & 0.5474 \\

GroupViT
& 4.02 & 0.1429 & 0.4099
& 2.21 & 0.0303 & 0.3032
& 3.99 & 0.1221 & 0.3829 \\

CLIPSeg
& 13.31 & 0.2471 & 0.6684
& 9.24 & 0.1506 & 0.6713
& 17.25 & 0.2126 & 0.6488 \\

DiffCut
& 6.31 & \textbf{0.3245} & 0.6668
& 5.54 & 0.1309 & 0.6834
& 6.20 & \textbf{0.2841} & 0.6329 \\

\midrule
\multicolumn{10}{l}{\textit{Fully supervised reference}} \\

Mask2Former
& 9.93 & {0.5590} & 0.8163
& 20.20 & 0.4278 & 0.8788
& 12.09 & 0.5565 & 0.8310 \\

\midrule
\textbf{DORMS (depth)}
& 5.93 & 0.2332 & 0.5119
& 25.84 & \textbf{0.2086} & 0.5866
& 7.89 & 0.2206 & 0.5126 \\

\bottomrule
\end{tabular}%
}
\end{minipage}%
\hspace{0.02\textwidth}%
%
\begin{minipage}[t]{0.225\textwidth}
\centering

\captionof{table}{Peak memory (MB) on COCO. CPU denotes process RSS and GPU denotes process VRAM. Our segmentation stage (seg.) and complete CPU pipeline (+DA-V2) are reported separately. See supplementary material for details.}
\label{tab:memory_comparison}

\setlength{\tabcolsep}{2pt}
\scriptsize
\resizebox{\linewidth}{!}{%
\begin{tabular}{@{}lrr@{}}
\toprule
Method & CPU & GPU \\
\midrule

\textbf{DORMS (seg.)} & 231 & -- \\
DORMS (+DA-V2) & 1128 & -- \\

\midrule
Felzenszwalb & 231 & -- \\
SLIC & 157 & -- \\

\midrule
GroupViT & 1116 & 786 \\
Mask2Former & 1344 & 972 \\
SAM (16) & 3434 & 4342 \\
SAM (32) & 3430 & 4342 \\
SAM2 (16) & 4472 & 6246 \\
SAM2 (32) & 4558 & 6246 \\
CLIPSeg & 6434 & 6396 \\
DiffCut & 8636 & 17360 \\

\bottomrule
\end{tabular}%
}
\end{minipage}%
}
\end{table*}

\subsection{Evaluation Metrics}
\label{sec:metrics}

We evaluate segmentation quality using class-agnostic Hungarian-matched mean Intersection-over-Union (mIoU) and Segmentation Covering. Unlike conventional semantic mIoU, our evaluation does not require predicted category labels.

For Hungarian-matched mIoU, predicted regions are matched one-to-one with ground-truth panoptic segments by maximizing their total IoU. For an image with $M$ predicted regions and $T$ ground-truth segments, we compute
\begin{equation}
\label{eq:hungarian_miou}
\mathrm{mIoU}_{\mathrm{H}} =
\frac{1}{\max(M,T)}
\sum_{(i,j)\in\mathcal{M}}
\operatorname{IoU}(P_i,G_j),
\end{equation}
where $\mathcal{M}$ denotes the optimal one-to-one matching. Unmatched regions contribute zero, penalizing both over-segmentation and under-segmentation.

Segmentation Covering measures the best overlap between each ground-truth segment and the predicted partition:
\begin{equation}
\label{eq:covering}
\mathrm{Covering} =
\frac{1}{\sum_{j=1}^{T}|G_j|}
\sum_{j=1}^{T}
|G_j|\max_i \operatorname{IoU}(G_j,P_i).
\end{equation}
Unlike Hungarian-matched mIoU, Covering does not enforce one-to-one matching and therefore responds differently to region granularity. Both metrics exclude void or unlabeled ground-truth pixels. We report averages across images, together with the mean number of predicted regions after cleanup.

We do not report instance Average Precision (AP), as our evaluation concerns class-agnostic, full-scene partitions rather than confidence-ranked object-instance detections.
Dataset-specific configurations, hyperparameter selection procedures, depth model checkpoints, and implementation details are provided in the supplementary material.

\section{Experimental Results and Analysis}
\label{sec:results}

\subsection{Comparison with Existing Segmentation Methods}
\label{sec:comparison}

Table~\ref{tab:segmentation_comparison} compares DORMS with classical and neural segmentation approaches on COCO Panoptic, Cityscapes, and ADE20K. DORMS achieves H-mIoU scores of 0.2332, 0.2086, and 0.2206, respectively, outperforming all evaluated classical baselines across the three datasets. It also exceeds their H-mIoU-selected Covering scores, although COCO Felzenszwalb achieves marginally higher Covering when optimized separately (0.5124 vs.\ 0.5119), using substantially more regions (29.22 vs.\ 5.93). On Cityscapes, DORMS achieves the highest H-mIoU, exceeding the strongest zero-shot baseline, CLIPSeg (0.1506). On COCO and ADE20K, stronger neural baselines achieve higher H-mIoU, while the COCO-supervised Mask2Former provides a substantially stronger supervised reference.

The metrics capture complementary properties: H-mIoU uses one-to-one matching, whereas Covering measures the best overlap for each ground-truth region. Consequently, DiffCut and CLIPSeg achieve higher Covering on Cityscapes despite lower H-mIoU. Table~\ref{tab:memory_comparison} further shows that our parameter-free segmentation stage requires only 231\,MB of CPU RAM on COCO, increasing to 1,128\,MB for the complete pipeline including depth estimation with Depth Anything V2 (DA-V2). Detailed baseline configurations, computational measurements, and geometric ablations are provided in the supplementary material.

\subsection{Contribution of Global Depth Detrending}
\label{sec:detrending_results}

We evaluate global depth detrending by comparing segmentation performance with and without this preprocessing step. As shown in Table~\ref{tab:detrending_ablation}, detrending improves H-mIoU from 0.2037 to 0.2332 on COCO Panoptic (+0.0295) and from 0.1753 to 0.2206 on ADE20K (+0.0453). On Cityscapes stereo disparity, H-mIoU increases from 0.0612 to 0.1045 at matched segmentation parameters, although differing threshold-search settings prevent attributing the entire gain to detrending alone (Sec.~\ref{sec:sensor_depth_ablation}).

\begin{figure}[t]
\centering
\includegraphics[width=0.7\columnwidth]{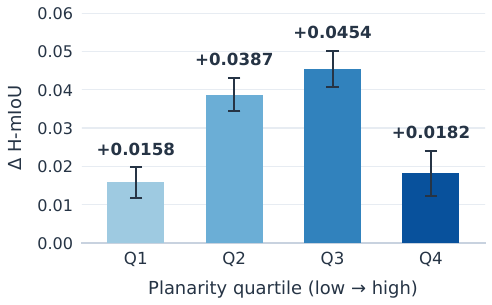}
\caption{Effect of global depth detrending across COCO planarity quartiles, ordered by increasing plane-fit $R^2$ (1,250 images per quartile). 
}
\label{fig:planarity_quartiles}
\end{figure}

Figure~\ref{fig:planarity_quartiles} shows statistically significant improvements across all four COCO planarity quartiles, with the largest gains in intermediate quartiles rather than the most planar scenes. This suggests that detrending benefits geometric merging beyond simply removing dominant planes. However, the effect is method-dependent: on COCO, detrending marginally improves Felzenszwalb (+0.0020) while reducing SLIC performance (-0.0086). These results indicate that its benefits are closely linked to the proposed depth-ordered merging mechanism. Additional analysis is provided in the supplementary material.

\begin{table}[t]
\centering
\caption{Effect of global depth detrending on Hungarian-matched mIoU. COCO and ADE20K results use their full validation sets. 
}
\label{tab:detrending_ablation}
\small
\setlength{\tabcolsep}{4pt}
\begin{tabular}{lccc}
\toprule
Dataset / Method & Without & With & $\Delta$ \\
\midrule
COCO / DORMS & 0.2037 & 0.2332 & +0.0295 \\
Cityscapes / DORMS (stereo) & 0.0612 & 0.1045 & +0.0433 \\
ADE20K / DORMS & 0.1753 & 0.2206 & +0.0453 \\
\midrule
COCO / Felzenszwalb & 0.1496 & 0.1516 & +0.0020 \\
COCO / SLIC & 0.1712 & 0.1626 & -0.0086 \\
\bottomrule
\end{tabular}
\end{table}

\begin{figure*}[t]
\centering
\setlength{\tabcolsep}{1pt}
\begin{tabular}{@{}ccccc@{}}
{\scriptsize RGB} & {\scriptsize Raw depth} & {\scriptsize Detrended depth} & {\scriptsize Ours (overlay)} & {\scriptsize GT (overlay)} \\
\includegraphics[width=0.19\textwidth]{figures/qualitative/coco/rgb.png} &
\includegraphics[width=0.19\textwidth]{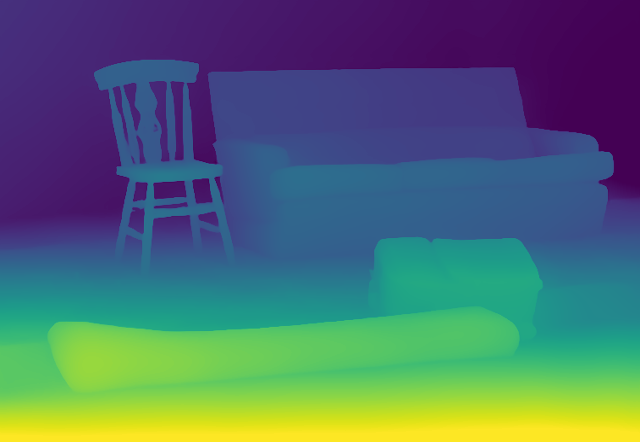} &
\includegraphics[width=0.19\textwidth]{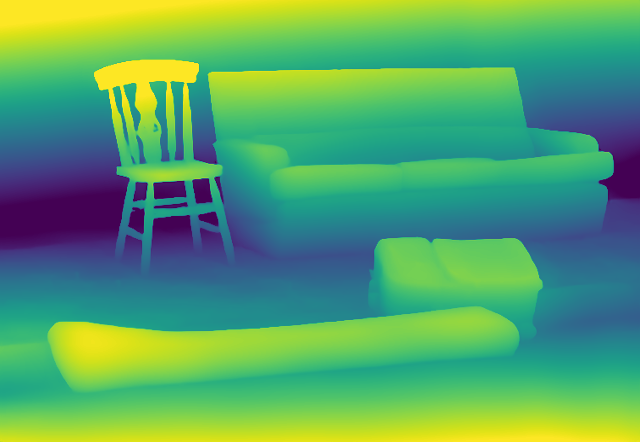} &
\includegraphics[width=0.19\textwidth]{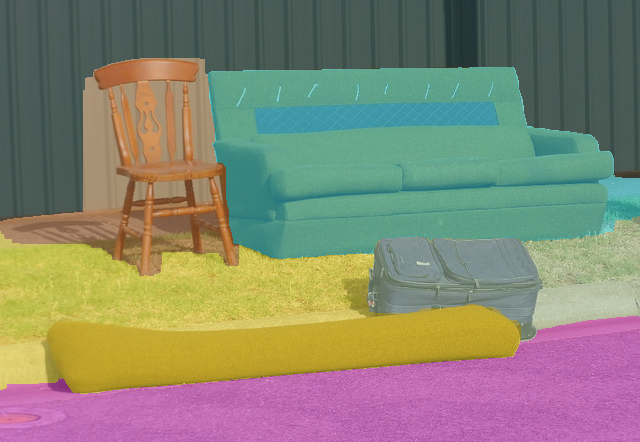} &
\includegraphics[width=0.19\textwidth]{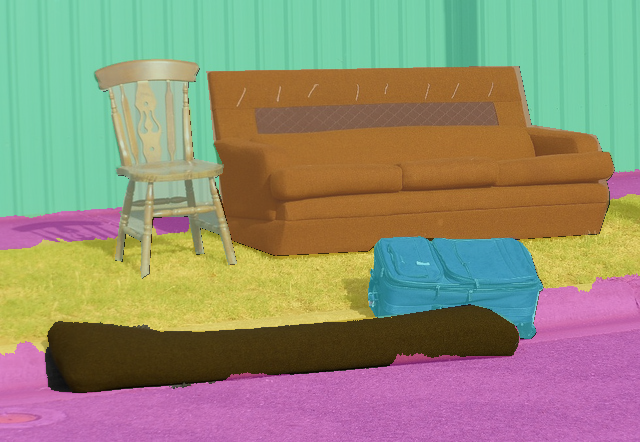} \\
\includegraphics[width=0.19\textwidth]{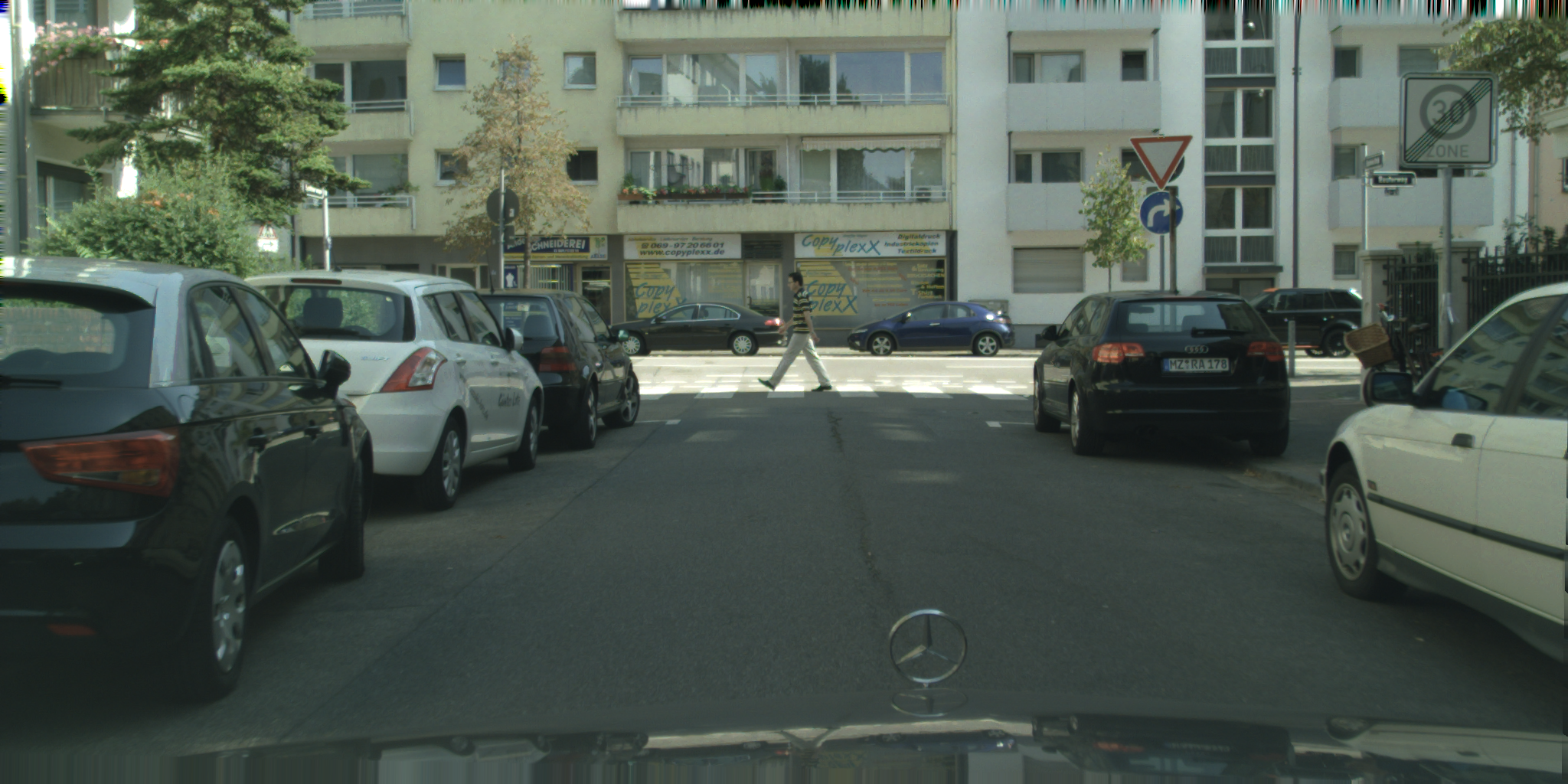} &
\includegraphics[width=0.19\textwidth]{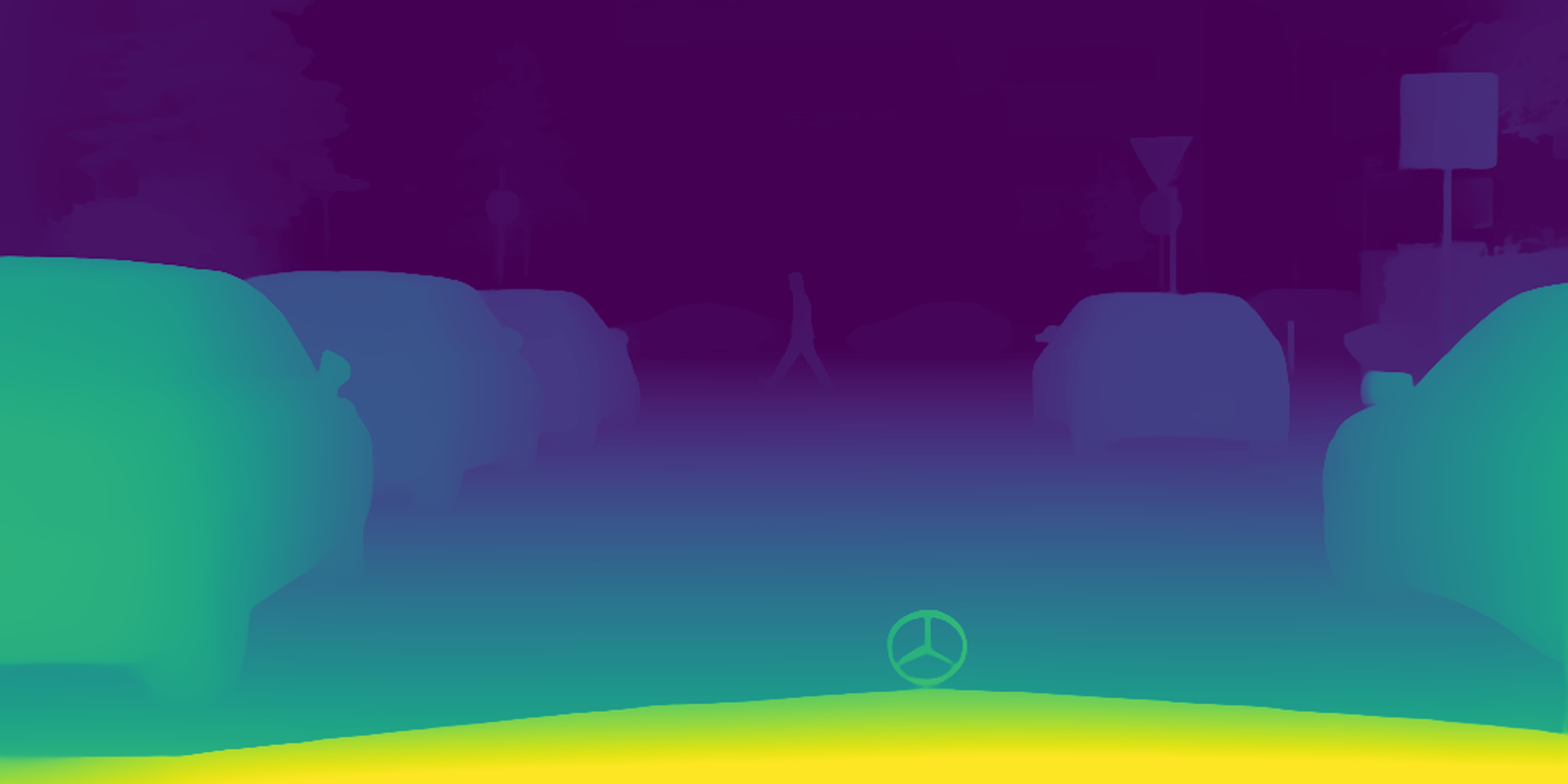} &
\includegraphics[width=0.19\textwidth]{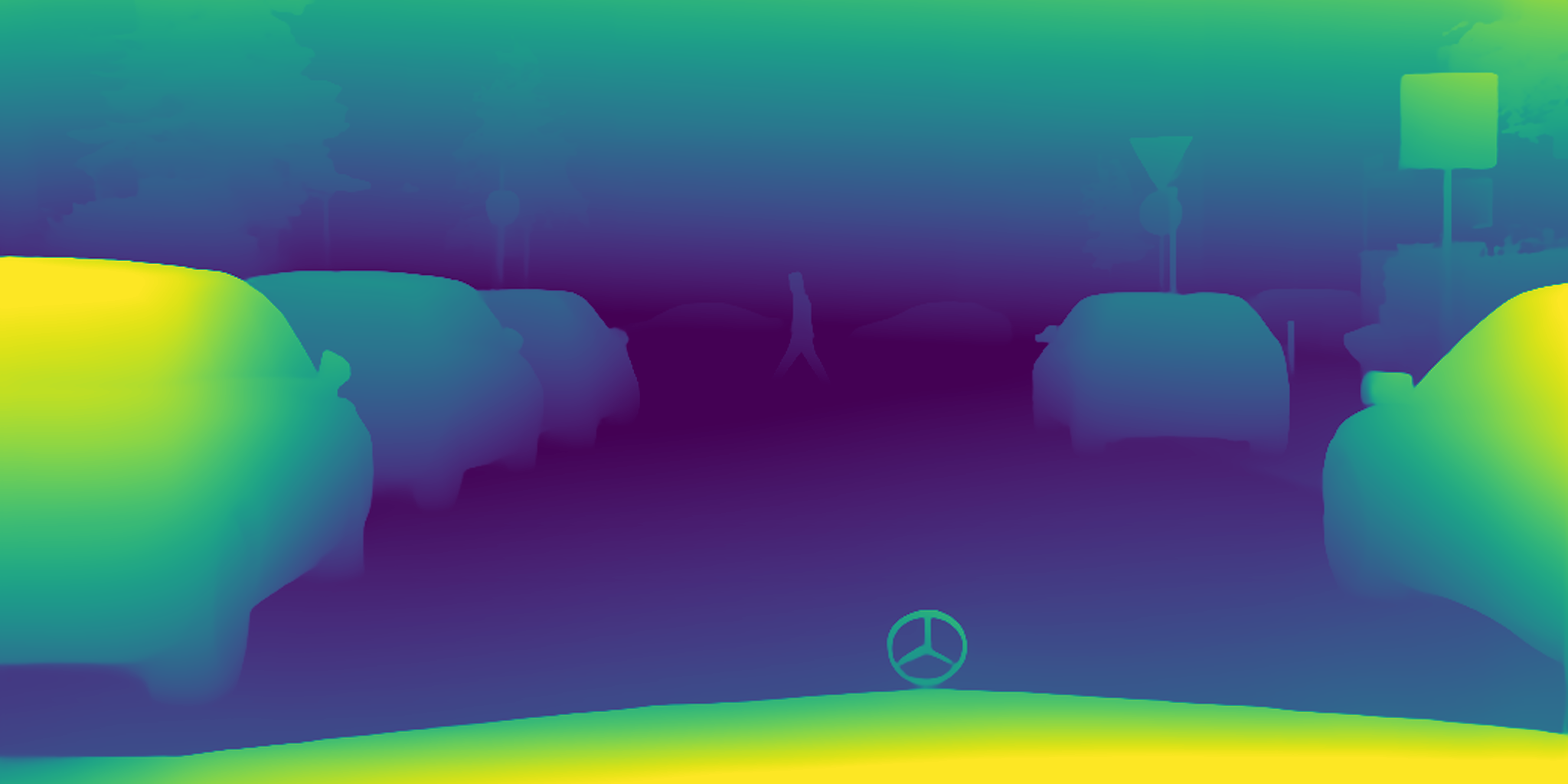} &
\includegraphics[width=0.19\textwidth]{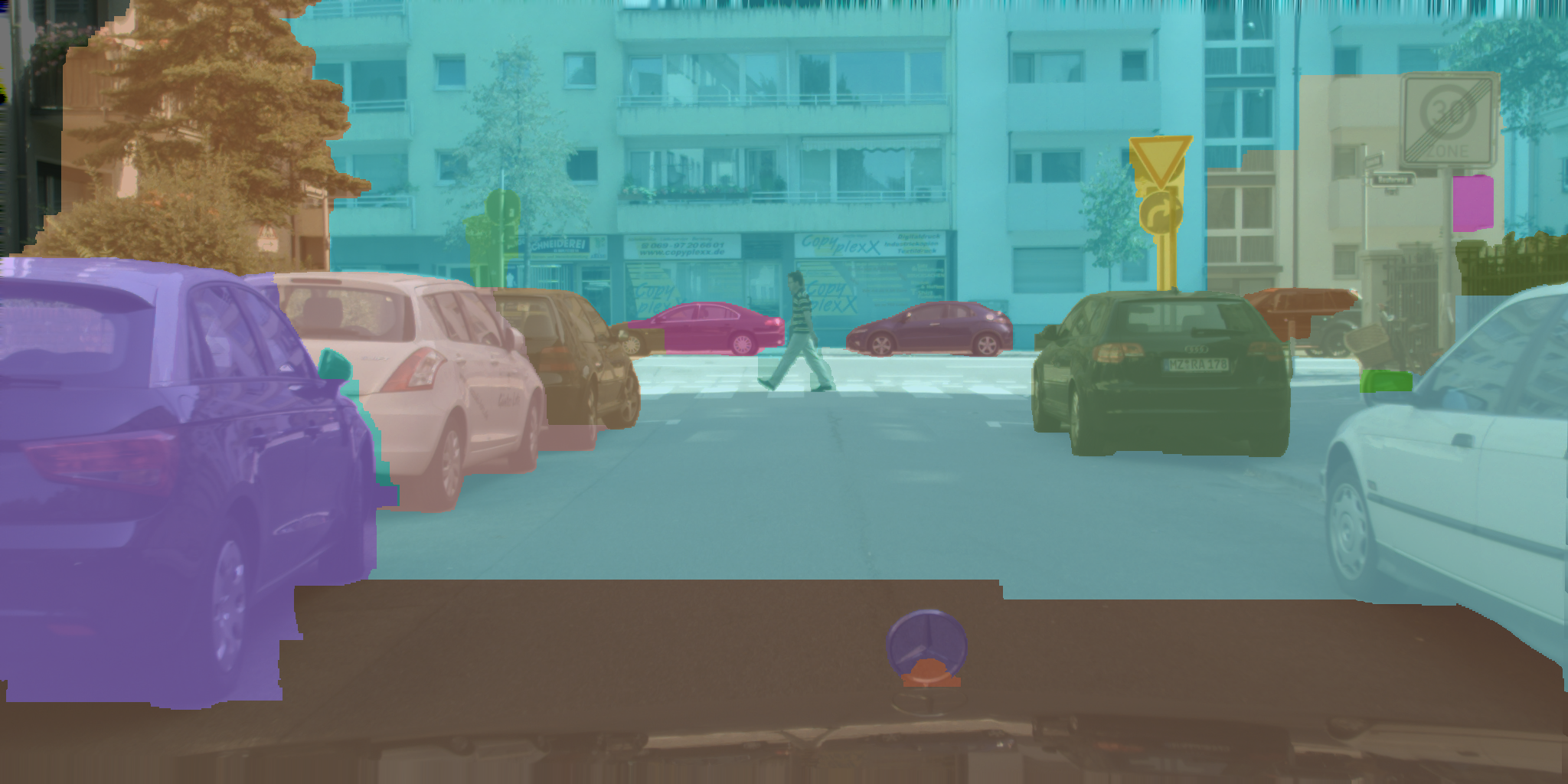} &
\includegraphics[width=0.19\textwidth]{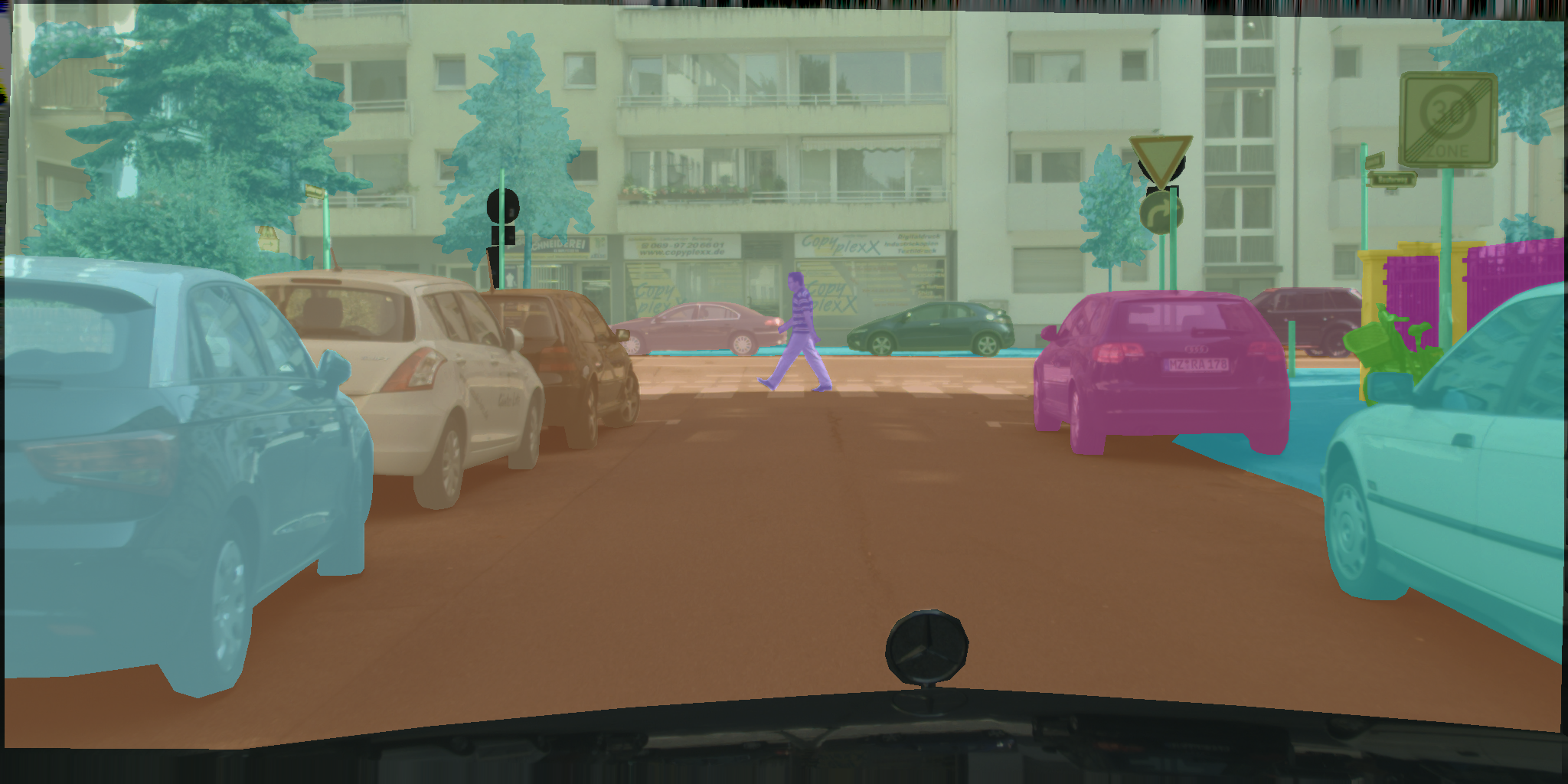} \\
\includegraphics[width=0.19\textwidth]{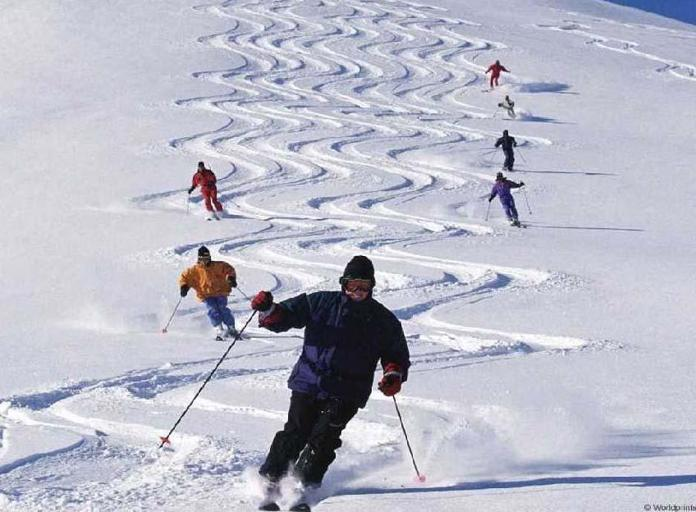} &
\includegraphics[width=0.19\textwidth]{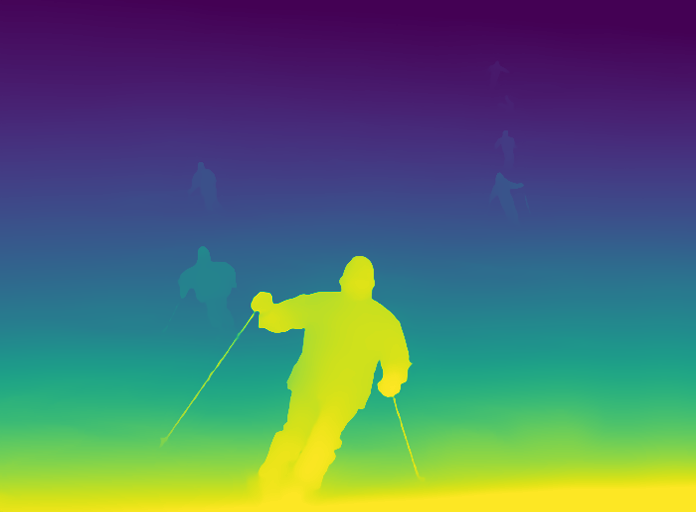} &
\includegraphics[width=0.19\textwidth]{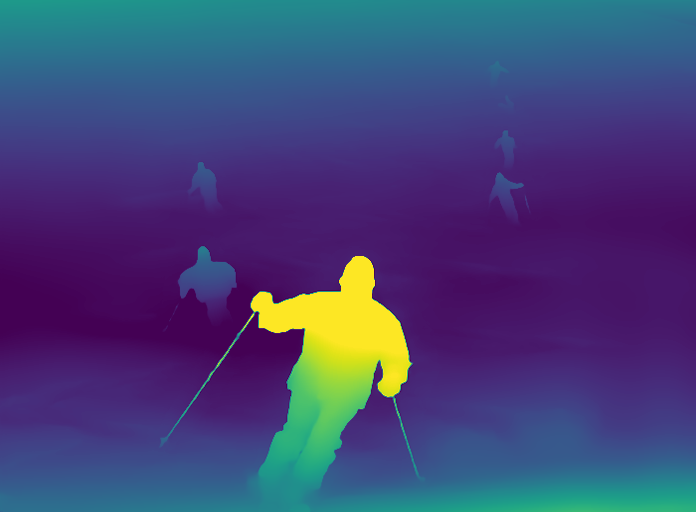} &
\includegraphics[width=0.19\textwidth]{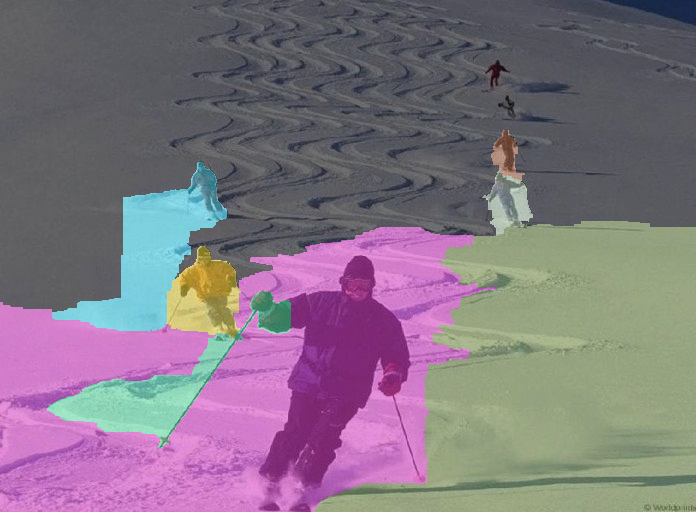} &
\includegraphics[width=0.19\textwidth]{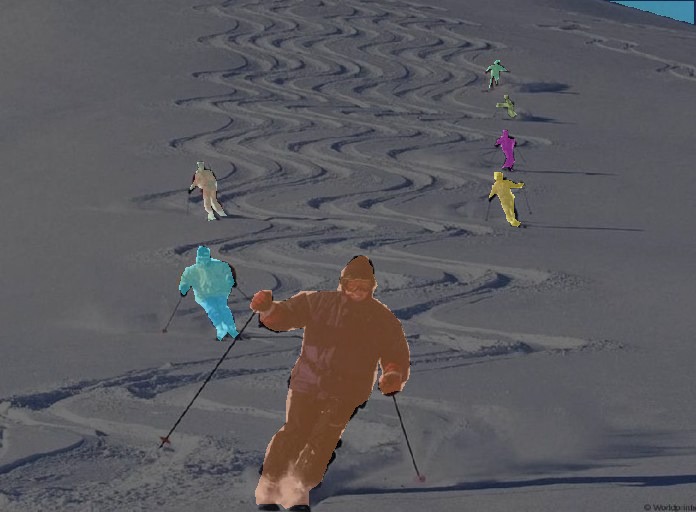} \\
\includegraphics[width=0.19\textwidth]{figures/qualitative/ade20k_failure/rgb.png} &
\includegraphics[width=0.19\textwidth]{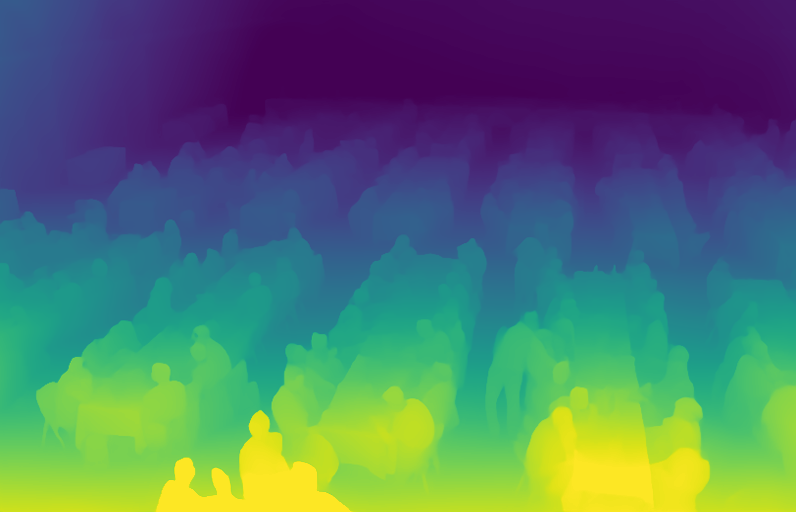} &
\includegraphics[width=0.19\textwidth]{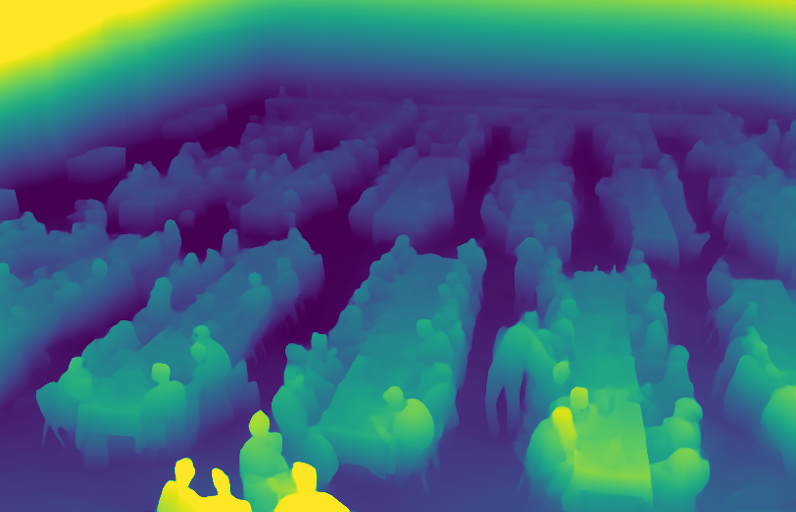} &
\includegraphics[width=0.19\textwidth]{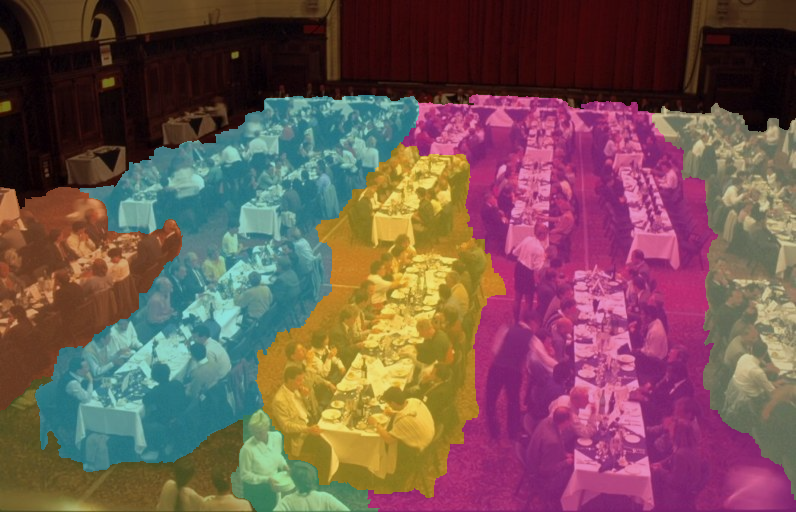} &
\includegraphics[width=0.19\textwidth]{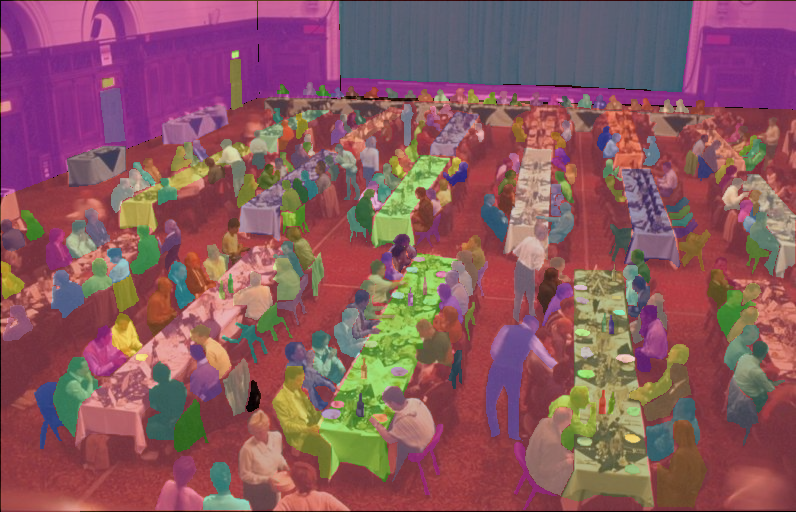}
\end{tabular}
\caption{Qualitative segmentation results from COCO Panoptic, Cityscapes, and ADE20K (top three rows), followed by a crowded ADE20K failure example (bottom row). From left to right: RGB image, estimated monocular depth, detrended depth, our predicted segmentation overlaid on RGB, and ground-truth regions overlaid on RGB. The bottom example illustrates under-segmentation in a scene containing many annotated regions.}
\label{fig:qualitative_examples}
\end{figure*}

\subsection{Cityscapes Sensor-Depth Ablation}
\label{sec:sensor_depth_ablation}

We investigate whether our framework can operate directly on Cityscapes stereo disparity~\cite{cordts2016cityscapes}, without a pretrained depth estimator. Unlike monocular depth predictions, disparity contains missing and quantized measurements, with nonlinear value spacing that compresses differences between distant surfaces. Since our merge criterion depends on depth differences, these properties can affect segmentation even when pixel ordering is preserved.

We compare the original disparity input with global depth detrending and a combination of detrending and disparity-specific preprocessing, including invalid-value handling, logarithmic transformation, and spatial filtering. The logarithmic transformation preserves disparity ordering while changing its numerical spacing, highlighting the importance of depth representation for persistence-based merging. The segmentation algorithm remains unchanged, although the original and subsequent experiments use different threshold-search settings, as detailed in the supplementary material.

Table~\ref{tab:sensor_depth_ablation} shows that the best observed H-mIoU increases from 0.0612 to 0.1045 with detrending and reaches 0.1163 with additional preprocessing; Covering increases from 0.2963 to 0.4994 overall. The final configuration outperforms both evaluated classical baselines on Cityscapes in H-mIoU and Covering, although it remains below the monocular-depth result (0.2086 H-mIoU). These findings demonstrate that our framework can produce meaningful geometric segmentations without any learned component, while highlighting its sensitivity to depth representation.

\begin{table}[t]
\centering
\caption{Cityscapes stereo-disparity ablation on the full 500-image validation set. 
}
\label{tab:sensor_depth_ablation}
\small
\setlength{\tabcolsep}{4pt}
\begin{tabular}{lcc}
\toprule
Disparity preprocessing & H-mIoU & Covering \\
\midrule
Original disparity & 0.0612 & 0.2963 \\
+ Depth detrending & 0.1045 ($k=20$) & 0.4751 ($k=7$) \\
+ Disparity preprocessing & \textbf{0.1163} ($k=25$) & \textbf{0.4994} ($k=7$) \\
\bottomrule
\end{tabular}
\vspace{-5mm}
\end{table}

\subsection{Qualitative Results}
\label{sec:qualitative_results}

Figure~\ref{fig:qualitative_examples} presents qualitative results across COCO Panoptic, Cityscapes, and ADE20K, showing the RGB images, estimated depth maps, detrended depth maps, predicted segmentations, and ground-truth annotations. The examples illustrate how depth-ordered geometric merging identifies meaningful scene regions without learned segmentation representations, while detrending emphasizes local geometric variations. Additional qualitative examples are provided in the supplementary material.

\section{Limitations and Future Directions}
\label{sec:limitations}

DORMS has several limitations. Its segmentation granularity depends on a dataset-specific target region count, which can lead to under-segmentation in crowded scenes. Performance also depends on the quality of the input depth map, particularly the accuracy of geometric boundaries and depth contrast between adjacent objects. Moreover, DORMS produces class-agnostic partitions without semantic supervision.

Future work will focus on improving DORMS through adaptive region-count selection, more efficient threshold search, and enhanced geometric merging strategies. Beyond natural images, segmentation plays an important role in medical imaging~\cite{isensee2021nnunet, gulle2026automated}, aviation~\cite{wang2024valnet,owens2025visual}, and industrial inspection~\cite{bergmann2019mvtec,roth2022total}. Investigating the applicability of DORMS to these domains, including suitable depth representations, presents an important direction for future research.


\section{Conclusion}
\label{sec:conclusion}

We presented DORMS, an algorithmic framework for image segmentation from monocular depth maps that combines depth-ordered merge trees, area--persistence-based scoring, gated spatial-contact constraints, and global depth detrending without learned segmentation parameters or semantic supervision. Experiments on COCO Panoptic, Cityscapes, and ADE20K demonstrate improvements over all evaluated classical baselines in Hungarian-matched mIoU and over their mIoU-selected configurations in Segmentation Covering, alongside competitive performance against zero-shot neural models. Our analysis highlights the importance of depth detrending and geometric merge constraints, demonstrating the potential of lightweight, algorithmic segmentation with increasingly capable depth estimators and sensing technologies.

\bibliographystyle{plainnat}
\bibliography{references}

@article{felzenszwalb2004efficient,
  author  = {Felzenszwalb, Pedro F. and Huttenlocher, Daniel P.},
  title   = {Efficient Graph-Based Image Segmentation},
  journal = {International Journal of Computer Vision},
  year    = {2004},
  volume  = {59},
  number  = {2},
  pages   = {167--181},
  doi     = {10.1023/B:VISI.0000022288.19776.77},
  url     = {https://doi.org/10.1023/B:VISI.0000022288.19776.77}
}

@article{achanta2012slic,
  author  = {Achanta, Radhakrishna and Shaji, Appu and Smith, Kevin and Lucchi, Aur{\'e}lien and Fua, Pascal and S{\"u}sstrunk, Sabine},
  title   = {{SLIC} Superpixels Compared to State-of-the-Art Superpixel Methods},
  journal = {IEEE Transactions on Pattern Analysis and Machine Intelligence},
  year    = {2012},
  volume  = {34},
  number  = {11},
  pages   = {2274--2282},
  doi     = {10.1109/TPAMI.2012.120},
  url     = {https://doi.org/10.1109/TPAMI.2012.120}
}

@article{arbelaez2011contour,
  author  = {Arbel{\'a}ez, Pablo and Maire, Michael and Fowlkes, Charless and Malik, Jitendra},
  title   = {Contour Detection and Hierarchical Image Segmentation},
  journal = {IEEE Transactions on Pattern Analysis and Machine Intelligence},
  year    = {2011},
  volume  = {33},
  number  = {5},
  pages   = {898--916},
  doi     = {10.1109/TPAMI.2010.161},
  url     = {https://doi.org/10.1109/TPAMI.2010.161}
}

@article{liu2016image,
  author  = {Liu, Ting and Seyedhosseini, Mojtaba and Tasdizen, Tolga},
  title   = {Image Segmentation Using Hierarchical Merge Tree},
  journal = {IEEE Transactions on Image Processing},
  year    = {2016},
  volume  = {25},
  number  = {10},
  pages   = {4596--4607},
  doi     = {10.1109/TIP.2016.2592704},
  url     = {https://doi.org/10.1109/TIP.2016.2592704}
}

@inproceedings{jones1997component,
  author    = {Jones, Ronald},
  title     = {Component Trees for Image Filtering and Segmentation},
  booktitle = {Proceedings of the IEEE/EURASIP Workshop on Nonlinear Signal and Image Processing ({NSIP})},
  year      = {1997},
  volume    = {1},
  note      = {Paper 311; no page range provided in the proceedings index},
  url       = {https://www.iwaenc.org/proceedings/1997/nsip97/pdf/author/ns970311.pdf}
}

@article{carr2003computing,
  author  = {Carr, Hamish and Snoeyink, Jack and Axen, Ulrike},
  title   = {Computing Contour Trees in All Dimensions},
  journal = {Computational Geometry},
  year    = {2003},
  volume  = {24},
  number  = {2},
  pages   = {75--94},
  doi     = {10.1016/S0925-7721(02)00093-7},
  url     = {https://doi.org/10.1016/S0925-7721(02)00093-7}
}

@inproceedings{letscher2007image,
  author    = {Letscher, David and Fritts, Jason},
  title     = {Image Segmentation Using Topological Persistence},
  booktitle = {Computer Analysis of Images and Patterns ({CAIP} 2007)},
  series    = {Lecture Notes in Computer Science},
  volume    = {4673},
  pages     = {587--595},
  publisher = {Springer},
  year      = {2007},
  doi       = {10.1007/978-3-540-74272-2_73},
  url       = {https://doi.org/10.1007/978-3-540-74272-2_73}
}

@article{assaf2018persistent,
  author  = {Assaf, Rabih and Goupil, Alban and Vrabie, Valeriu and Boudier, Thomas and Kacim, Mohammad},
  title   = {Persistent Homology for Object Segmentation in Multidimensional Grayscale Images},
  journal = {Pattern Recognition Letters},
  year    = {2018},
  volume  = {112},
  pages   = {277--284},
  doi     = {10.1016/j.patrec.2018.08.007},
  url     = {https://doi.org/10.1016/j.patrec.2018.08.007}
}

@article{chazal2013persistence,
  author  = {Chazal, Fr{\'e}d{\'e}ric and Guibas, Leonidas J. and Oudot, Steve Y. and Skraba, Primoz},
  title   = {Persistence-Based Clustering in Riemannian Manifolds},
  journal = {Journal of the ACM},
  year    = {2013},
  volume  = {60},
  number  = {6},
  pages   = {1--38},
  note    = {Article 41},
  doi     = {10.1145/2535927},
  url     = {https://doi.org/10.1145/2535927}
}

@inproceedings{dimiccoli2009hierarchical,
  author    = {Dimiccoli, Mariella and Salembier, Philippe},
  title     = {Hierarchical Region-Based Representation for Segmentation and Filtering with Depth in Single Images},
  booktitle = {Proceedings of the IEEE International Conference on Image Processing ({ICIP})},
  year      = {2009},
  pages     = {3533--3536},
  doi       = {10.1109/ICIP.2009.5414079},
  url       = {https://doi.org/10.1109/ICIP.2009.5414079}
}

@article{hasnat2016joint,
  author  = {Hasnat, Md. Abul and Alata, Olivier and Tr{\'e}meau, Alain},
  title   = {Joint Color-Spatial-Directional Clustering and Region Merging ({JCSD-RM}) for Unsupervised {RGB-D} Image Segmentation},
  journal = {IEEE Transactions on Pattern Analysis and Machine Intelligence},
  year    = {2016},
  volume  = {38},
  number  = {11},
  pages   = {2255--2268},
  doi     = {10.1109/TPAMI.2015.2513407},
  url     = {https://doi.org/10.1109/TPAMI.2015.2513407}
}

@article{antic2021depth,
  author  = {Anti{\'c}, Milo{\v s} and Zde{\v s}ar, Andrej and {\v S}krjanc, Igor},
  title   = {Depth-Image Segmentation Based on Evolving Principles for {3D} Sensing of Structured Indoor Environments},
  journal = {Sensors},
  year    = {2021},
  volume  = {21},
  number  = {13},
  pages   = {4395},
  doi     = {10.3390/s21134395},
  url     = {https://doi.org/10.3390/s21134395}
}

@inproceedings{yang2024depth,
  author    = {Yang, Lihe and Kang, Bingyi and Huang, Zilong and Zhao, Zhen and Xu, Xiaogang and Feng, Jiashi and Zhao, Hengshuang},
  title     = {{Depth Anything V2}},
  booktitle = {Advances in Neural Information Processing Systems},
  year      = {2024},
  volume    = {37},
  pages     = {21875--21911},
  doi       = {10.52202/079017-0688}
}

@inproceedings{cheng2022masked,
  author    = {Cheng, Bowen and Misra, Ishan and Schwing, Alexander G. and Kirillov, Alexander and Girdhar, Rohit},
  title     = {Masked-Attention Mask Transformer for Universal Image Segmentation},
  booktitle = {Proceedings of the IEEE/CVF Conference on Computer Vision and Pattern Recognition ({CVPR})},
  year      = {2022},
  pages     = {1290--1299},
  doi       = {10.1109/CVPR52688.2022.00135},
  url       = {https://openaccess.thecvf.com/content/CVPR2022/html/Cheng_Masked-Attention_Mask_Transformer_for_Universal_Image_Segmentation_CVPR_2022_paper.html}
}

@inproceedings{kirillov2023segment,
  author    = {Kirillov, Alexander and Mintun, Eric and Ravi, Nikhila and Mao, Hanzi and Rolland, Chloe and Gustafson, Laura and Xiao, Tete and Whitehead, Spencer and Berg, Alexander C. and Lo, Wan-Yen and Dollar, Piotr and Girshick, Ross},
  title     = {Segment Anything},
  booktitle = {Proceedings of the IEEE/CVF International Conference on Computer Vision ({ICCV})},
  year      = {2023},
  pages     = {4015--4026},
  doi       = {10.1109/ICCV51070.2023.00371},
  url       = {https://openaccess.thecvf.com/content/ICCV2023/html/Kirillov_Segment_Anything_ICCV_2023_paper.html}
}

@inproceedings{ravi2024sam2,
  author    = {Ravi, Nikhila and Gabeur, Valentin and Hu, Yuan-Ting and Hu, Ronghang and Ryali, Chaitanya and Ma, Tengyu and Khedr, Haitham and R{\"a}dle, Roman and Rolland, Chloe and Gustafson, Laura and Mintun, Eric and Pan, Junting and Alwala, Kalyan Vasudev and Carion, Nicolas and Wu, Chao-Yuan and Girshick, Ross and Dollar, Piotr and Feichtenhofer, Christoph},
  title     = {{SAM 2}: Segment Anything in Images and Videos},
  booktitle = {International Conference on Learning Representations ({ICLR})},
  year      = {2025}
}

@inproceedings{xu2022groupvit,
  author    = {Xu, Jiarui and De Mello, Shalini and Liu, Sifei and Byeon, Wonmin and Breuel, Thomas and Kautz, Jan and Wang, Xiaolong},
  title     = {{GroupViT}: Semantic Segmentation Emerges from Text Supervision},
  booktitle = {Proceedings of the IEEE/CVF Conference on Computer Vision and Pattern Recognition ({CVPR})},
  year      = {2022},
  pages     = {18134--18144},
  doi       = {10.1109/CVPR52688.2022.01760},
  url       = {https://openaccess.thecvf.com/content/CVPR2022/html/Xu_GroupViT_Semantic_Segmentation_Emerges_From_Text_Supervision_CVPR_2022_paper.html}
}

@inproceedings{lueddecke2022image,
  author    = {L{\"u}ddecke, Timo and Ecker, Alexander},
  title     = {Image Segmentation Using Text and Image Prompts},
  booktitle = {Proceedings of the IEEE/CVF Conference on Computer Vision and Pattern Recognition ({CVPR})},
  year      = {2022},
  pages     = {7086--7096},
  doi       = {10.1109/CVPR52688.2022.00695},
  url       = {https://openaccess.thecvf.com/content/CVPR2022/html/Luddecke_Image_Segmentation_Using_Text_and_Image_Prompts_CVPR_2022_paper.html}
}

@inproceedings{couairon2024diffcut,
  author    = {Couairon, Paul and Shukor, Mustafa and Haugeard, Jean-Emmanuel and Cord, Matthieu and Thome, Nicolas},
  title     = {{DiffCut}: Catalyzing Zero-Shot Semantic Segmentation with Diffusion Features and Recursive Normalized Cut},
  booktitle = {Advances in Neural Information Processing Systems},
  year      = {2024},
  volume    = {37},
  pages     = {13548--13578},
  doi       = {10.52202/079017-0433}
}

@inproceedings{long2015fully,
  author    = {Long, Jonathan and Shelhamer, Evan and Darrell, Trevor},
  title     = {Fully Convolutional Networks for Semantic Segmentation},
  booktitle = {Proceedings of the IEEE Conference on Computer Vision and Pattern Recognition ({CVPR})},
  year      = {2015},
  pages     = {3431--3440},
  doi       = {10.1109/CVPR.2015.7298965},
  url       = {https://openaccess.thecvf.com/content_cvpr_2015/html/Long_Fully_Convolutional_Networks_2015_CVPR_paper.html}
}

@inproceedings{he2017mask,
  author    = {He, Kaiming and Gkioxari, Georgia and Dollar, Piotr and Girshick, Ross},
  title     = {{Mask R-CNN}},
  booktitle = {Proceedings of the IEEE International Conference on Computer Vision ({ICCV})},
  year      = {2017},
  pages     = {2961--2969},
  doi       = {10.1109/ICCV.2017.322},
  url       = {https://openaccess.thecvf.com/content_iccv_2017/html/He_Mask_R-CNN_ICCV_2017_paper.html}
}

@inproceedings{kirillov2019panoptic,
  author    = {Kirillov, Alexander and He, Kaiming and Girshick, Ross and Rother, Carsten and Dollar, Piotr},
  title     = {Panoptic Segmentation},
  booktitle = {Proceedings of the IEEE/CVF Conference on Computer Vision and Pattern Recognition ({CVPR})},
  year      = {2019},
  pages     = {9404--9413},
  doi       = {10.1109/CVPR.2019.00963},
  url       = {https://openaccess.thecvf.com/content_CVPR_2019/html/Kirillov_Panoptic_Segmentation_CVPR_2019_paper.html}
}

@inproceedings{lin2014microsoft,
  author    = {Lin, Tsung-Yi and Maire, Michael and Belongie, Serge and Hays, James and Perona, Pietro and Ramanan, Deva and Doll{\'a}r, Piotr and Zitnick, C. Lawrence},
  title     = {{Microsoft COCO}: Common Objects in Context},
  booktitle = {Computer Vision -- {ECCV} 2014},
  series    = {Lecture Notes in Computer Science},
  volume    = {8693},
  pages     = {740--755},
  publisher = {Springer},
  year      = {2014},
  doi       = {10.1007/978-3-319-10602-1_48},
  url       = {https://doi.org/10.1007/978-3-319-10602-1_48}
}

@inproceedings{cordts2016cityscapes,
  author    = {Cordts, Marius and Omran, Mohamed and Ramos, Sebastian and Rehfeld, Timo and Enzweiler, Markus and Benenson, Rodrigo and Franke, Uwe and Roth, Stefan and Schiele, Bernt},
  title     = {The {Cityscapes} Dataset for Semantic Urban Scene Understanding},
  booktitle = {Proceedings of the IEEE Conference on Computer Vision and Pattern Recognition ({CVPR})},
  year      = {2016},
  pages     = {3213--3223},
  doi       = {10.1109/CVPR.2016.350},
  url       = {https://openaccess.thecvf.com/content_cvpr_2016/html/Cordts_The_Cityscapes_Dataset_CVPR_2016_paper.html}
}

@inproceedings{zhou2017scene,
  author    = {Zhou, Bolei and Zhao, Hang and Puig, Xavier and Fidler, Sanja and Barriuso, Adela and Torralba, Antonio},
  title     = {Scene Parsing Through {ADE20K} Dataset},
  booktitle = {Proceedings of the IEEE Conference on Computer Vision and Pattern Recognition ({CVPR})},
  year      = {2017},
  pages     = {633--641},
  doi       = {10.1109/CVPR.2017.544},
  url       = {https://openaccess.thecvf.com/content_cvpr_2017/html/Zhou_Scene_Parsing_Through_CVPR_2017_paper.html}
}

@article{gulle2026automated,
  title={Automated Segmentation of Intrinsic Foot Muscles in Ultrasound Using Deep Neural Networks: Development and Validation Study},
  author={Gulle, Halime and Mumcu, Furkan and Goodrich, Austin G and Bekit, Lokman and Haksal, Pinar Kuyulu and Johnson, A Wayne and Davis, Irene and Yilmaz, Yasin},
  journal={Ultrasound in Medicine \& Biology},
  year={2026},
  publisher={Elsevier}
}

@article{isensee2021nnunet,
  title={nnU-Net: A Self-Configuring Method for Deep Learning-Based Biomedical Image Segmentation},
  author={Isensee, Fabian and Jaeger, Paul F. and Kohl, Simon A. A. and Petersen, Jens and Maier-Hein, Klaus H.},
  journal={Nature Methods},
  volume={18},
  number={2},
  pages={203--211},
  year={2021},
  doi={10.1038/s41592-020-01008-z}
}

@article{wang2024valnet,
  title={VALNet: Vision-Based Autonomous Landing with Airport Runway Instance Segmentation},
  author={Wang, Qiang and Feng, Wenquan and Zhao, Hongbo and Liu, Binghao and Lyu, Shuchang},
  journal={Remote Sensing},
  volume={16},
  number={12},
  pages={2161},
  year={2024},
  doi={10.3390/rs16122161}
}

@inproceedings{owens2025visual,
  title={Visual Segmentation for Autonomous Aircraft Landing on Austere Runways},
  author={Owens, Alissa and Taylor, Clark and Nykl, Scott},
  booktitle={Proceedings of the 2025 International Technical Meeting of the Institute of Navigation},
  pages={23--36},
  year={2025},
  doi={10.33012/2025.19972}
}

@inproceedings{bergmann2019mvtec,
  title={MVTec AD -- A Comprehensive Real-World Dataset for Unsupervised Anomaly Detection},
  author={Bergmann, Paul and Fauser, Michael and Sattlegger, David and Steger, Carsten},
  booktitle={Proceedings of the IEEE/CVF Conference on Computer Vision and Pattern Recognition},
  pages={9592--9600},
  year={2019},
  doi={10.1109/CVPR.2019.00982}
}

@inproceedings{roth2022total,
  title={Towards Total Recall in Industrial Anomaly Detection},
  author={Roth, Karsten and Pemula, Latha and Zepeda, Joaquin and Sch{\"o}lkopf, Bernhard and Brox, Thomas and Gehler, Peter},
  booktitle={Proceedings of the IEEE/CVF Conference on Computer Vision and Pattern Recognition},
  pages={14318--14328},
  year={2022},
  doi={10.1109/CVPR52688.2022.01392}
}

\clearpage
\onecolumn
\startcontents[appendix]

\begingroup
\normalsize
\setlength{\parskip}{0pt}
\setlength{\baselineskip}{11.5pt}
\vspace{-2mm}

\section*{Appendix Contents}
\printcontents[appendix]{}{1}{
    \setcounter{tocdepth}{2}
}

\endgroup

\clearpage
\twocolumn
\appendix
\section{Implementation Details}
\label{sec:supp_implementation}

All reported configurations of our method use global depth detrending, area--persistence-based merge scoring, and the gated spatial-contact constraint described in Sec.~\ref{sec:method}. The segmentation stage operates exclusively on depth values and does not require learned parameters. This section provides the dataset-specific configurations, hyperparameter selection procedures, depth inputs, and implementation settings used in our experiments.

\subsection{Hyperparameter Configurations and Selection}
\label{sec:supp_hyperparameters}

The principal hyperparameters are the target region count $k$, the minimum accumulated contact threshold $m$, and the area-gate fraction $g$. Table~\ref{tab:method_configs} summarizes the configurations used for the main monocular-depth experiments.

\begin{table}[t]
\centering
\caption{Dataset-specific configurations of the proposed method. The area gate $g$ is expressed as a percentage of the image area.}
\label{tab:method_configs}
\small
\begin{tabular}{lccc}
\toprule
Dataset & $k$ & $m$ & $g$ \\
\midrule
COCO Panoptic & 6 & 16 & 0.2\% \\
Cityscapes & 30 & 3 & 10\% \\
ADE20K & 8 & 8 & 2\% \\
\bottomrule
\end{tabular}
\end{table}

\paragraph{COCO Panoptic.}
The target region count is selected using a disjoint subset of COCO train2017, with reweighting to reflect the natural distribution of scene complexity. Table~\ref{tab:coco_k_selection} presents the corresponding Hungarian-matched mIoU values. The best result occurs at $k=6$, with lower performance at both neighboring values $k=5$ and $k=7$. Thus, the selected configuration is an interior maximum of the evaluated range rather than a boundary value, and there is no tie. The remaining hyperparameters, $m=16$ and $g=0.2\%$, were fixed beforehand based on separate val2017 experiments and were not independently selected using the held-out subset. Consequently, only the selection of $k$ uses the disjoint held-out data.

\begin{table}[t]
\centering
\caption{Target region count selection on the held-out COCO train2017 subset, using naturally reweighted Hungarian-matched mIoU.}
\label{tab:coco_k_selection}
\small
\setlength{\tabcolsep}{5pt}
\resizebox{\columnwidth}{!}{%
\begin{tabular}{lcccccc}
\toprule
$k$ & 4 & 5 & 6 & 7 & 8 & 10 \\
\midrule
H-mIoU & 0.20114 & 0.20273 & \textbf{0.20357} & 0.20102 & 0.19624 & 0.18822 \\
\bottomrule
\end{tabular}%
}
\end{table}

\paragraph{Cityscapes.}
We evaluate the parameter grid $k\in\{4,5,6,7,8,10,12,15,17,20,25,30\}$ and $m\in\{1,3,16\}$, with the area gate fixed at $g=10\%$. The highest Hungarian-matched mIoU is 0.2086 at $k=30$. This maximum is attained by both $m=3$ and $m=16$, which produce identical recorded results to floating-point precision across all 12 evaluated target counts. We adopt $m=3$ as the representative configuration. The area-gate fraction was not swept on Cityscapes; $g=10\%$ is the only evaluated setting. These parameters were selected using the Cityscapes validation set itself rather than a separate held-out selection split.

\paragraph{ADE20K.}
We evaluate $k\in\{4,5,6,7,8,10,12,15,17\}$, $m\in\{1,3,5,8,16,32\}$, and $g\in\{2\%,10\%\}$. The highest Hungarian-matched mIoU, 0.2206, occurs uniquely at $(k,m,g)=(8,8,2\%)$, with no tied configurations. As with Cityscapes, these parameters are selected from sweeps on the evaluation split, rather than a separate held-out set. Accordingly, the reported Cityscapes and ADE20K results should be interpreted as validation-selected configurations.

\subsection{Threshold Search and Region Selection}
\label{sec:supp_threshold_search}

The threshold search employs bounded adaptive subdivision, as described in Sec.~\ref{sec:threshold_search}. The search interval starts at $\tau_{\min}=10^{-6}$ and extends to $\tau_{\max}=\max(10^6,20s_{\max})$, where $s_{\max}$ is the maximum merge score used to establish the image-dependent upper bound. The reported runners use a maximum of 2,000 segmentation replays, with a minimum bracket ratio of 1.0005.

To limit exploration of unnecessarily large region counts, the search uses a cap of $\max(24,3k_{\max})$, where $k_{\max}$ is the largest target region count in the evaluated grid. The cap is computed once for that grid. A search bracket is pruned only when the region counts at both endpoints exceed the cap. Consequently, the algorithm does not exhaustively resolve all region counts, particularly those substantially above the requested target.

For each discovered region count, the search retains a witness threshold that produced it. After selecting the preferred count, the segmentation is reconstructed through a fresh replay at the corresponding witness threshold. This avoids assuming that an arbitrary threshold within a bracket reproduces the same count, which is not guaranteed because the region-count function can be non-monotonic.

We use the \texttt{prefer\_over} selection policy: the smallest discovered count greater than or equal to $k$ is preferred, with the closest discovered count used only when no such candidate exists. In the latter case, equal-distance ties favor the smaller count. Across all three primary evaluation datasets, the preferred-over-target branch was selected for every image: 5,000/5,000 on COCO, 500/500 on Cityscapes, and 2,000/2,000 on ADE20K. Thus, all selected region counts before cleanup were at least $k$; no fallback to a below-target count was required.

\subsection{Small-Region Cleanup}
\label{sec:supp_cleanup}

The gated spatial-contact constraint may leave small components that cannot accumulate sufficient contact to merge with neighboring regions. To address this, we apply the small-region cleanup procedure described in Sec.~\ref{sec:region_cleanup}. The minimum component size is
\begin{equation}
A_{\min}=\max\left(2,\left\lfloor 0.0005|\Omega|\right\rfloor\right).
\end{equation}
The absolute minimum size and relative threshold are identical across all datasets and reported configurations. For illustration, this corresponds to 153 pixels at a resolution of $640\times480$ and 1,048 pixels at $2048\times1024$.

Cleanup is performed only when the neck-width constraint is enabled ($m>1$) and $A_{\min}>1$. Components smaller than the threshold are removed, and their pixels are assigned to the nearest surviving regions through a Euclidean distance transform, followed by contiguous relabeling. When $m=1$, cleanup is disabled.

Table~\ref{tab:cleanup_effect} quantifies the effect of this operation. On COCO, the mean region count decreases from 6.054 to 5.927, while on ADE20K it decreases from 8.021 to 7.886. The effect is considerably larger on Cityscapes, where the mean count decreases from 30.034 to 25.842, corresponding to approximately 4.2 removed regions per image. Cleanup reduces the final count below the specified target in 7.4\% of COCO images, 92.0\% of Cityscapes images, and 8.8\% of ADE20K images.

\begin{table}[t]
\centering
\caption{Effect of small-region cleanup on the three evaluation datasets. Region counts are averaged over images. The final column reports the number and percentage of images with fewer than $k$ regions after cleanup.}
\label{tab:cleanup_effect}
\small
\setlength{\tabcolsep}{4pt}
\begin{tabular}{lcccc}
\toprule
Dataset & Before & After & Removed & Below $k$ \\
\midrule
COCO & 6.054 & 5.927 & 0.127 & 371 (7.4\%) \\
Cityscapes & 30.034 & 25.842 & 4.192 & 460 (92.0\%) \\
ADE20K & 8.021 & 7.886 & 0.135 & 175 (8.8\%) \\
\bottomrule
\end{tabular}
\end{table}

These results also clarify the relationship between the target region count and the reported segmentation granularity. In all three datasets, the search selects counts close to the requested target before cleanup. The substantially lower final average on Cityscapes is primarily a consequence of removing small residual components, rather than an inability of the threshold search to reach the requested region count. All region counts reported in the main experimental comparisons are measured after cleanup.

\subsection{Depth Inputs and Preprocessing}
\label{sec:supp_depth_inputs}

For the main experiments, depth maps are generated using pretrained Depth Anything V2 (DA-V2) models~\cite{yang2024depth}. The exact checkpoints are summarized in Table~\ref{tab:depth_checkpoints}.

\begin{table}[t]
\centering
\caption{Pretrained depth estimation models used in the main experiments. Model sizes refer to approximate parameter counts.}
\label{tab:depth_checkpoints}
\small
\begin{tabular}{lcc}
\toprule
Dataset & DA-V2 checkpoint & Parameters \\
\midrule
COCO Panoptic & Small & 24.8M \\
Cityscapes & Large & 335.3M \\
ADE20K & Small & 24.8M \\
\bottomrule
\end{tabular}
\end{table}

The corresponding Hugging Face model identifiers are (depth-anything/) \texttt{Depth-Anything-V2-Small-hf} and \texttt{Depth-Anything-V2-Large-hf}. COCO and ADE20K use the Small checkpoint, whereas Cityscapes uses Large. The Cityscapes depth maps were retained from an earlier depth-source comparison that evaluated multiple estimators at the Large model scale. Thus, the checkpoint difference reflects the depth-source evaluation protocol rather than a uniform model-size selection across datasets.

For the separate Cityscapes sensor-depth ablation, we use the stereo disparity maps distributed with the dataset instead of neural-estimated depth. These are stored as 16-bit PNG images with valid disparity values decoded as $(p-1)/256$, where $p$ is the encoded pixel value and $p=0$ denotes an invalid measurement. Approximately 22.1\% of pixels contain invalid measurements. These pixels are initially filled using their nearest valid neighbors before segmentation. The additional disparity-specific preprocessing variants and their effects are examined in Sec.~\ref{sec:sensor_depth_ablation} and the supplementary disparity analysis.

After obtaining the input depth representation, the main pipeline applies global linear detrending and min--max normalization before constructing the depth-ordered segmentation. No RGB features, semantic category labels, or learned segmentation parameters are used in the subsequent algorithm.

\section{Detailed DORMS Algorithm}
\label{app:dorms_algorithm}

We provide procedural details of DORMS (Depth-Ordered Region Merging for Segmentation). Its four stages are global depth detrending, a depth-ordered union--find sweep at a prescribed merge threshold, bounded threshold selection for a target number of regions, and conditional cleanup. The input depth convention is that larger values indicate nearer locations; when a sensor reports metric distance, its ordering must be reversed before the sweep. All stages after depth estimation are deterministic and require no learned parameters.

\paragraph{Notation and preprocessing.}
Let $\Omega$ denote the $H\times W$ pixel grid, $N=HW$, and $\mathcal{N}_4(p)$ the four-connected neighbors of pixel $p$. Given the input depth map $d$, DORMS estimates a global plane by ordinary least squares and normalizes the residual to $[0,1]$:
\begin{align}
(\hat a,\hat b,\hat c)&=\arg\min_{a,b,c}\sum_{(x,y)\in\Omega}\!
[d(x,y)-ax-by-c]^2,\nonumber\\
\tilde d&=\operatorname{Norm}_{[0,1]}\big(d-\hat a x-\hat b y-\hat c\big).
\label{eq:supp_dorms_detrending}
\end{align}
Pixels are activated in descending order of $\tilde d$, with ties resolved by a fixed pixel index. Each active component $C$ maintains its area $A(C)$ and birth depth $b(C)$, inherited from its earliest activated pixel. When components meet, the elder is the one with the earlier birth (larger birth depth); equal-depth births are resolved by the smallest root index. The junior is determined by birth order, not by component size.

\begin{algorithm}[tbp]
\caption{DORMS: complete segmentation pipeline}
\label{alg:supp_dorms_pipeline}
\footnotesize
\begin{algorithmic}[1]
\Require Depth map $d$, target count $k$, contact threshold $m$, area gate $g$
\Ensure Final partition $\widehat L$
\State $\tilde d\gets\Call{DetrendAndNormalize}{d}$ \Comment{Eq.~\eqref{eq:supp_dorms_detrending}}
\State $s_{\max}\gets\Call{MaxMergeScoreNoVeto}{\tilde d}$
\State \hspace{1em}\Comment{One preliminary full depth sweep per image}
\State $(\tau^\star,L,R)\gets\Call{SelectThreshold}{\tilde d,k,m,g,s_{\max}}$
\State $a_{\min}\gets\max(2,\lfloor 0.0005N\rfloor)$
\If{$m>1$ \textbf{and} $a_{\min}>1$}
    \State $L\gets\Call{CleanupSmallRegions}{L,a_{\min}}$
\EndIf
\State \Return $\widehat L$
\end{algorithmic}
\end{algorithm}

\paragraph{Fixed-threshold sweep.}
For a specified threshold $\tau$, Algorithm~\ref{alg:supp_dorms_sweep} constructs a partition using a union--find forest. A newly activated pixel with no active neighbors creates a component; with one neighboring component, it is absorbed without a merge test. When multiple distinct neighboring components are encountered, the pixel is assigned to their elder, and every other component is considered as a junior merge candidate. The initial neighbor set contains distinct active roots; since accepted junior components are attached only to the elder, each unprocessed junior remains a valid root during the candidate loop. Crucially, the activating pixel joins the elder whether or not any junior merge is approved. The same surviving component pair may be encountered at subsequent bridging pixels and evaluated again using its current state.

\begin{algorithm}[tbp]
\caption{\textsc{Replay}: one fixed-threshold depth sweep}
\label{alg:supp_dorms_sweep}
\footnotesize
\begin{algorithmic}[1]
\Require Detrended depth $\tilde d$, threshold $\tau$, parameters $m,g$
\Ensure Partition $L$, number of regions $R$
\State $\pi\gets\Call{SortDescending}{\tilde d}$ \Comment{Tie: pixel index}
\State Initialize empty union--find forest and contact counters $c$
\For{$p$ in $\pi$}
    \State $S\gets\{\Call{Find}{q}:q\in\mathcal N_4(p),\ q\text{ active}\}$
    \State Remove duplicate roots from $S$
    \If{$S=\varnothing$}
        \State Create component $\{p\}$ with $A=1$, $b=\tilde d(p)$
    \ElsIf{$|S|=1$}
        \State Attach $p$ to the sole component in $S$
    \Else
        \State $e\gets\Call{Eldest}{S}$ \Comment{Birth depth; root-index tie}
        \For{$j\in S\setminus\{e\}$}
            \If{$\Call{AdmitMerge}{e,j,p,\tau,m,g,c}$}
                \State $\Call{UnionIntoElder}{e,j}$ \Comment{Update current area}
            \EndIf
        \EndFor
        \State Attach $p$ to elder $e$ regardless of prior decisions
    \EndIf
\EndFor
\State $L\gets$ labels induced by the final union--find roots
\State $R\gets$ number of distinct roots in $L$
\State \Return $(L,R)$
\end{algorithmic}
\end{algorithm}

\paragraph{Merge admissibility.}
For an encounter at bridging pixel $p$, the junior-component persistence and area--persistence score are
\begin{equation}
P(p,j)=b(C_j)-\tilde d(p),\qquad
\varphi(p,j)=A(C_j)\,P(p,j).
\label{eq:supp_dorms_score}
\end{equation}
Algorithm~\ref{alg:supp_dorms_admit} tests accumulated spatial contact before this score. Whenever $m>1$, contact is counted at every bridging encounter, including encounters at which the area gate is inactive. The contact veto is applied only when \emph{both} components have area at least $gN$; otherwise the candidate proceeds directly to the score test. The score permits merging only when $\varphi<\tau$. Thus, a contact-based deferral may eventually reach the score test as contact accumulates, whereas a score-based rejection is irreversible: the junior's area and persistence cannot decrease during the descending-depth sweep. Deferred contact does not make the score test easier to pass.\footnote{The optimized contact-counter hash table limits each lookup to 64 probes. If this limit is exceeded, the implementation treats contact as sufficient for that encounter (a fail-open contact check); the area--persistence score test still applies. The pseudocode states the intended merge rule without this storage-level fallback.}

\begin{algorithm}[tbp]
\caption{\textsc{AdmitMerge}: merge decision}
\label{alg:supp_dorms_admit}
\footnotesize
\begin{algorithmic}[1]
\Require Current elder $e$, junior $j$, bridging pixel $p$, threshold $\tau$
\Ensure Boolean decision
\If{$m>1$}
    \State $c[e,j]\gets c[e,j]+1$ \Comment{Independent of area gate}
    \If{$A(e)\ge gN$ \textbf{and} $A(j)\ge gN$}
        \If{$c[e,j]<m$}
            \State \Return \textbf{false} \Comment{Contact-deferred}
        \EndIf
    \EndIf
\EndIf
\State $P\gets b(j)-\tilde d(p)$
\State $s\gets A(j)\,P$ \Comment{Current junior area}
\State \Return $(s<\tau)$ \Comment{Reject at $s\ge\tau$}
\end{algorithmic}
\end{algorithm}

\paragraph{Bounded threshold selection.}
Only the outer search in Algorithm~\ref{alg:supp_dorms_search} receives the target count $k$; a fixed-threshold replay has no knowledge of $k$. Before selecting thresholds, an additional full no-veto sweep computes the maximum candidate area--persistence score $s_{\max}$, used to set a per-image upper threshold. The search records a \emph{witness threshold} for each discovered region count $R(\tau)$. Since earlier merge decisions affect later component states, $R(\tau)$ is not guaranteed to be monotonic and standard bisection is inappropriate. The implementation recursively subdivides only brackets whose endpoints have \emph{different} counts, using the geometric mean and exploring the lower-threshold half before the upper half. Subdivision also stops at the replay budget, the minimum bracket ratio, or when both endpoint counts exceed the region-count cap. Consequently, the search selects among \emph{discovered} counts, not an exhaustive set of all achievable counts.

\begin{algorithm}[tbp]
\caption{\textsc{SelectThreshold}: bounded adaptive search}
\label{alg:supp_dorms_search}
\footnotesize
\begin{algorithmic}[1]
\Require Depth $\tilde d$, target $k$, parameters $m,g$, precomputed $s_{\max}$
\Ensure Witness $\tau^\star$, partition $L^\star$, pre-cleanup count $R^\star$
\State $\tau_\ell\gets10^{-6}$; $\tau_u\gets\max(10^6,20s_{\max})$
\State $B\gets$ replay budget; $\rho\gets1.0005$
\State $C\gets\max(24,3k_{\max})$ \Comment{$k_{\max}=k$ for one target}
\State $\mathcal W\gets\varnothing$ \Comment{Count $\mapsto$ witness threshold}
\State $(L_\ell,R_\ell)\gets\Call{Replay}{\tilde d,\tau_\ell,m,g}$
\State $(L_u,R_u)\gets\Call{Replay}{\tilde d,\tau_u,m,g}$
\State Store $\mathcal W[R_\ell]\gets\tau_\ell$; set $\mathcal W[R_u]\gets\tau_u$ if $R_u\notin\operatorname{dom}(\mathcal W)$
\State $Q\gets[(\tau_\ell,\tau_u,R_\ell,R_u)]$ \Comment{LIFO stack}
\While{$Q\ne\varnothing$ \textbf{and} replays $<B$}
    \State $(a,b,R_a,R_b)\gets\Call{Pop}{Q}$
    \If{$R_a=R_b$ \textbf{or} $b/a<\rho$}
        \State \textbf{continue} \Comment{No endpoint change; narrow bracket}
    \EndIf
    \If{$R_a>C$ \textbf{and} $R_b>C$}
        \State \textbf{continue} \Comment{Both counts above cap}
    \EndIf
    \State $t\gets\sqrt{ab}$ \Comment{Geometric midpoint}
    \State $(L_t,R_t)\gets\Call{Replay}{\tilde d,t,m,g}$
    \If{$R_t\notin\operatorname{dom}(\mathcal W)$}
        \State $\mathcal W[R_t]\gets t$
    \EndIf
    \State \Call{Push}{$Q,(t,b,R_t,R_b)$} \Comment{Right half first}
    \State \Call{Push}{$Q,(a,t,R_a,R_t)$} \Comment{Left half explored first}
\EndWhile
\State $\mathcal R\gets\operatorname{dom}(\mathcal W)$
\If{$\{R\in\mathcal R:R\ge k\}\ne\varnothing$}
    \State $R^\star\gets\min\{R\in\mathcal R:R\ge k\}$
\Else
    \State $R^\star\gets\arg\min_{R\in\mathcal R}|R-k|$
    \State \hspace{1em}\Comment{Ties favor smaller $R$}
\EndIf
\State $\tau^\star\gets\mathcal W[R^\star]$
\State $(L^\star,R^\star)\gets\Call{Replay}{\tilde d,\tau^\star,m,g}$
\State \Return $(\tau^\star,L^\star,R^\star)$
\end{algorithmic}
\end{algorithm}

Here $s_{\max}$ is obtained from the preliminary no-veto sweep in Algorithm~\ref{alg:supp_dorms_pipeline}, rather than assumed to be known. The main experimental runners allow up to 2,000 search replays per image; the lower-level routine defaults to 500, used in some separate analyses. For a multi-$k$ evaluation grid, $k_{\max}$ is set from the complete grid so the cap remains constant across targets. Equal endpoint counts terminate refinement even though non-monotonicity means that distinct interior counts cannot be ruled out. The fixed left-first depth-first order, together with the finite budget, determines which counts are actually discovered. The per-image upper bound and cap are supplied by the evaluation runners and should not be confused with lower-level routine defaults.

\paragraph{Conditional small-region cleanup.}
When $m>1$, components smaller than $a_{\min}=\max(2,\lfloor0.0005N\rfloor)$ are reassigned to the nearest surviving region by Euclidean distance transform and relabeled consecutively; cleanup is skipped when $m=1$ or the size threshold is at most one. Threshold selection operates on the pre-cleanup partition, whereas reported region counts are post-cleanup. On COCO with $k=6$, the mean count decreases from 6.054 to 5.927 after cleanup; none of the 5,000 search outputs undershot $k$.

\section{Depth-Source and Depth-Estimator Ablation}
\label{app:depth_backbone_ablation}

We investigate how the choice of depth source affects DORMS using five depth representations on Cityscapes: sensor-derived stereo disparity, Depth Anything V1~\cite{yang2024depth}, Depth Anything V2~\cite{yang2024depth}, ZoeDepth, and Marigold. The evaluated checkpoints are \texttt{LiheYoung/depth-anything-large-hf}, \texttt{depth-anything/Depth-Anything-V2-Large-hf}, \texttt{Intel/zoedepth-nyu-kitti}, and \texttt{prs-eth/marigold-depth-v1-0}. Unlike the learned depth estimators, stereo disparity is obtained directly from the Cityscapes sensor measurements. In all cases, the subsequent segmentation algorithm remains unchanged.

\paragraph{Experimental Setup.}
We evaluate 50 stratified Cityscapes validation images, considering each depth source with and without global plane subtraction. For each of the resulting ten conditions, we sweep the target region count $k\in\{7,20\}$, spatial-contact threshold $m\in\{1,3,5,8,16,32\}$, and area-gating fraction $g\in\{2\%,10\%\}$. This produces 12,000 evaluated configurations across the 50 images, with no execution failures. We report the highest mean Hungarian-matched mIoU (H-mIoU) obtained for each depth representation and preprocessing condition. Since configurations are selected using the same 50-image subset, these results characterize sensitivity to depth representation rather than independently validated performance.

\begin{table}[t]
\centering
\caption{Depth-source ablation on 50 stratified Cityscapes images. We report the highest mean H-mIoU over the evaluated segmentation configurations for each depth source, with and without global depth detrending. The improvement ($\Delta$) compares the best configurations within each condition, which may use different segmentation hyperparameters.}
\label{tab:supp_depth_backbones}
\small
\setlength{\tabcolsep}{5pt}
\begin{tabular}{lccc}
\toprule
Depth source & Original & Detrended & $\Delta$ \\
\midrule
Stereo disparity & 0.0648 & 0.1023 & +0.0375 \\
Marigold & 0.0807 & 0.1195 & +0.0388 \\
Depth Anything V1 & 0.0884 & 0.1625 & +0.0741 \\
ZoeDepth & 0.0922 & 0.1307 & +0.0385 \\
Depth Anything V2 & \textbf{0.1110} & \textbf{0.1919} & \textbf{+0.0809} \\
\bottomrule
\end{tabular}
\end{table}

\paragraph{Results.}
Table~\ref{tab:supp_depth_backbones} shows substantial variation across depth sources. Without detrending, all four monocular depth estimators outperform stereo disparity, with Depth Anything V2 achieving the highest H-mIoU (0.1110 versus 0.0648). The paired improvement of Depth Anything V2 over disparity is 0.0461, with a 95\% bootstrap confidence interval of [0.0301, 0.0609], and improvements on 43 of the 50 images.

Global detrending improves the best observed H-mIoU for every depth source, with the largest gain obtained using Depth Anything V2 (+0.0809). Combining Depth Anything V2 with detrending increases H-mIoU from 0.0648 for the original disparity baseline to 0.1919, an absolute difference of 0.1271 (approximately $2.96\times$). These comparisons suggest that segmentation quality depends substantially on the information and numerical structure preserved by the input depth representation. However, because each condition uses its best configuration from the evaluated grid, the differences should not be interpreted as strictly controlled, single-factor effects.

We additionally examine whether missing stereo measurements explain the gap between disparity and monocular depth. The per-image analysis finds negligible correlation with the invalid-disparity fraction ($r=0.017$, $p=0.91$), providing no evidence that missing measurements alone account for the observed difference. Moreover, the H-mIoU variation across spatial-contact thresholds increases from approximately 0.0005 in the earlier disparity-based analysis to 0.0107 with detrended Depth Anything V2. This indicates that the effectiveness of geometric merge constraints also depends on the depth representation.

\paragraph{Scope.}
This ablation is limited to 50 Cityscapes images and should not be interpreted as a full-dataset model ranking. The subsequent full-validation experiment evaluates detrended Depth Anything V2 Large on all 500 Cityscapes validation images, achieving 0.2086 H-mIoU under the final DORMS configuration, as reported in Table~\ref{tab:segmentation_comparison}. COCO and ADE20K use Depth Anything V2 Small in the main experiments; cross-estimator comparisons were not conducted on those datasets. Metric3D v2 and UniDepth v2 were considered but excluded because of environment compatibility constraints.

Overall, these experiments demonstrate that DORMS can operate with different learned or sensor-derived depth sources, while revealing substantial sensitivity to depth estimation quality and representation. They further support the importance of global detrending and show that the contribution of spatial-contact constraints can vary with the input depth source.

\section{Evaluation Metrics and Scoring Protocol}
\label{app:evaluation_protocol}

We evaluate all segmentation methods using a unified class-agnostic protocol based on Hungarian-matched mean Intersection-over-Union (H-mIoU) and Segmentation Covering. Both metrics operate on complete image partitions and disregard predicted semantic category labels. For each image, pixels marked as void or unlabeled in the ground-truth annotations are excluded from evaluation. All methods are evaluated using the same ground-truth partitions and scoring implementation, rather than adopting results reported under potentially different protocols in their original publications.

\subsection{Hungarian-Matched mIoU}
\label{app:hungarian_miou}

Let $\mathcal{P}=\{P_1,\ldots,P_M\}$ and $\mathcal{G}=\{G_1,\ldots,G_T\}$ denote the predicted and ground-truth regions, respectively, restricted to valid evaluation pixels. We compute the pairwise Intersection-over-Union between each predicted and ground-truth region:
\begin{equation}
\operatorname{IoU}(P_i,G_j)=
\frac{|P_i\cap G_j|}{|P_i\cup G_j|}.
\end{equation}

We construct an $M\times T$ IoU matrix and use the Hungarian algorithm to obtain the optimal one-to-one assignment:
\begin{equation}
\mathcal{M}^{*}=
\arg\max_{\mathcal{M}}
\sum_{(i,j)\in\mathcal{M}}
\operatorname{IoU}(P_i,G_j),
\end{equation}
where $\mathcal{M}$ ranges over valid one-to-one matchings. The image-level score is
\begin{equation}
\mathrm{mIoU}_{\mathrm{H}}=
\frac{1}{\max(M,T)}
\sum_{(i,j)\in\mathcal{M}^{*}}
\operatorname{IoU}(P_i,G_j).
\end{equation}

The normalization by $\max(M,T)$ ensures that unmatched regions contribute zero, penalizing both excessive fragmentation and insufficient region separation. Unlike conventional semantic mIoU, this metric does not aggregate IoU by semantic class and does not require predicted category labels.

\subsection{Segmentation Covering}
\label{app:covering}

Segmentation Covering evaluates how well each ground-truth region is represented by the predicted partition. For each $G_j$, we identify the predicted region with the highest IoU and weight the result by the ground-truth region's area:
\begin{equation}
\mathrm{Covering}=
\frac{
\sum_{j=1}^{T}|G_j|
\max_{1\leq i\leq M}\operatorname{IoU}(G_j,P_i)
}{
\sum_{j=1}^{T}|G_j|
}.
\end{equation}

Unlike Hungarian-matched mIoU, Covering does not impose a one-to-one correspondence between predicted and ground-truth regions. Multiple ground-truth regions may therefore select the same predicted region. Furthermore, Covering does not explicitly penalize differences between the numbers of predicted and ground-truth regions, although such differences can still affect their IoU values.

These distinctions explain why the two metrics can produce different method rankings. Hungarian-matched mIoU directly penalizes unmatched regions, whereas Covering emphasizes the quality of the best available overlap for each ground-truth region. Reporting both metrics provides complementary information about segmentation quality and region granularity.

\subsection{Dataset-Level Aggregation and Instance AP}
\label{app:metric_aggregation}

Both metrics are computed independently for each image and then averaged across the evaluation split, giving each image equal weight. We additionally report the average number of predicted regions. For our method, this count is measured after the applicable small-region cleanup step.

We do not report instance Average Precision (AP), since our evaluation concerns class-agnostic, full-scene partitions rather than confidence-ranked object-instance proposals. The predicted regions may represent objects, object parts, or background structures, and do not naturally correspond to ranked instance detections. Hungarian-matched mIoU and Segmentation Covering are therefore better aligned with the output and evaluation objectives of our segmentation framework.

\section{Dataset and Annotation Preparation}
\label{app:dataset_preparation}

We evaluate our method on COCO Panoptic, Cityscapes, and ADE20K using the validation splits described in the main paper. Table~\ref{tab:supp_dataset_statistics} summarizes the dataset sizes and average numbers of ground-truth segments. All primary experiments use depth maps generated by pretrained Depth Anything V2 (DA-V2), while a separate Cityscapes experiment evaluates the method using sensor-derived stereo disparity. The segmentation algorithm operates exclusively on the supplied depth representation, without accessing RGB features or semantic category labels.

\begin{table}[t]
\centering
\caption{Evaluation datasets and ground-truth segmentation statistics. The average number of segments is computed per image.}
\label{tab:supp_dataset_statistics}
\small
\begin{tabular}{lccc}
\toprule
Dataset & Split & Images & GT regions/image \\
\midrule
COCO Panoptic & val2017 & 5,000 & 11.35 \\
Cityscapes & Fine val & 500 & 29.34 \\
ADE20K & Val & 2,000 & 15.21 \\
\bottomrule
\end{tabular}
\end{table}

\subsection{COCO Panoptic and Cityscapes}
\label{app:coco_cityscapes_preparation}

For COCO Panoptic~\cite{lin2014microsoft,kirillov2019panoptic}, we use the 5,000 images from val2017 and their corresponding panoptic segmentation annotations. Ground-truth segments are treated as class-agnostic regions, without distinguishing semantic categories during evaluation. Void and unlabeled pixels are excluded from the scoring procedure.

For Cityscapes~\cite{cordts2016cityscapes}, we use the 500 images from the fine validation split and their corresponding ground-truth segmentation regions. The primary Cityscapes experiment uses DA-V2 Large depth estimates, whereas COCO uses DA-V2 Small. These depth inputs are processed using the same segmentation framework, with dataset-specific geometric parameters documented in Sec.~\ref{sec:supp_implementation}.

We additionally evaluate the stereo disparity maps distributed with Cityscapes to investigate segmentation without a neural depth estimator. The disparity values are decoded from 16-bit PNG images according to
\begin{equation}
d_{\mathrm{disp}}(x,y)=\frac{p(x,y)-1}{256},
\end{equation}
where $p(x,y)$ denotes the encoded pixel value and $p=0$ indicates an invalid measurement. Approximately 22.1\% of disparity pixels are invalid. These measurements are filled using the nearest valid disparity values before subsequent processing. Since larger disparity generally corresponds to smaller physical depth, the decoded representation follows the near-to-far ordering required by our algorithm.

The stereo-disparity experiments are evaluated separately from the monocular-depth results. In particular, the main Cityscapes H-mIoU of 0.2086 is obtained using DA-V2 Large depth estimates, not stereo disparity. The effects of disparity-specific transformations are analyzed in Sec.~\ref{sec:sensor_depth_ablation}.

\subsection{ADE20K Panoptic Annotation Construction}
\label{app:ade20k_panoptic}

ADE20K~\cite{zhou2017scene} provides semantic segmentation annotations and separate instance-level annotations, but the publicly accessible annotations used in our experiments are not supplied as a directly usable panoptic partition. We therefore construct class-agnostic panoptic ground truth by combining the publicly available 150-class semantic masks with the instance-ID annotations.

The construction follows four rules. First, the 100 instance-level ``thing'' categories are mapped to their corresponding semantic categories using canonical class names. We require exact canonical-name correspondence rather than approximate substring or synonym matching, which can incorrectly associate distinct classes. Second, instance annotations are assigned individual segment identifiers, preserving the distinction between annotated instances. Third, instance annotations take priority wherever their labels disagree with the semantic segmentation mask. Finally, pixels not assigned to an instance are grouped according to their remaining semantic category, with one segment per semantic class per image. Pixels without a valid annotation remain excluded from evaluation.

Importantly, the semantic fallback does not split each category into separate connected components. A single semantic segment can therefore contain spatially disconnected regions. This convention is consistent with panoptic annotations in which a segment identifier need not correspond to one spatially connected component. For reference, an inspection of COCO Panoptic annotations found that approximately 33.7\% of ground-truth segments contain disconnected components. Splitting ADE20K semantic regions by connectivity would instead substantially increase the number of target regions and change the evaluation protocol.

Using this construction, the ADE20K validation set contains an average of 15.21 ground-truth segments per image. As an internal consistency check, evaluating each constructed annotation against itself produces a Hungarian-matched mIoU of 1.0. This confirms consistency of the constructed labels with the scoring implementation, but does not independently establish equivalence to an official panoptic annotation release.

All ADE20K segmentation results use these constructed ground-truth partitions and DA-V2 Small depth estimates. The same annotations and evaluation implementation are used for our method and all compared baselines, ensuring that the reported differences are not caused by using different ground-truth constructions.

\section{Baseline Implementation Details}
\label{app:baseline_implementations}

We compare our approach against classical segmentation algorithms, zero-shot neural segmentation methods, and a fully supervised reference under the unified class-agnostic evaluation protocol described in Sec.~\ref{app:evaluation_protocol}. Because these methods differ in their input modalities, training supervision, and output representations, we document their configurations and the procedures used to obtain comparable image partitions. All reported scores are computed using our evaluation implementation rather than taken from the methods' original publications.

\subsection{Classical Segmentation Baselines}
\label{app:classical_baselines}

We evaluate Felzenszwalb's graph-based segmentation~\cite{felzenszwalb2004efficient} and SLIC~\cite{achanta2012slic} using RGB images and depth-based inputs where available. For depth-based experiments, the algorithms operate on scalar depth representations rather than RGB features. We evaluate multiple parameter configurations and report the configurations achieving the highest Hungarian-matched mIoU.

For Felzenszwalb, we vary the graph-segmentation scale parameter, while for SLIC, we vary the requested number of superpixels. These parameters influence the granularity of the resulting partitions, and the number of predicted regions may differ from the nominal parameter values. The comparison tables report the actual average number of regions produced.

We also examine depth preprocessing separately for the classical baselines. On COCO, the reported depth-based Felzenszwalb configuration uses globally detrended depth, while depth-based SLIC uses the original depth representation, as these achieve the stronger results in their respective parameter sweeps. The corresponding best H-mIoU configurations use scale $s=1000$ for Felzenszwalb and $n=6$ for SLIC. On ADE20K, the reported depth-based configurations use detrended depth for both methods. The Cityscapes SLIC depth-input comparison instead uses the dataset's sensor-derived stereo disparity, whereas our main Cityscapes result uses monocular depth estimated by Depth Anything V2.

The reported classical-baseline configurations are selected according to H-mIoU rather than independently optimizing each evaluation metric. These objectives can favor different segmentation granularities. For example, on COCO, depth-based Felzenszwalb achieves H-mIoU of 0.1516 and Covering of 0.4901 in its H-mIoU-selected configuration, whereas a separately selected configuration reaches Covering of 0.5124 with an average of 29.22 regions. We retain the H-mIoU-selected configuration in the main comparison table for consistency.

\subsection{SAM and SAM2}
\label{app:sam_baselines}

We evaluate SAM~\cite{kirillov2023segment} and SAM2~\cite{ravi2024sam2} using RGB images with point-based prompting configurations of 16 and 32 points. Both models rely on learned visual representations, but do not receive dataset-specific semantic category vocabularies during evaluation.

SAM and SAM2 produce mask proposals that may overlap and need not cover the entire image. To evaluate these outputs as complete partitions, we apply a common mask-assignment procedure. For each pixel covered by multiple proposals, we select the covering mask with the highest confidence score. Ties are resolved in favor of the smaller mask. Pixels not covered by any proposal are assigned to a single residual region, which is included in both the predicted region count and the segmentation evaluation.

The residual region can occupy a substantial fraction of the image. Across the evaluated configurations, approximately 41--46\% of COCO pixels and 61--80\% of Cityscapes pixels remain uncovered by the predicted masks. In particular, SAM2 with 16-point prompting leaves approximately 80\% of Cityscapes pixels in the residual region. Consequently, the resulting segmentation metrics reflect both the predicted masks and the additional residual region. This distinction is especially relevant for Segmentation Covering, where a large residual region may provide substantial overlap with ground-truth background or other extended regions.

We retain the uncovered pixels rather than excluding them from evaluation, ensuring that SAM and SAM2 are scored as full-image partitions under the same protocol as the other approaches.

\subsection{Vocabulary-Conditioned Segmentation}
\label{app:vocabulary_baselines}

We evaluate GroupViT~\cite{xu2022groupvit} and CLIPSeg~\cite{lueddecke2022image} using RGB images and dataset-specific semantic category vocabularies. The vocabularies contain 133 panoptic categories for COCO, 19 evaluation categories for Cityscapes, and 150 semantic categories for ADE20K.

These models receive the corresponding category names as text input, allowing them to exploit semantic information about the categories present in each benchmark. Although no dataset-specific model fine-tuning is performed, this constitutes an information advantage relative to methods that operate without semantic category descriptions, including our depth-based segmentation framework. The category vocabulary is therefore treated as part of the evaluation configuration rather than an unspecified implementation detail.

Predicted semantic category identities are not used when computing our class-agnostic evaluation metrics. Instead, the resulting segmentation partitions are evaluated using the same region-overlap measures applied to the other methods. The reported results consequently measure partition agreement rather than conventional semantic segmentation accuracy.

\subsection{DiffCut and Fully Supervised Reference}
\label{app:other_neural_baselines}

We additionally evaluate DiffCut~\cite{couairon2024diffcut}, which performs segmentation using features extracted from a pretrained diffusion model. Our comparisons use the SSD-1B-based configuration. Unlike GroupViT and CLIPSeg, DiffCut does not receive a dataset-specific category vocabulary during inference. Its segmentation output is evaluated directly under our class-agnostic partition protocol.

For COCO Panoptic, we also include Mask2Former~\cite{cheng2022masked} trained with dataset supervisions. This model serves as a fully supervised reference rather than a zero-shot baseline. Its training setting differs substantially from the other methods, and its results are therefore reported separately.

\subsection{Evaluation Consistency and Comparison Scope}
\label{app:baseline_evaluation}

All baseline predictions are evaluated against the same ground-truth partitions using Hungarian-matched mIoU and Segmentation Covering. We additionally report the average number of predicted regions to expose differences in segmentation granularity. The reported metrics should not be directly equated with semantic mIoU or other performance measures published in the original baseline papers, since their matching procedures, category definitions, and evaluation objectives may differ.

All COCO and ADE20K comparisons use their complete validation splits of 5,000 and 2,000 images, respectively. On Cityscapes, all methods are evaluated on 500 images except DiffCut, for which one image failed during inference because of a recursion-depth error, resulting in 499 evaluated images.

The comparison methods also differ in the information available during inference. Our approach uses estimated monocular depth without RGB features or semantic category labels during segmentation. The classical methods use RGB or depth-based inputs, SAM and SAM2 use learned RGB representations with point prompts, and GroupViT and CLIPSeg additionally receive dataset-specific category vocabularies. DiffCut uses pretrained diffusion features, while the supervised Mask2Former reference benefits from labeled COCO training data. These distinctions should be considered when interpreting the relative performance of the methods.

\section{Extended Baseline Comparison Tables}
\label{app:extended_baselines}

We provide detailed per-dataset comparisons complementing the consolidated results in the main paper. Each table reports the input representation, evaluated configuration, mean number of predicted regions, Hungarian-matched mIoU (H-mIoU), and Segmentation Covering. All numerical results are obtained using the unified class-agnostic evaluation protocol described in Sec.~\ref{app:evaluation_protocol}, rather than copied from the original publications.

For classical segmentation methods, the reported configurations are selected according to H-mIoU from their evaluated parameter sweeps. Neural methods are evaluated under their corresponding prompting, vocabulary, or feature-extraction settings. These distinctions are important because methods differ in their input information, supervision, and resulting region granularity.

\subsection{COCO Panoptic}
\label{app:coco_baselines}

Table~\ref{tab:supp_coco_baselines} presents the complete comparison on COCO Panoptic val2017, comprising 5,000 images. Our method uses depth estimated by DA-V2 Small, followed by global detrending and segmentation with $(k,m,g)=(6,16,0.2\%)$. The target region count was selected using a disjoint train2017 subset, while the remaining geometric parameters were established using validation-set experiments.

\begin{table*}[t]
\centering
\caption{Extended class-agnostic segmentation comparison on COCO Panoptic val2017 (5,000 images). Classical configurations correspond to their H-mIoU-selected settings. Mask2Former uses a COCO-trained Swin-Tiny checkpoint as a fully supervised reference. Bold indicates the highest H-mIoU among non-supervised methods. The fixed-grid control is reported separately from the segmentation methods.}
\label{tab:supp_coco_baselines}
\small
\setlength{\tabcolsep}{4pt}
\begin{tabular}{llllrrr}
\toprule
Method & Input & Configuration & Setting & \#R & H-mIoU & Covering \\
\midrule
DiffCut & RGB & SSD-1B backbone & Zero-shot & 6.31 & \textbf{0.3245} & 0.6668 \\
SAM2 & RGB & 16 points & Zero-shot & 10.11 & 0.2783 & 0.6002 \\
SAM2 & RGB & 32 points & Zero-shot & 14.59 & 0.2648 & 0.6107 \\
CLIPSeg & RGB & 133-category vocabulary & Zero-shot & 13.31 & 0.2471 & 0.6684 \\
\textbf{DORMS} & DA-V2 Small & $k=6,\;m=16,\;g=0.2\%$ & Depth-only & 5.93 & 0.2332 & 0.5119 \\
SLIC & Depth (raw) & $n=6$ & Classical & 5.16 & 0.1712 & 0.3823 \\
SAM & RGB & 16 points & Zero-shot & 27.17 & 0.1608 & 0.5765 \\
SLIC & RGB & $n=8$ & Classical & 3.49 & 0.1537 & 0.3931 \\
Felzenszwalb & Depth (detrended) & $s=1000$ & Classical & 11.10 & 0.1516 & 0.4901 \\
GroupViT & RGB & 133-category vocabulary & Zero-shot & 4.02 & 0.1429 & 0.4099 \\
Felzenszwalb & RGB & $s=3000$ & Classical & 14.04 & 0.1349 & 0.4638 \\
SAM & RGB & 32 points & Zero-shot & 47.41 & 0.1168 & 0.5813 \\
\midrule
\multicolumn{7}{l}{\textit{Fully supervised reference}} \\
Mask2Former & RGB & Swin-Tiny (COCO) & Supervised & 9.93 & 0.5590 & 0.8163 \\
\midrule
\multicolumn{7}{l}{\textit{Fixed-grid control}} \\
Grid control & Depth & 9-region grid & Control & 9.00 & 0.1310 & 0.2765 \\
\bottomrule
\end{tabular}
\end{table*}

The proposed method achieves higher H-mIoU than all evaluated classical baselines while producing relatively few regions. Among the classical methods, depth-based SLIC achieves the highest H-mIoU (0.1712), followed by RGB-based SLIC (0.1537) and detrended-depth Felzenszwalb (0.1516). The results also demonstrate that using depth does not uniformly outperform RGB for every classical algorithm.

The table reports the H-mIoU-selected configuration for each classical baseline. When Covering is optimized independently, depth-based Felzenszwalb achieves a slightly higher COCO Covering score than our method (0.5124 versus 0.5119), but requires 29.22 regions on average, compared with 5.93 for our approach. This illustrates the influence of segmentation granularity on comparisons between the two metrics.

The nine-region grid control provides an additional reference for partition quality without sophisticated geometric grouping. Our method substantially exceeds this control in both H-mIoU and Covering.

\subsection{Cityscapes}
\label{app:cityscapes_baselines}

Table~\ref{tab:supp_cityscapes_baselines} reports the comparison on the Cityscapes fine validation set. Our main configuration uses DA-V2 Large depth estimates with $(k,m,g)=(30,3,10\%)$. The classical depth-based baselines shown here use Cityscapes stereo disparity rather than the monocular depth input used by our headline configuration. Therefore, the comparisons involve different depth representations as well as different segmentation procedures.

\begin{table*}[t]
\centering
\caption{Extended class-agnostic segmentation comparison on Cityscapes fine validation. All methods use 500 images except DiffCut, which is evaluated on 499 images following one inference failure. The depth-based classical rows use sensor-derived stereo disparity, whereas DORMS uses DA-V2 Large monocular depth. Mask2Former uses a Cityscapes-trained Swin-Tiny checkpoint as a fully supervised reference. Bold indicates the highest H-mIoU among non-supervised methods.}
\label{tab:supp_cityscapes_baselines}
\small
\setlength{\tabcolsep}{4pt}
\begin{tabular}{llllrrr}
\toprule
Method & Input & Configuration & Setting & \#R & H-mIoU & Covering \\
\midrule
\textbf{DORMS} & DA-V2 Large & $k=30,\;m=3,\;g=10\%$ & Depth-only & 25.84 & \textbf{0.2086} & 0.5866 \\
CLIPSeg & RGB & 19-category vocabulary & Zero-shot & 9.24 & 0.1506 & 0.6713 \\
SAM2 & RGB & 32 points & Zero-shot & 41.35 & 0.1464 & 0.4356 \\
SAM2 & RGB & 16 points & Zero-shot & 22.77 & 0.1365 & 0.4063 \\
DiffCut & RGB & SSD-1B backbone & Zero-shot & 5.54 & 0.1309 & 0.6834 \\
SAM & RGB & 16 points & Zero-shot & 26.53 & 0.1234 & 0.4536 \\
SAM & RGB & 32 points & Zero-shot & 52.98 & 0.1050 & 0.4766 \\
Felzenszwalb & RGB & $s=1000$ & Classical & 35.21 & 0.0917 & 0.4562 \\
SLIC & Stereo disparity & $n=25$ & Classical & 25.91 & 0.0735 & 0.1871 \\
SLIC & RGB & $n=25$ & Classical & 20.09 & 0.0732 & 0.2350 \\
Felzenszwalb & Stereo disparity & $s=30$ & Classical & 70.16 & 0.0522 & 0.4598 \\
GroupViT & RGB & 19-category vocabulary & Zero-shot & 2.21 & 0.0303 & 0.3032 \\
\midrule
\multicolumn{7}{l}{\textit{Fully supervised reference}} \\
Mask2Former & RGB & Swin-Tiny (Cityscapes) & Supervised & 20.20 & 0.4278 & 0.8788 \\
\bottomrule
\end{tabular}
\end{table*}

Our method obtains the highest H-mIoU among the evaluated approaches, exceeding CLIPSeg (0.1506), SAM2 with 32 points (0.1464), and DiffCut (0.1309). Its Covering score of 0.5866 also exceeds the classical baselines and both SAM2 configurations, although CLIPSeg and DiffCut achieve higher Covering.

The classical comparisons demonstrate that using sensor disparity directly does not necessarily produce strong segmentation performance. As investigated in Sec.~\ref{sec:sensor_depth_ablation}, our method also supports a separate model-free configuration operating on the same sensor-derived disparity. With appropriate preprocessing, that configuration reaches an H-mIoU of 0.1163 and a maximum observed Covering of 0.4994. These results are analyzed separately from the main DA-V2-based comparison because they use a different depth representation and segmentation configuration.

\subsection{ADE20K}
\label{app:ade20k_baselines}

Table~\ref{tab:supp_ade20k_baselines} presents the comparison on the 2,000-image ADE20K validation set using the constructed panoptic annotations described in Sec.~\ref{app:ade20k_panoptic}. Our method uses DA-V2 Small depth maps with $(k,m,g)=(8,8,2\%)$. Unlike COCO, this configuration was selected using parameter sweeps on the evaluation split.

\begin{table*}[t]
\centering
\caption{Extended class-agnostic segmentation comparison on ADE20K validation (2,000 images). All methods are evaluated using the same constructed panoptic annotations. Classical configurations correspond to their H-mIoU-selected settings. Mask2Former uses an ADE20K-trained Swin-Large panoptic checkpoint as a fully supervised reference; its backbone is larger than the Swin-Tiny checkpoint used for COCO and Cityscapes. Bold indicates the highest H-mIoU among non-supervised methods.}
\label{tab:supp_ade20k_baselines}
\small
\setlength{\tabcolsep}{4pt}
\begin{tabular}{llllrrr}
\toprule
Method & Input & Configuration & Setting & \#R & H-mIoU & Covering \\
\midrule
DiffCut & RGB & SSD-1B backbone & Zero-shot & 6.20 & \textbf{0.2841} & 0.6329 \\
SAM2 & RGB & 32 points & Zero-shot & 11.92 & 0.2696 & 0.5474 \\
SAM2 & RGB & 16 points & Zero-shot & 8.66 & 0.2615 & 0.5357 \\
\textbf{DORMS} & DA-V2 Small & $k=8,\;m=8,\;g=2\%$ & Depth-only & 7.89 & 0.2206 & 0.5126 \\
CLIPSeg & RGB & 150-category vocabulary & Zero-shot & 17.25 & 0.2126 & 0.6488 \\
SAM & RGB & 16 points & Zero-shot & 26.95 & 0.1946 & 0.5558 \\
SLIC & RGB & $n=12$ & Classical & 6.77 & 0.1528 & 0.3684 \\
SAM & RGB & 32 points & Zero-shot & 47.43 & 0.1472 & 0.5635 \\
Felzenszwalb & RGB & $s=3000$ & Classical & 14.11 & 0.1413 & 0.4425 \\
SLIC & Depth (detrended) & $n=6$ & Classical & 4.75 & 0.1366 & 0.3460 \\
Felzenszwalb & Depth (detrended) & $s=1000$ & Classical & 10.55 & 0.1336 & 0.4348 \\
GroupViT & RGB & 150-category vocabulary & Zero-shot & 3.99 & 0.1221 & 0.3829 \\
\midrule
\multicolumn{7}{l}{\textit{Fully supervised reference}} \\
Mask2Former & RGB & Swin-Large (ADE20K) & Supervised & 12.09 & 0.5565 & 0.8310 \\
\bottomrule
\end{tabular}
\end{table*}

Our method obtains an H-mIoU of 0.2206, outperforming CLIPSeg, SAM, GroupViT, and all evaluated classical baselines. DiffCut and both SAM2 configurations achieve higher H-mIoU. The Covering results show a different ranking, with CLIPSeg, DiffCut, and several other neural approaches exceeding our method.

Among classical methods, RGB-based SLIC achieves the highest H-mIoU (0.1528), followed by RGB-based Felzenszwalb (0.1413). Both remain substantially below our method in H-mIoU and Covering. The results demonstrate that the advantage of the proposed geometric merging framework is not simply attributable to using depth rather than RGB, since directly applying classical segmentation methods to depth maps produces weaker results.

\subsection{Interpretation of Baseline Comparisons}
\label{app:baseline_comparison_discussion}

The comparisons should be interpreted in light of the differing information available to each approach. Our segmentation algorithm receives depth values without semantic labels, whereas GroupViT and CLIPSeg additionally receive dataset-specific category vocabularies. SAM and SAM2 use pretrained RGB representations and point-based prompting, while DiffCut relies on diffusion-model features. Mask2Former is included exclusively as a fully supervised COCO reference.

The reported segmentation metrics also depend on the number and structure of predicted regions. In particular, Hungarian-matched mIoU penalizes unmatched regions through one-to-one assignment, whereas Segmentation Covering allows multiple ground-truth regions to select the same predicted region. Accordingly, independently optimizing these metrics can produce different preferred segmentation granularities.

The purpose of these comparisons is therefore to evaluate class-agnostic partition quality under a common scoring protocol, rather than to claim equivalence in supervision, input information, or computational requirements.

\section{Additional Depth Detrending Analysis}
\label{app:additional_detrending}

We provide additional analysis of global depth detrending, focusing on its relationship with scene planarity and its effect on classical depth-based segmentation methods. These experiments complement the main-paper results in Sec.~\ref{sec:detrending_results} and investigate whether the observed improvements arise primarily from removing dominant planar structures or from a more general interaction between depth detrending and geometric region merging.

\subsection{Performance Across Planarity Quartiles}
\label{app:planarity_analysis}

To examine the relationship between detrending and scene geometry, we divide the 5,000 COCO Panoptic validation images into four equally sized quartiles according to the proportion of depth variation explained by a fitted linear plane. For each image, we compute the coefficient of determination
\begin{equation}
R^2 = 1 -
\frac{\sum_{(x,y)\in\Omega}[d(x,y)-\hat d(x,y)]^2}
{\sum_{(x,y)\in\Omega}[d(x,y)-\bar d]^2},
\end{equation}
where $\hat d(x,y)=\hat a x+\hat b y+\hat c$ is the least-squares plane fitted to the depth map and $\bar d$ is its mean depth value. Higher $R^2$ indicates that a larger fraction of the depth variation can be explained by a global linear spatial trend. This measure characterizes the structure of the estimated depth map rather than directly identifying physical planar surfaces.

We sort images by $R^2$ and divide them into quartiles Q1--Q4, containing 1,250 images each. Q1 corresponds to the weakest global linear trends, while Q4 contains images whose depth maps are most strongly explained by a linear plane. For every image, we compare segmentation with and without detrending using identical segmentation parameters. The resulting per-image differences provide a paired evaluation of the preprocessing step.

\begin{table*}[t]
\centering
\caption{Effect of global depth detrending across COCO Panoptic planarity quartiles. Each quartile contains 1,250 images. Improvements are calculated from paired image-level H-mIoU differences, with 95\% confidence intervals obtained using 10,000 paired bootstrap resamples. Statistical significance is assessed using two-sided paired $t$-tests. Displayed values are rounded independently.}
\label{tab:supp_planarity_quartiles}
\small
\setlength{\tabcolsep}{5pt}
\begin{tabular}{lccccccc}
\toprule
Quartile & $R^2$ range & Without & With & $\Delta$ H-mIoU & 95\% CI & $p$-value & Improved images \\
\midrule
Q1 & [0.0003, 0.3640] & 0.2127 & 0.2284 & +0.0158 & [0.0117, 0.0198] & $1.08\times10^{-13}$ & 761 / 1250 \\
Q2 & [0.3641, 0.5883] & 0.1954 & 0.2341 & +0.0387 & [0.0344, 0.0431] & $6.35\times10^{-61}$ & 915 / 1250 \\
Q3 & [0.5884, 0.7538] & 0.1974 & 0.2428 & \textbf{+0.0454} & [0.0408, 0.0502] & $2.96\times10^{-70}$ & 929 / 1250 \\
Q4 & [0.7539, 0.9939] & 0.2092 & 0.2274 & +0.0182 & [0.0123, 0.0241] & $1.26\times10^{-9}$ & 802 / 1250 \\
\bottomrule
\end{tabular}
\end{table*}

As shown in Table~\ref{tab:supp_planarity_quartiles}, detrending improves average H-mIoU in all four quartiles. The improvements are statistically significant in every group, with all 95\% confidence intervals excluding zero. The largest gains occur in Q2 (+0.0387) and Q3 (+0.0454), whereas the improvements are smaller in Q1 (+0.0158) and Q4 (+0.0182). The mean improvement across all 5,000 images is +0.0295, consistent with the aggregate COCO result reported in the main paper. Figure~\ref{fig:planarity_quartiles} visualizes these quartile-level improvements and their confidence intervals.

These findings indicate that the benefit of detrending is not concentrated in images with the strongest global linear depth structure. Instead, the largest improvements occur at intermediate levels of planarity. One possible explanation is that removing a global depth trend makes local variations more influential during component formation and persistence-based merging. In images with weak global trends, there may be relatively little structure to remove, while images dominated by a strong global gradient may contain less residual geometric variation after subtraction. Although consistent with the observed pattern, this interpretation does not establish a unique causal mechanism.

The quartile-level averages also conceal substantial variation across individual images. Detrending improves H-mIoU for 761, 915, 929, and 802 images in Q1--Q4, respectively, while reducing performance for some remaining images. Thus, detrending provides a consistent average benefit without guaranteeing an improvement for every scene.

\subsection{Effect on Classical Depth-Based Segmentation}
\label{app:classical_detrending}

We additionally evaluate whether global depth detrending improves conventional segmentation algorithms applied directly to depth maps. Specifically, we compare Felzenszwalb's graph-based segmentation and SLIC on COCO Panoptic using depth representations with and without plane subtraction. Both methods use the configurations and parameter-selection procedures described in Sec.~\ref{app:classical_baselines}. The results are summarized in Table~\ref{tab:supp_classical_detrending}.

\begin{table}[t]
\centering
\caption{Effect of depth detrending on classical segmentation baselines on COCO Panoptic val2017. Each input variant uses its H-mIoU-selected configuration. Differences are computed from the displayed rounded scores.}
\label{tab:supp_classical_detrending}
\small
\setlength{\tabcolsep}{4pt}
\begin{tabular}{lccc}
\toprule
Method & Without & With & $\Delta$ \\
\midrule
Felzenszwalb (depth) & 0.1496 & 0.1516 & +0.0020 \\
SLIC (depth) & 0.1712 & 0.1626 & -0.0086 \\
\bottomrule
\end{tabular}
\end{table}

Detrending has only a marginal positive effect on Felzenszwalb, increasing H-mIoU from 0.1496 to 0.1516. In contrast, SLIC decreases from 0.1712 to 0.1626 after detrending. These results differ considerably from the +0.0295 improvement observed with our depth-ordered merge-tree framework on the same dataset.

The contrasting behavior can be understood in terms of how the algorithms use depth information. Felzenszwalb relies on local graph-edge differences and component-based merging, while SLIC groups pixels according to spatial and feature similarity. Removing a global depth gradient changes the feature distances underlying these decisions and does not necessarily produce more favorable partitions. In our framework, depth values additionally determine pixel activation order and component persistence, making the interaction between detrending and subsequent merging fundamentally different.

These comparisons demonstrate that global depth detrending is not a universally beneficial preprocessing step for depth-based segmentation. Its contribution depends on the segmentation mechanism, supporting its use as an integrated component of the proposed framework rather than as a generic enhancement applied indiscriminately to existing algorithms.

\section{Extended Cityscapes Stereo-Disparity Analysis}
\label{app:stereo_disparity}

We provide additional analysis of the Cityscapes stereo-disparity experiments presented in Sec.~\ref{sec:sensor_depth_ablation}. Unlike the main experiments using pretrained monocular depth estimation, these experiments operate directly on sensor-derived disparity maps, allowing the complete segmentation pipeline to run without a neural network. We investigate how disparity encoding, numerical value spacing, and preprocessing affect depth-ordered geometric merging.

\subsection{Disparity Representation and Its Effects}
\label{app:disparity_representation}

Cityscapes provides stereo disparity maps encoded as 16-bit images. Valid disparity values are recovered using $d_{\mathrm{disp}}=(p-1)/256$, where $p=0$ indicates an invalid measurement. Approximately 22.1\% of pixels contain invalid measurements, which are initially replaced using nearest-valid interpolation.

Two properties of this representation are particularly relevant to our method. First, disparity is quantized, producing many pixels with identical values. Since our algorithm processes pixels in descending depth order and resolves equal-depth ties by pixel index, large groups of identical values can affect component activation and merging. For example, in the Cityscapes image \texttt{frankfurt\_000001\_048355}, the disparity map contains only 25,344 distinct values across approximately 2.1 million pixels, compared with 2,002,886 distinct values in the corresponding Depth Anything V2 estimate. Approximately 23.0\% of disparity pixels belong to tie groups containing more than 1,000 pixels. These statistics characterize this particular image rather than the entire dataset.

Second, stereo disparity is inversely related to physical depth, causing numerical differences between distant surfaces to become compressed. This is particularly important because our area--persistence merge score depends on differences between component birth values and the depth of subsequent bridging pixels. Consequently, two depth representations with the same pixel ordering can produce different merge decisions when their numerical spacing differs.

Global linear detrending addresses part of the first issue by introducing position-dependent residual values, which can break large groups of equal disparity values. However, detrending does not necessarily correct the nonlinear compression of disparity values. We therefore investigate additional transformations that modify value spacing while preserving, where possible, the underlying depth ordering.

\subsection{Preprocessing Variants and Experimental Protocol}
\label{app:disparity_preprocessing}

We examine several preprocessing operations applied before the standard global detrending step. The evaluated operations include alternative invalid-value filling strategies, logarithmic and inverse-depth transformations, spatial smoothing, rank-based transformation, and resolution changes. The operations are implemented in \texttt{eval/disparity\_prep.py}, with filling performed before smoothing to avoid filtering directly across pixels representing missing measurements.

A particularly informative control is the logarithmic transformation. When applied to the filled disparity values before plane subtraction, this strictly increasing transformation preserves their ordering while changing their numerical spacing. We verified that the pixel activation order remains identical to that of the original disparity representation. Thus, differences observed before detrending can be attributed to changes in value spacing rather than changes in sweep order. Once global detrending is applied, however, identical ordering is no longer guaranteed.

We first screen 12 preprocessing variants on 20 Cityscapes images at $k=30$. Table~\ref{tab:supp_disparity_screen} summarizes the results. These experiments serve only to identify promising configurations for further evaluation and are not treated as full-validation performance estimates.

\begin{table}[t]
\centering
\caption{Initial screening of disparity preprocessing variants on 20 Cityscapes images at $k=30$. Results are preliminary and are used only to select configurations for subsequent experiments.}
\label{tab:supp_disparity_screen}
\small
\setlength{\tabcolsep}{5pt}
\begin{tabular}{lcc}
\toprule
Preprocessing variant & H-mIoU & Covering \\
\midrule
Fill split & \textbf{0.1184} & 0.4287 \\
Log transform & 0.1170 & \textbf{0.5100} \\
Fill far & 0.1152 & 0.3875 \\
Median ($3\times3$) & 0.1143 & 0.4107 \\
Bilateral filtering & 0.1142 & 0.3957 \\
Half resolution & 0.1131 & 0.4007 \\
Baseline & 0.1130 & 0.4033 \\
Median ($5\times5$) & 0.1123 & 0.4078 \\
Feather & 0.1096 & 0.4082 \\
Rank transform & 0.1094 & 0.4128 \\
Inverse depth 1 & 0.1036 & 0.4678 \\
Inverse depth 2 & 0.1011 & 0.4483 \\
\bottomrule
\end{tabular}
\end{table}

The logarithmic transformation produces the largest improvement in Covering, increasing it from 0.4033 to 0.5100 on the screening subset. Since this transformation preserves the original disparity ordering before detrending, the result indicates that numerical spacing, rather than pixel ordering alone, influences the segmentation. In contrast, the rank transformation, which preserves ordering while discarding the original spacing, produces lower H-mIoU than the baseline. This further indicates that the numerical relationships between depth values contain information relevant to our merge criterion.

The H-mIoU differences among the best individual preprocessing variants are relatively small on this 20-image subset, and no confidence intervals were computed. We therefore do not interpret these preliminary rankings as statistically established differences.

Following the screening experiments, we evaluate combinations of promising preprocessing operations across $k\in\{15,20,25,30,35,40\}$ on the same 20 images. The combination \texttt{fill\_split+log+median3+half} achieves the highest observed H-mIoU in this stage, reaching 0.1284 at $k=20$. We select this configuration for full-validation evaluation. This selection reflects the H-mIoU objective rather than independent optimization for Covering.

Throughout the full-validation disparity study, the segmentation algorithm itself remains unchanged. The experiments use $m=16$, $g=10\%$, a minimum-region fraction of $0.0005$, and the preferred-over-target region-count selection policy. Unlike the main monocular-depth experiments, the disparity study uses a fixed search upper bound $\tau_{\max}=10^9$, a maximum of 500 sweep replays, and a region-count cap of $\max(24,3k)$ computed separately for each target.

\subsection{Full-Validation Region-Count Sensitivity}
\label{app:disparity_k_sweep}

We evaluate the selected preprocessing configuration on all 500 Cityscapes validation images over nine target region counts. As a control, we also evaluate the original filled disparity representation with global plane subtraction but without the additional disparity-specific preprocessing. The control is evaluated over five target counts using the same disparity-study search settings.

Table~\ref{tab:supp_disparity_k_sweep} reports the complete full-validation results. H-mIoU, Covering, and the mean final region count are computed over the 500 images for each configuration.

\begin{table*}[t]
\centering
\caption{Full-validation Cityscapes stereo-disparity results across target region counts. The proposed disparity preprocessing uses \texttt{fill\_split+log+median3+half} followed by global detrending. The control uses nearest-filled disparity with global detrending only. Region counts are measured after cleanup. A dash indicates a target count not evaluated for the control.}
\label{tab:supp_disparity_k_sweep}
\small
\setlength{\tabcolsep}{6pt}
\begin{tabular}{c|ccc|ccc}
\toprule
& \multicolumn{3}{c|}{Disparity preprocessing + detrending}
& \multicolumn{3}{c}{Detrending only} \\
\cmidrule(lr){2-4}
\cmidrule(lr){5-7}
$k$ & H-mIoU & Covering & \#R & H-mIoU & Covering & \#R \\
\midrule
4  & 0.0688 & 0.4708 & 4.0  & -- & -- & -- \\
5  & 0.0772 & 0.4887 & 4.9  & -- & -- & -- \\
7  & 0.0892 & \textbf{0.4994} & 6.9 & 0.0858 & \textbf{0.4751} & 7.0 \\
10 & 0.1002 & 0.4969 & 9.9  & -- & -- & -- \\
15 & 0.1112 & 0.4913 & 14.4 & 0.1014 & 0.4343 & 14.6 \\
20 & 0.1159 & 0.4847 & 18.7 & \textbf{0.1045} & 0.4152 & 19.4 \\
25 & \textbf{0.1163} & 0.4790 & 22.5 & 0.1044 & 0.3968 & 24.0 \\
30 & 0.1142 & 0.4733 & 25.9 & 0.1010 & 0.3827 & 28.6 \\
35 & 0.1111 & 0.4684 & 29.2 & -- & -- & -- \\
\bottomrule
\end{tabular}
\end{table*}

The selected disparity preprocessing achieves its highest H-mIoU of 0.1163 at $k=25$, with lower values at both $k=20$ and $k=30$. Its highest Covering of 0.4994 occurs at $k=7$, also bracketed by evaluated target counts with lower Covering. These results demonstrate that the two metrics favor different segmentation granularities.

For the detrending-only control, the highest observed H-mIoU is 0.1045 at $k=20$, with lower values at $k=15$ and $k=25$. However, the highest observed Covering of 0.4751 occurs at $k=7$, the lowest target count evaluated for this control. Therefore, the control's Covering maximum is not fully localized, and a higher score at a smaller target count cannot be excluded.

Comparisons at matched target counts further illustrate the benefit of the additional disparity preprocessing. At $k=20$, H-mIoU increases from 0.1045 to 0.1159, while at $k=25$ it increases from 0.1044 to 0.1163. These comparisons use the same target count and disparity-study search settings, providing more direct evidence for the additional preprocessing than comparisons between independently selected maxima.

\subsection{Performance Attribution and Interpretation}
\label{app:disparity_attribution}

The original Cityscapes disparity experiment without plane subtraction achieved H-mIoU of 0.0612 and Covering of 0.2963. Introducing global detrending raises the best observed scores to 0.1045 and 0.4751, respectively, while the additional disparity-specific preprocessing raises them further to 0.1163 and 0.4994.

The largest numerical change is associated with the introduction of global detrending. However, the original no-detrending experiment and the later detrended experiments used different threshold-search configurations. Therefore, the difference between their reported maxima should not be interpreted as a strictly controlled estimate of the isolated effect of plane subtraction. Comparisons at matched segmentation parameters support the same direction of improvement, but the search-setting difference remains.

The additional preprocessing contributes a further 0.0118 H-mIoU when comparing the independently selected maxima of the detrending-only and fully preprocessed configurations. The corresponding observed Covering difference is 0.0243, although this value is an upper bound on the difference between the true optimized maxima because the detrending-only control was not evaluated below $k=7$. No paired confidence intervals were computed for these full-validation preprocessing comparisons, so we do not claim statistical significance for the additional improvement.

The corrected sensor-derived disparity configuration exceeds the evaluated classical segmentation baselines on Cityscapes in both H-mIoU and Covering, even when their respective metrics are optimized independently. Nevertheless, it remains below the main DA-V2 Large configuration, which achieves H-mIoU of 0.2086 and Covering of 0.5866.

These findings illustrate that our framework can operate using geometric information from a stereo sensor without a learned depth estimator. They also demonstrate sensitivity to the numerical representation of depth information: preserving pixel ordering alone is insufficient to guarantee similar segmentation behavior when depth-value spacing changes. The experiments do not establish that newer or higher-resolution depth sensors will necessarily improve performance, since the interaction between sensor characteristics and area--persistence-based merging requires further investigation.

\subsection{Experimental Scope and Limitations}
\label{app:disparity_limitations}

The preprocessing variants were initially screened and combined using a subset of 20 images, without confidence intervals. Their rankings should therefore be interpreted as configuration-selection evidence rather than general conclusions about the relative effectiveness of every operation.

The full-validation results use 500 images, but the selected preprocessing configuration and target region counts were determined using evaluation data rather than an independent held-out validation split. The resulting scores should be interpreted as validation-selected performance. In addition, the disparity tie statistics presented above were measured on one representative image and should not be interpreted as dataset-wide estimates.

Finally, the two reported metrics are optimized over potentially different target region counts, and the detrending-only Covering sweep does not bracket its maximum. We retain these distinctions when reporting the results and avoid attributing the full improvement to any single preprocessing operation.

\section{Computational Efficiency and Resource Usage}
\label{app:computational_efficiency}

We evaluate the computational cost of our method and the segmentation baselines in terms of inference latency, memory consumption, and learned parameter counts. In addition to the memory comparison reported in Table~\ref{tab:memory_comparison}, we provide detailed measurements for different hardware configurations and image resolutions. All benchmarks were conducted on the same machine equipped with an AMD Ryzen 9 7950X CPU (16 physical cores) and an NVIDIA RTX 4090 GPU. The measurements characterize single-image inference rather than batched throughput.

\subsection{Benchmark Protocol}
\label{app:efficiency_protocol}

For latency measurements, models are loaded before timing and two warm-up passes are discarded. We report the median runtime across repeated executions of each image, followed by the median across evaluated images. All GPU timings use explicit CUDA synchronization on the device hosting the model. CPU benchmarks allow 16 threads, although our depth-ordered sweep executes sequentially within each image. Image loading, evaluation metrics, and label-map conversion are excluded from timing.

The GPU benchmarks use the first eight COCO Panoptic validation images, with three repetitions per image. CPU benchmarks use the same ordered image set, although computationally expensive models are evaluated on fewer images. The precise sample counts are reported in Table~\ref{tab:supp_runtime}.

Memory is measured independently using a dedicated process that imports the required libraries, loads one model, processes one image, and exits. Peak CPU memory is measured using the process resident set size (RSS). For GPU inference, we separately record peak allocated tensor memory, peak memory reserved by the CUDA caching allocator, and GPU-process memory reported by \texttt{nvidia-smi} after inference. The last quantity additionally includes CUDA context overhead. CPU RSS and GPU VRAM represent different resources and should not be directly equated.

\subsection{Inference Latency}
\label{app:efficiency_latency}

Table~\ref{tab:supp_runtime} compares CPU and GPU inference times. Our segmentation stage requires no GPU and operates directly on the supplied depth map. With the per-image threshold search used for the reported segmentation results, its median CPU latency is 1.1230\,s. A single sweep at a prescribed threshold takes 0.0328\,s, approximately 34 times faster. Thus, threshold selection accounts for approximately 97\% of the segmentation stage's measured runtime.

\begin{table*}[t]
\centering
\caption{Single-image inference latency on COCO Panoptic, in seconds. CPU and GPU measurements use the same hardware but separate execution configurations. GPU measurements use eight images and three repetitions per image; the final column specifies the CPU measurement sample count and repetitions. Dashes indicate unavailable or unmeasured configurations. The fixed-$\tau$ variant does not perform per-image region-count targeting. A dash indicates that the corresponding device configuration
was not benchmarked; DiffCut was evaluated on GPU only.}
\label{tab:supp_runtime}
\small
\setlength{\tabcolsep}{7pt}
\begin{tabular}{lrrc}
\toprule
Method & CPU (s) & GPU (s) & CPU images $\times$ repeats \\
\midrule
\multicolumn{4}{l}{\textit{Proposed segmentation stage}} \\
Ours (full $\tau$ search) & 1.1230 & -- & $8\times3$ \\
Ours (fixed $\tau$) & 0.0328 & -- & $8\times3$ \\
\midrule
\multicolumn{4}{l}{\textit{Classical methods}} \\
Felzenszwalb (depth) & 0.2719 & -- & $8\times3$ \\
SLIC (depth) & 0.0330 & -- & $8\times3$ \\
\midrule
\multicolumn{4}{l}{\textit{Depth estimators}} \\
Depth Anything V2 Small & 0.1333 & 0.0069 & $4\times2$ \\
Depth Anything V2 Large & 1.1161 & 0.0326 & $4\times2$ \\
\midrule
\multicolumn{4}{l}{\textit{Neural segmentation methods}} \\
GroupViT & 0.0850 & 0.0123 & $4\times2$ \\
Mask2Former & 0.2650 & 0.0210 & $4\times2$ \\
SAM (16 points) & 6.9720 & 0.2621 & $3\times1$ \\
SAM (32 points) & 22.8705 & 0.8382 & $3\times1$ \\
SAM2 (16 points) & 7.7445 & 0.2909 & $3\times1$ \\
SAM2 (32 points) & 24.2669 & 0.8868 & $3\times1$ \\
CLIPSeg & 16.8332 & 0.5448 & $3\times1$ \\
DiffCut & -- & 0.8827 & -- \\
\bottomrule
\end{tabular}
\end{table*}

The fixed-threshold result characterizes the cost of the underlying geometric segmentation operation, but does not represent the configuration used to obtain our reported accuracy results. In particular, a prescribed threshold does not ensure that the resulting partition approaches the specified target region count. The additional computational cost of adaptive threshold selection is therefore a tradeoff associated with controlling segmentation granularity.

The CPU latency of Depth Anything V2 Small is 0.1333\,s, compared with 1.1230\,s for our segmentation stage with threshold search. Adding these separately measured medians gives an estimated CPU pipeline latency of 1.2563\,s; this value is not an independently timed end-to-end measurement. Several neural models, including GroupViT and Mask2Former, achieve lower CPU latency than our full-search segmentation stage. Consequently, the primary computational advantage of our method is its parameter-free, accelerator-independent segmentation operation rather than uniformly faster inference.

For SAM and SAM2, increasing the point count from 16 to 32 substantially increases runtime. The GPU latency of SAM rises from 0.2621\,s to 0.8382\,s, while SAM2 increases from 0.2909\,s to 0.8868\,s. The point count changes the number of processed batches rather than the maximum working batch size, explaining why the corresponding GPU memory measurements remain unchanged.

\subsection{Memory Usage Across Image Resolutions}
\label{app:efficiency_memory_resolution}

Table~\ref{tab:supp_memory_resolution} compares the peak CPU memory of our segmentation stage and the classical baselines across three image resolutions. Our method requires 231\,MB on COCO, 233\,MB on ADE20K, and 465\,MB on Cityscapes. These figures include the interpreter, imported libraries, and algorithmic working memory.

\begin{table}[t]
\centering
\caption{Peak CPU process memory (MB) across image resolutions. The measurements use one image from each dataset in a dedicated process. The complete pipeline includes depth estimation and segmentation in the same CPU process.}
\label{tab:supp_memory_resolution}
\small
\setlength{\tabcolsep}{4pt}
\begin{tabular}{lrrr}
\toprule
Method & COCO & ADE20K & Cityscapes \\
\midrule
Ours (segmentation) & \textbf{231} & \textbf{233} & \textbf{465} \\
Ours (complete pipeline) & 1128 & 1311 & 3210 \\
Felzenszwalb & 231 & 257 & 811 \\
SLIC & 157 & 162 & 321 \\
\bottomrule
\end{tabular}
\end{table}

The evaluated image resolutions are $426\times640$ for COCO, $512\times683$ for ADE20K, and $1024\times2048$ for Cityscapes. At Cityscapes resolution, our segmentation stage uses substantially less memory than Felzenszwalb (465\,MB versus 811\,MB), although SLIC requires only 321\,MB. The observed increase in memory with resolution is consistent with the pixel-dependent storage requirements of the geometric segmentation algorithm.

The complete CPU pipeline additionally includes the pretrained depth estimator. Its measured peak memory reaches 1,128\,MB on COCO and 3,210\,MB on Cityscapes, considerably above the segmentation-only measurements. This difference highlights the importance of distinguishing the proposed parameter-free segmentation procedure from the upstream learned depth representation.

The segmentation-only memory measurements are obtained without importing the PyTorch runtime. In contrast, the complete pipeline requires PyTorch for depth estimation, increasing the process memory footprint even before model weights are loaded. The fixed-threshold memory measurements are omitted because the benchmarking procedure first obtains a threshold through a full search in the same process. Since peak RSS is a process-lifetime high-water mark, these measurements do not isolate the memory required by a standalone fixed-threshold replay.

\subsection{GPU Memory and Learned Parameters}
\label{app:efficiency_gpu_memory}

Table~\ref{tab:supp_gpu_memory} provides the GPU memory breakdown and learned parameter counts for the depth estimators and neural segmentation baselines. GPU-process memory includes reserved CUDA memory and context overhead, whereas allocated memory accounts only for active tensor allocations. The distinction is particularly relevant for SAM and SAM2, whose reserved memory substantially exceeds their allocated tensor memory.

\begin{table*}[t]
\centering
\caption{Detailed resource usage of neural models during GPU inference on COCO. Parameter counts are in millions (M), and all memory measurements are in MB. Alloc. and Reserv. denote peak CUDA-allocated and peak CUDA-reserved memory. Process VRAM is the GPU-process memory reported after inference, including the CUDA context. Host RSS is peak CPU process memory during GPU execution and is not equivalent to the CPU-only inference measurements in Table~\ref{tab:memory_comparison}.}
\label{tab:supp_gpu_memory}
\small
\setlength{\tabcolsep}{6pt}
\begin{tabular}{lrrrrr}
\toprule
Method & Params (M) & Alloc. & Reserv. & Process VRAM & Host RSS \\
\midrule
Depth Anything V2 Small & 24.8 & 219 & 272 & 732 & 1407 \\
Depth Anything V2 Large & 335.3 & 1500 & 1758 & 2218 & 2246 \\
\midrule
GroupViT & 55.7 & 278 & 326 & 786 & 1693 \\
Mask2Former & 47.4 & 375 & 510 & 972 & 1773 \\
SAM (16 points) & 93.7 & 2426 & 3880 & 4342 & 1698 \\
SAM (32 points) & 93.7 & 2426 & 3880 & 4342 & 1708 \\
SAM2 (16 points) & 216.9 & 3743 & 5784 & 6246 & 1798 \\
SAM2 (32 points) & 216.9 & 3743 & 5784 & 6246 & 1799 \\
CLIPSeg & 150.7 & 5686 & 5936 & 6396 & 1674 \\
DiffCut & 1415.0 & 15502 & 16878 & 17360 & 8636 \\
\bottomrule
\end{tabular}
\end{table*}

Our segmentation stage contains no learned parameters. The full monocular-depth pipeline nevertheless requires the pretrained Depth Anything V2 estimator, with 24.8 million parameters for the Small model used on COCO and ADE20K and 335.3 million for the Large model used on Cityscapes. Thus, the absence of learned parameters applies specifically to segmentation, not to the complete monocular-depth pipeline.

Among the neural baselines, the GPU-process memory ranges from 786\,MB for GroupViT to 17,360\,MB for DiffCut. SAM and SAM2 require 4,342\,MB and 6,246\,MB, respectively, under both point configurations. Their identical GPU memory measurements are consistent with the fixed processing batch size used by the automatic mask-generation pipeline.

DiffCut also exhibits substantial host memory consumption during GPU inference, reaching 8,636\,MB of peak CPU RSS. Unlike the CPU-only baseline measurements in Table~\ref{tab:memory_comparison}, this value includes model preparation and host-side operations associated with GPU execution. We report it separately rather than interpreting it as a CPU-only inference requirement.

\subsection{Measurement Limitations}
\label{app:efficiency_limitations}

These experiments characterize computational requirements on a single hardware configuration and should not be interpreted as hardware-independent rankings. The latency measurements use a limited number of images, with fewer repetitions for computationally expensive CPU models. Memory measurements use one image per configuration and may vary with image content and resolution. Batched throughput, energy consumption, and cross-hardware performance are not evaluated.

CPU RSS and GPU memory describe different resources, and GPU-process memory includes allocator reservation and CUDA context overhead. Furthermore, the SAM and SAM2 benchmarks use \texttt{points\_per\_batch=64}, whereas the corresponding accuracy experiments use 128; their reported latency and memory measurements therefore characterize the benchmark configuration. Finally, although fixed-threshold replay is substantially faster than per-image threshold search, it does not preserve the same adaptive region-count selection procedure. All principal segmentation accuracy results use the full threshold-search configuration.

\section{Geometric Constraints and Region-Count Sensitivity}
\label{app:geometric_ablation}

We investigate the influence of the target region count $k$ and the gated neck-width constraint on segmentation performance. Table~\ref{tab:geometric_ablation} presents representative results on COCO Panoptic and Cityscapes. For COCO, we report the naturally reweighted results from the held-out train2017 subset used for selecting $k$. For Cityscapes, we compare configurations with and without the neck-width constraint on the validation set. All configurations use detrended depth maps.

\begin{table}[t]
\centering
\caption{Sensitivity to target region count $k$ and the gated neck-width constraint, measured using Hungarian-matched mIoU. COCO results are from the 500-image held-out train2017 subset with natural-distribution reweighting, using the adopted neck-width configuration. Cityscapes results are from the full 500-image validation set, with the neck-width constraint disabled ($m=1$) or enabled ($m=3$).}
\label{tab:geometric_ablation}
\small
\setlength{\tabcolsep}{4pt}
\begin{tabular}{cc|ccc}
\toprule
\multicolumn{2}{c|}{COCO (held-out)}
& \multicolumn{3}{c}{Cityscapes (val)} \\
\cmidrule(lr){1-2}
\cmidrule(lr){3-5}
$k$ & H-mIoU & $k$ & Veto off & Veto on \\
\midrule
4  & 0.2241 & 10 & 0.1543 & 0.1550 \\
5  & 0.2329 & 20 & 0.1952 & 0.1996 \\
6  & \textbf{0.2357} & 25 & 0.1987 & 0.2067 \\
7  & 0.2300 & 30 & 0.1945 & \textbf{0.2086} \\
8  & 0.2252 & 35 & 0.1844 & 0.2061 \\
10 & 0.2174 & 40 & 0.1728 & 0.2027 \\
\bottomrule
\end{tabular}
\end{table}

\textbf{Target Region Count.}
Segmentation performance depends on the specified target region count, with the optimal granularity varying across datasets. On COCO, Hungarian-matched mIoU peaks at $k=6$ on the held-out subset and decreases for both smaller and larger targets. On Cityscapes, the best configuration occurs at $k=30$, reflecting the greater region granularity required for its more complex urban scenes. These results illustrate the importance of balancing region separation against excessive fragmentation. The specified $k$ remains a target rather than a guaranteed final count, since threshold selection operates over discovered achievable region counts and subsequent cleanup may further merge small regions.

\textbf{Gated Neck-Width Constraint.}
The neck-width constraint prevents premature merging between sufficiently large components connected through limited spatial contact. Its contribution becomes more pronounced at higher target region counts. On Cityscapes, enabling the constraint increases Hungarian-matched mIoU from 0.1945 to 0.2086 at $k=30$, an absolute improvement of 0.0141. The advantage grows with $k$ over the evaluated range, allowing higher region counts to remain effective without excessive merging. The area gate is essential to this behavior, as applying the contact constraint indiscriminately can prevent small components from merging and produce severe fragmentation.

Interestingly, the sensitivity to the neck-width threshold differs across datasets. On COCO, varying the threshold produces graded changes in performance, whereas on Cityscapes, thresholds of $m=3$ and $m=16$ yield identical results across their jointly evaluated configurations. ADE20K exhibits both behaviors, with graded sensitivity at smaller target counts and saturation at $k\geq10$. Despite these differences, the benefit of enabling the constraint generally increases with the target region count on Cityscapes and ADE20K, where systematic comparisons across multiple target counts are available. This suggests that the constraint is particularly useful when preserving a larger number of geometrically distinct regions.

\section{Additional Qualitative Results}
\label{app:additional_qualitative}

\begin{figure*}[t]
\centering
\setlength{\tabcolsep}{1pt}
\begin{tabular}{@{}ccccc@{}}
{\scriptsize RGB} & {\scriptsize Raw depth} & {\scriptsize Detrended depth} & {\scriptsize Ours (overlay)} & {\scriptsize GT (overlay)} \\
\includegraphics[width=0.19\textwidth]{figures/supplementary/qualitative/coco/000000145781/rgb.png} &
\includegraphics[width=0.19\textwidth]{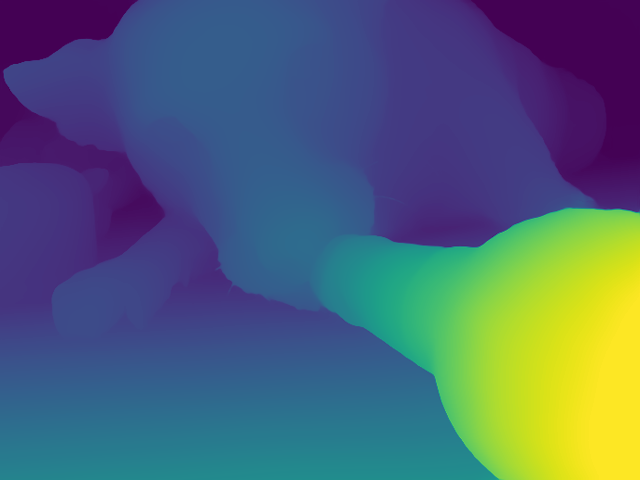} &
\includegraphics[width=0.19\textwidth]{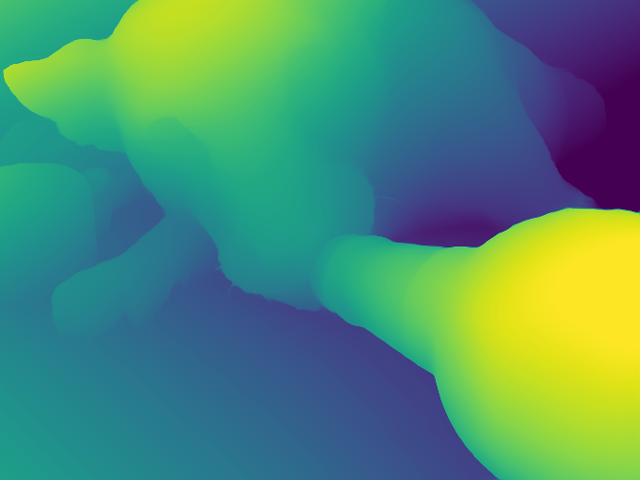} &
\includegraphics[width=0.19\textwidth]{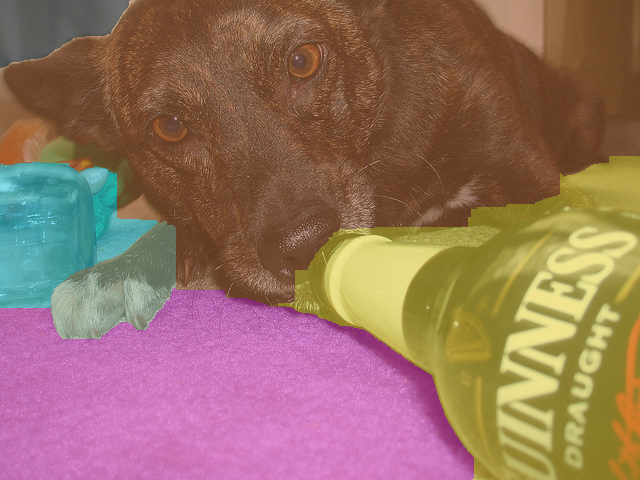} &
\includegraphics[width=0.19\textwidth]{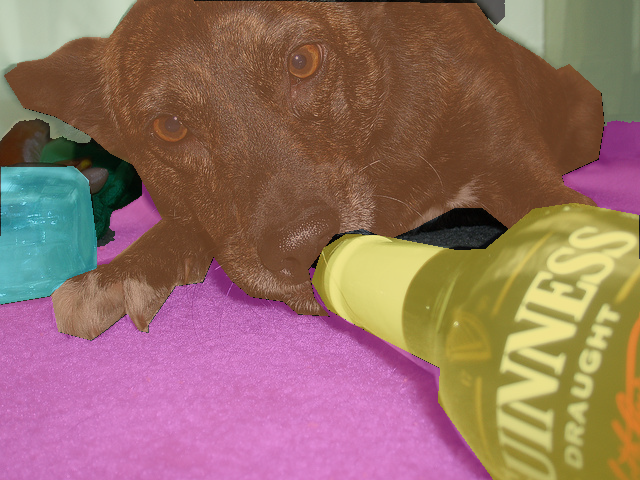} \\
\includegraphics[width=0.19\textwidth]{figures/supplementary/qualitative/coco/000000070048/rgb.png} &
\includegraphics[width=0.19\textwidth]{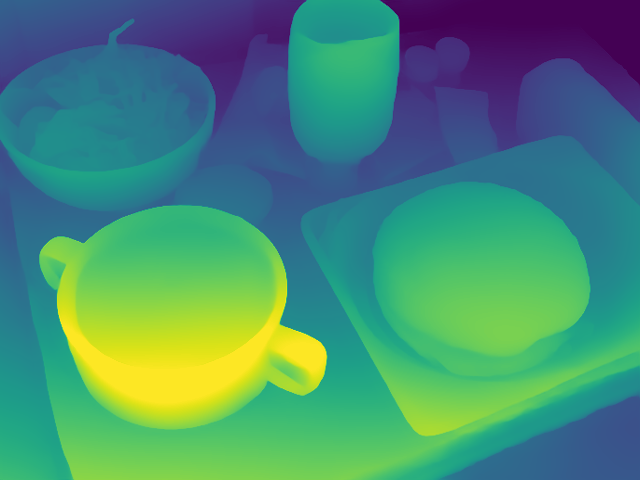} &
\includegraphics[width=0.19\textwidth]{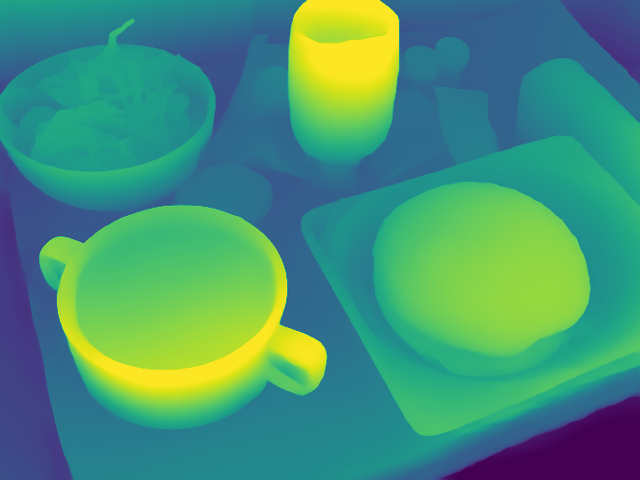} &
\includegraphics[width=0.19\textwidth]{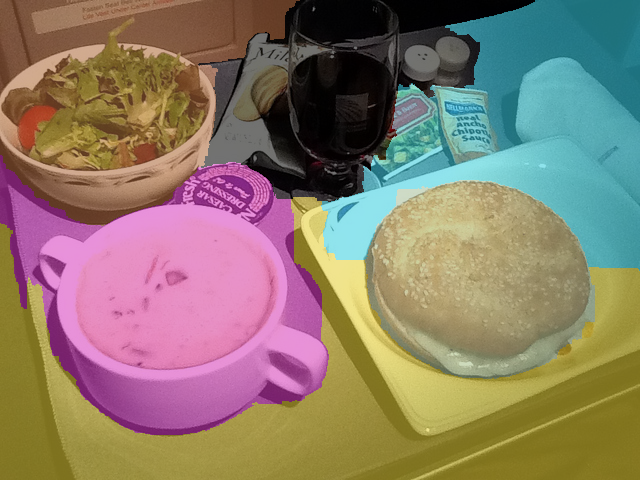} &
\includegraphics[width=0.19\textwidth]{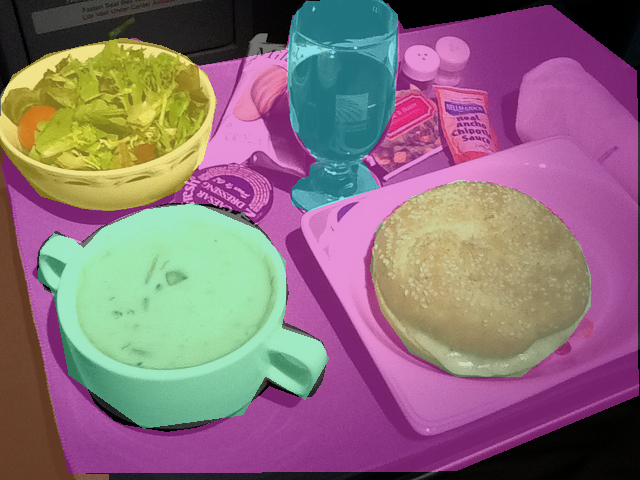} \\
\includegraphics[width=0.19\textwidth]{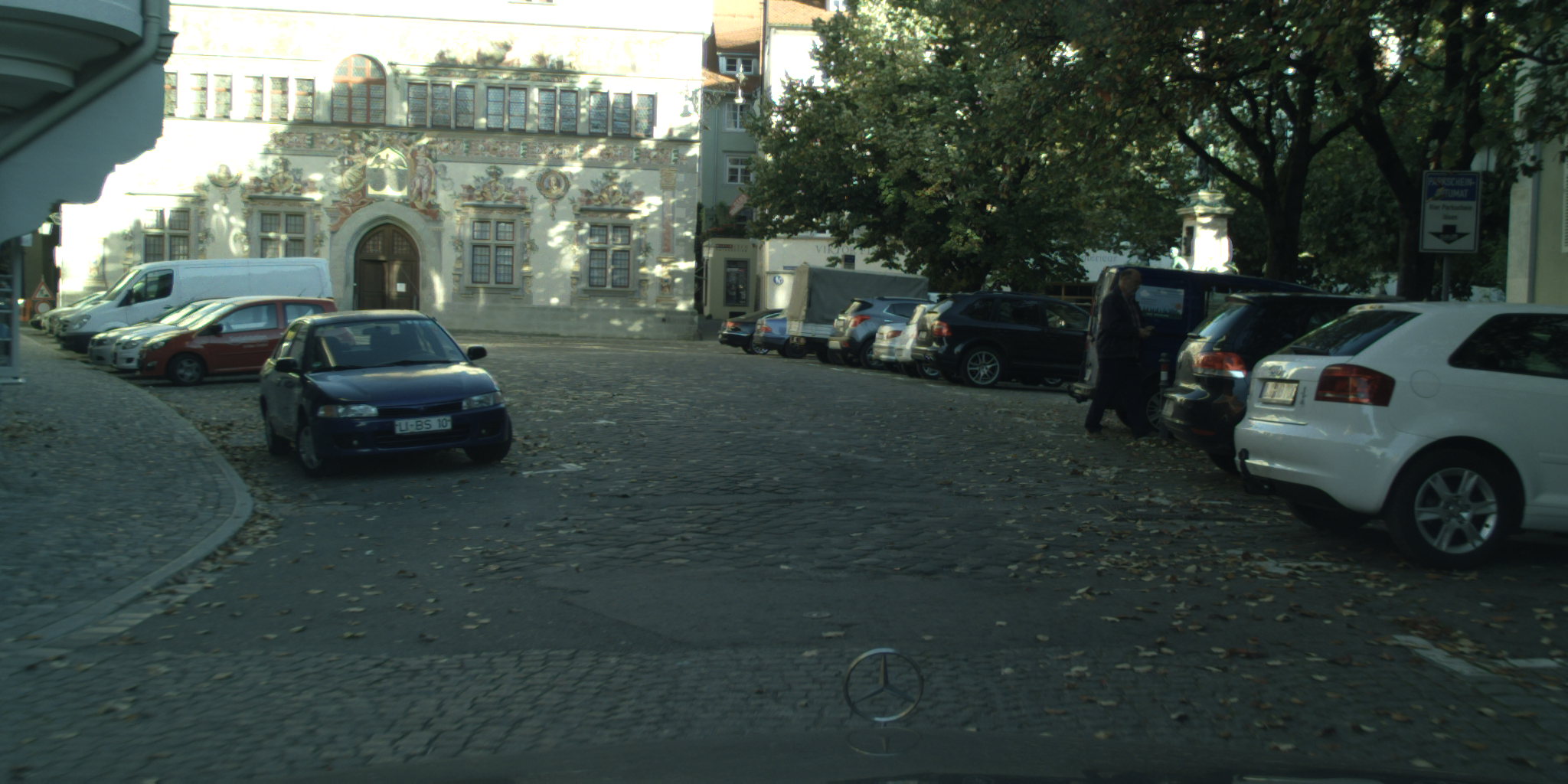} &
\includegraphics[width=0.19\textwidth]{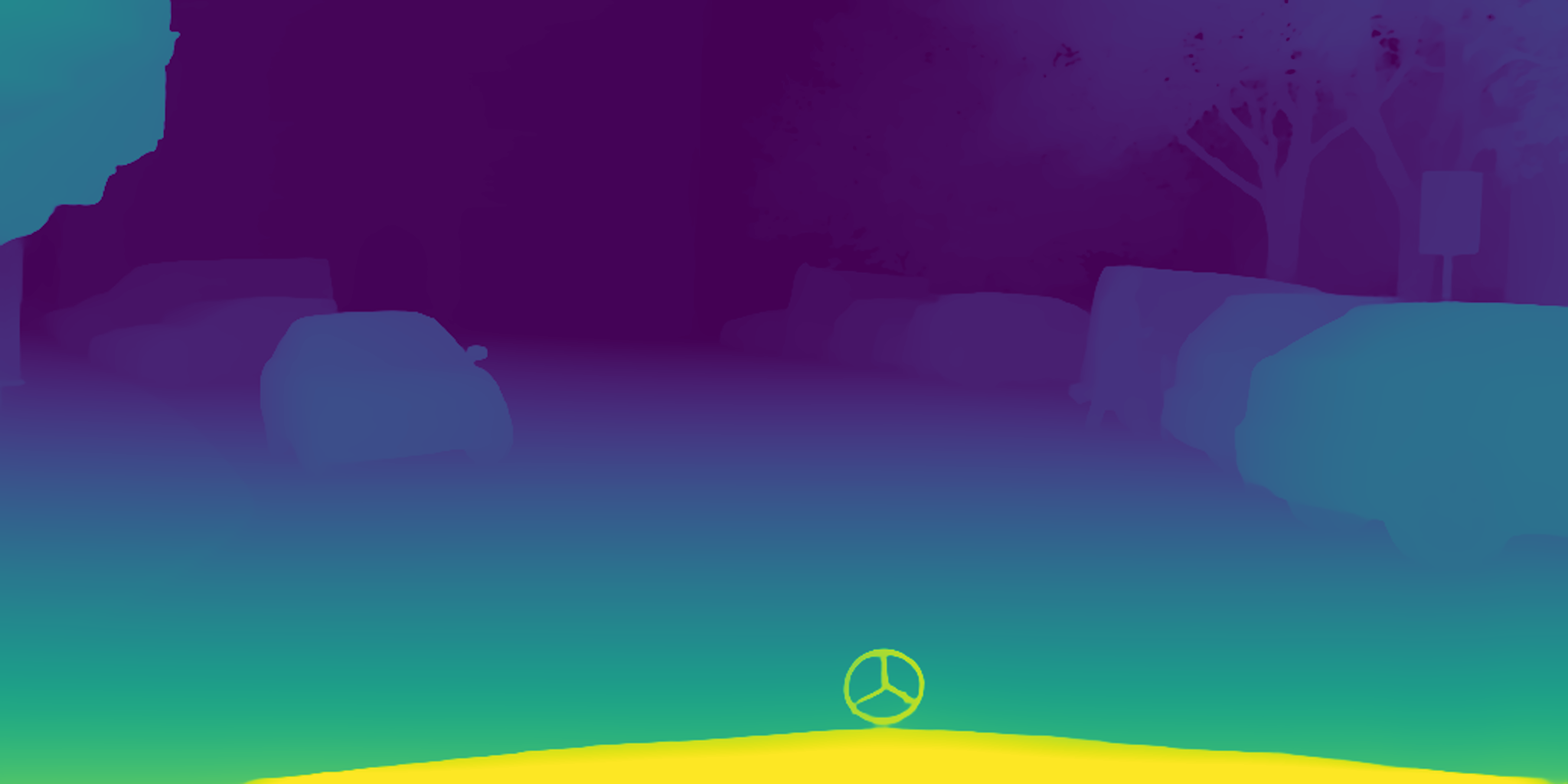} &
\includegraphics[width=0.19\textwidth]{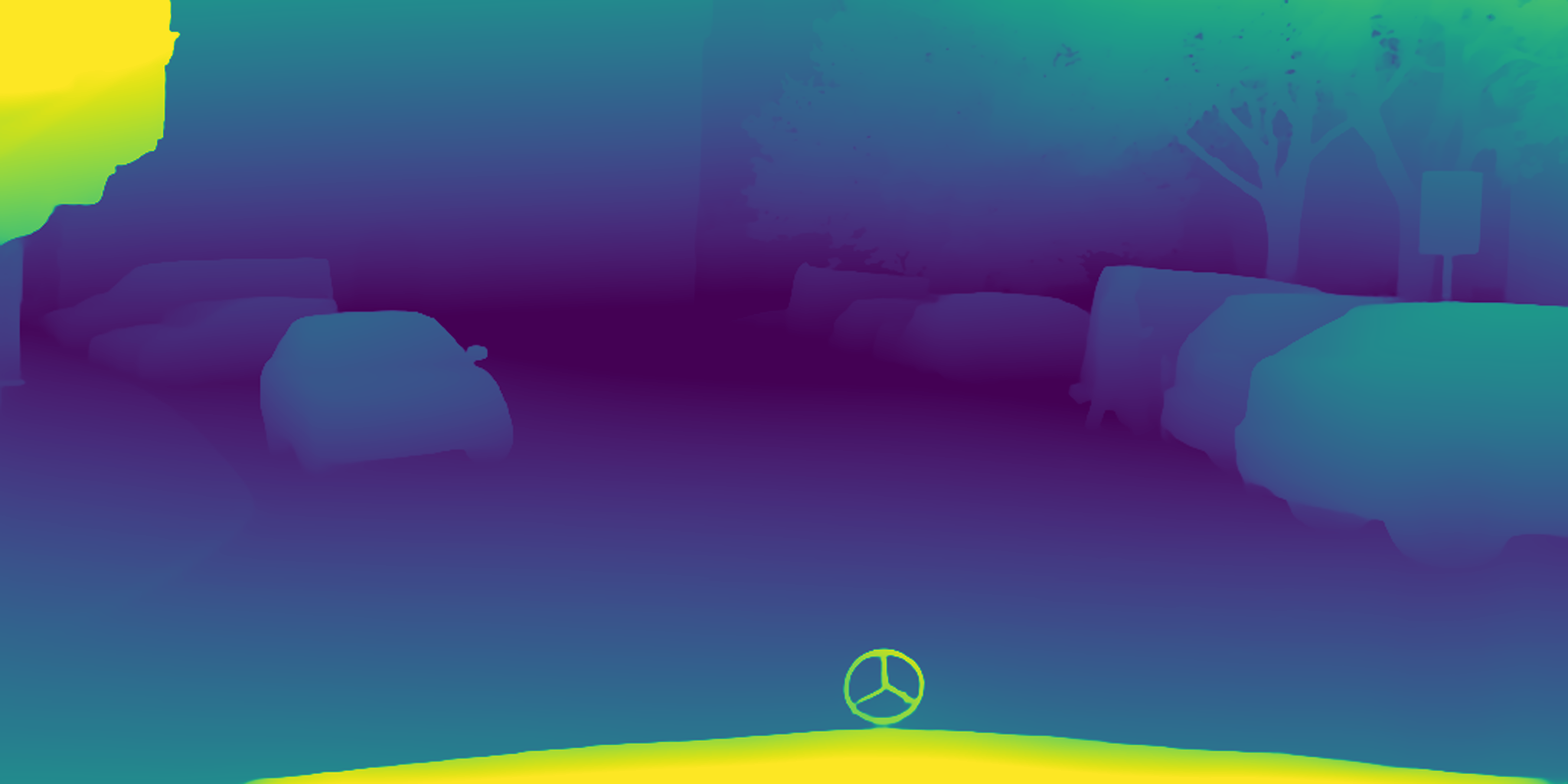} &
\includegraphics[width=0.19\textwidth]{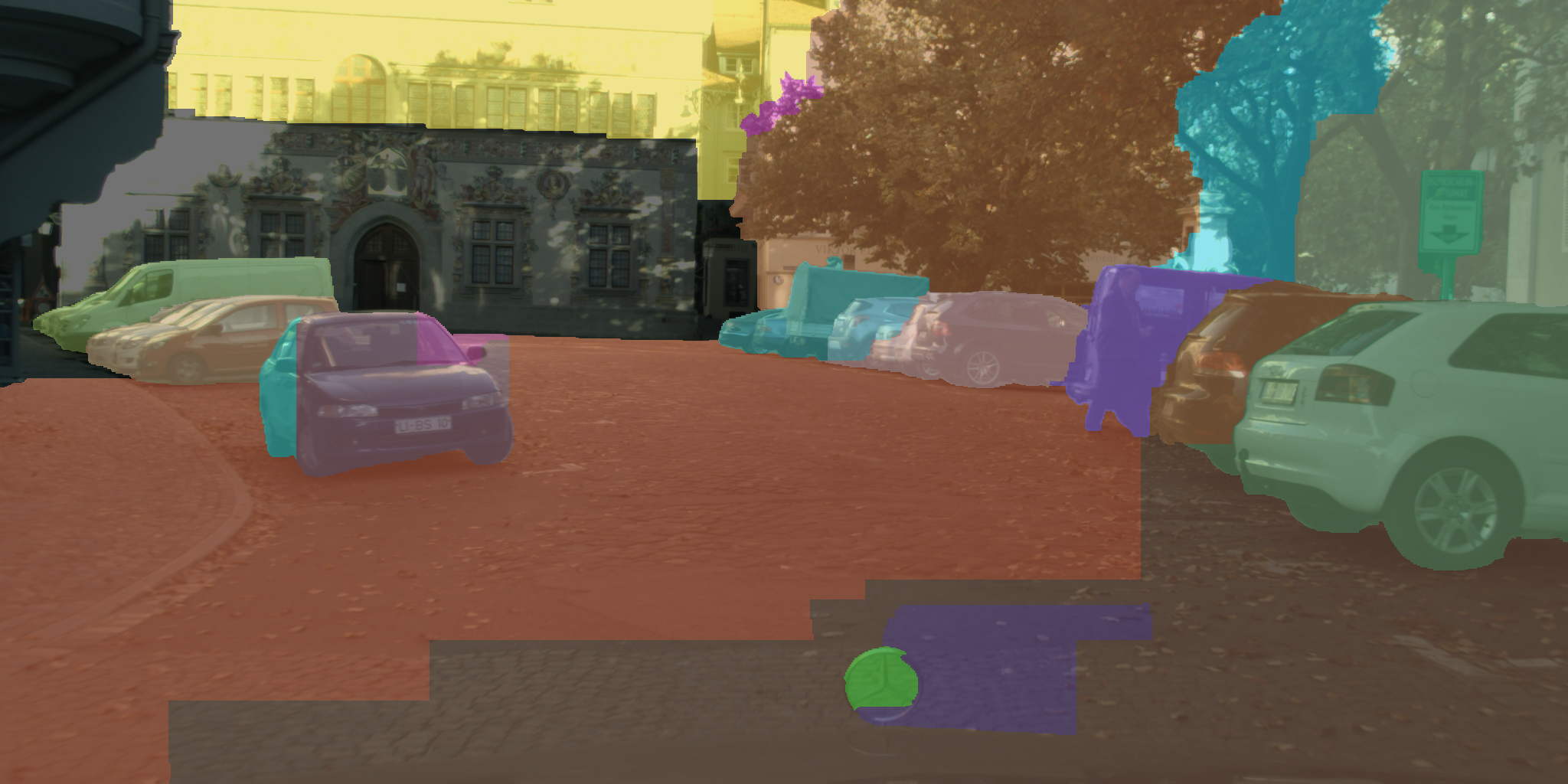} &
\includegraphics[width=0.19\textwidth]{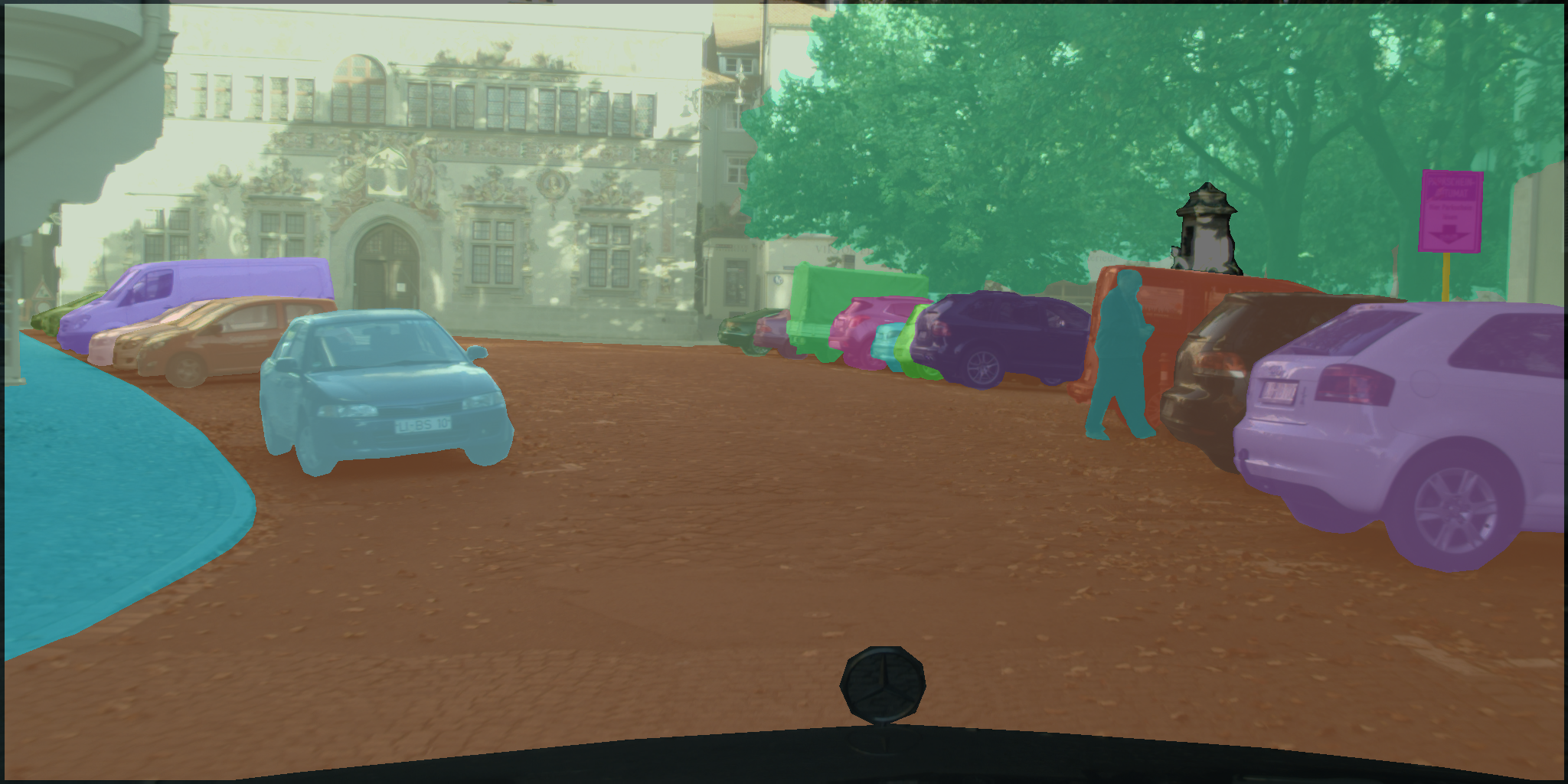} \\
\includegraphics[width=0.19\textwidth]{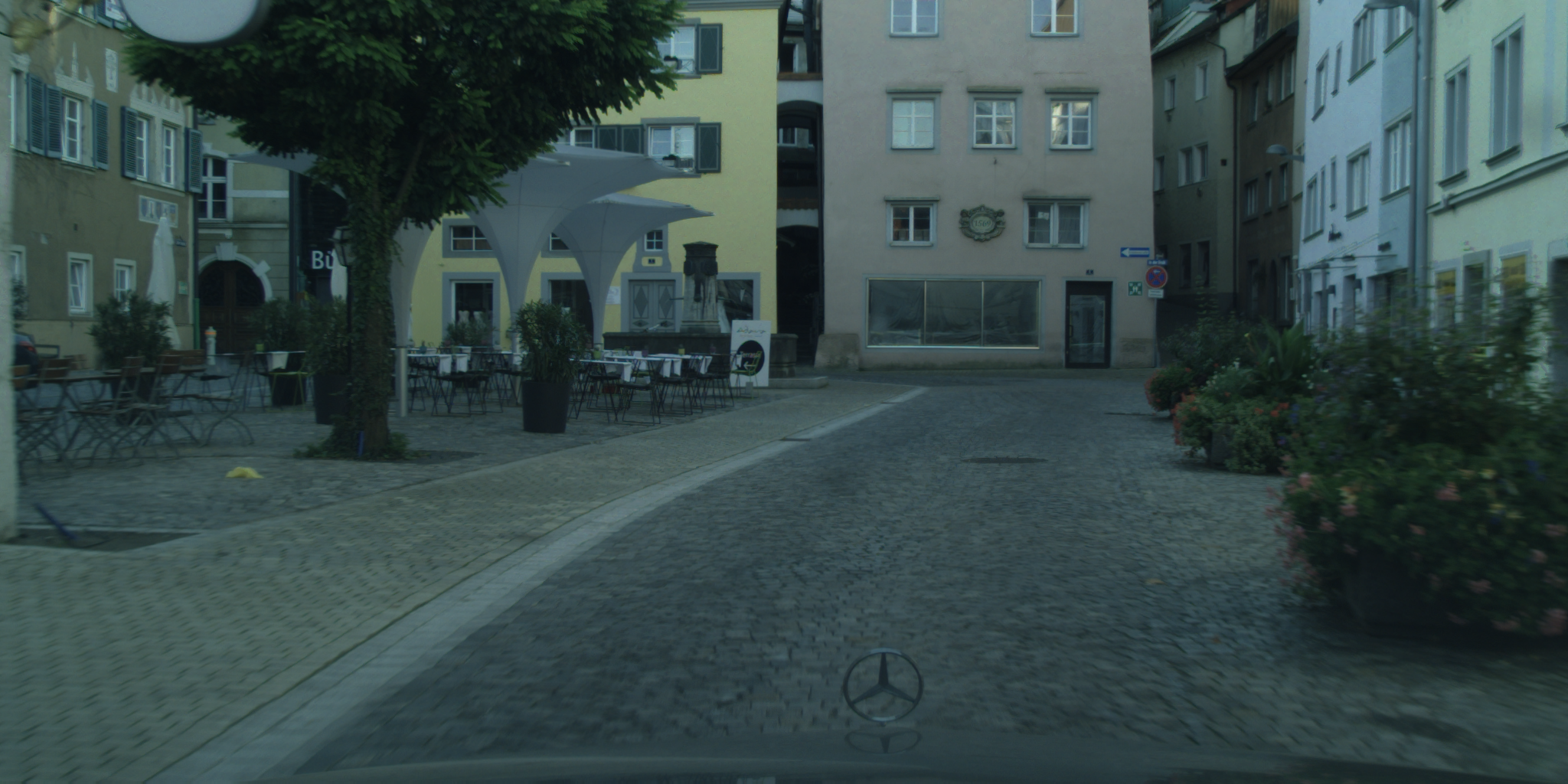} &
\includegraphics[width=0.19\textwidth]{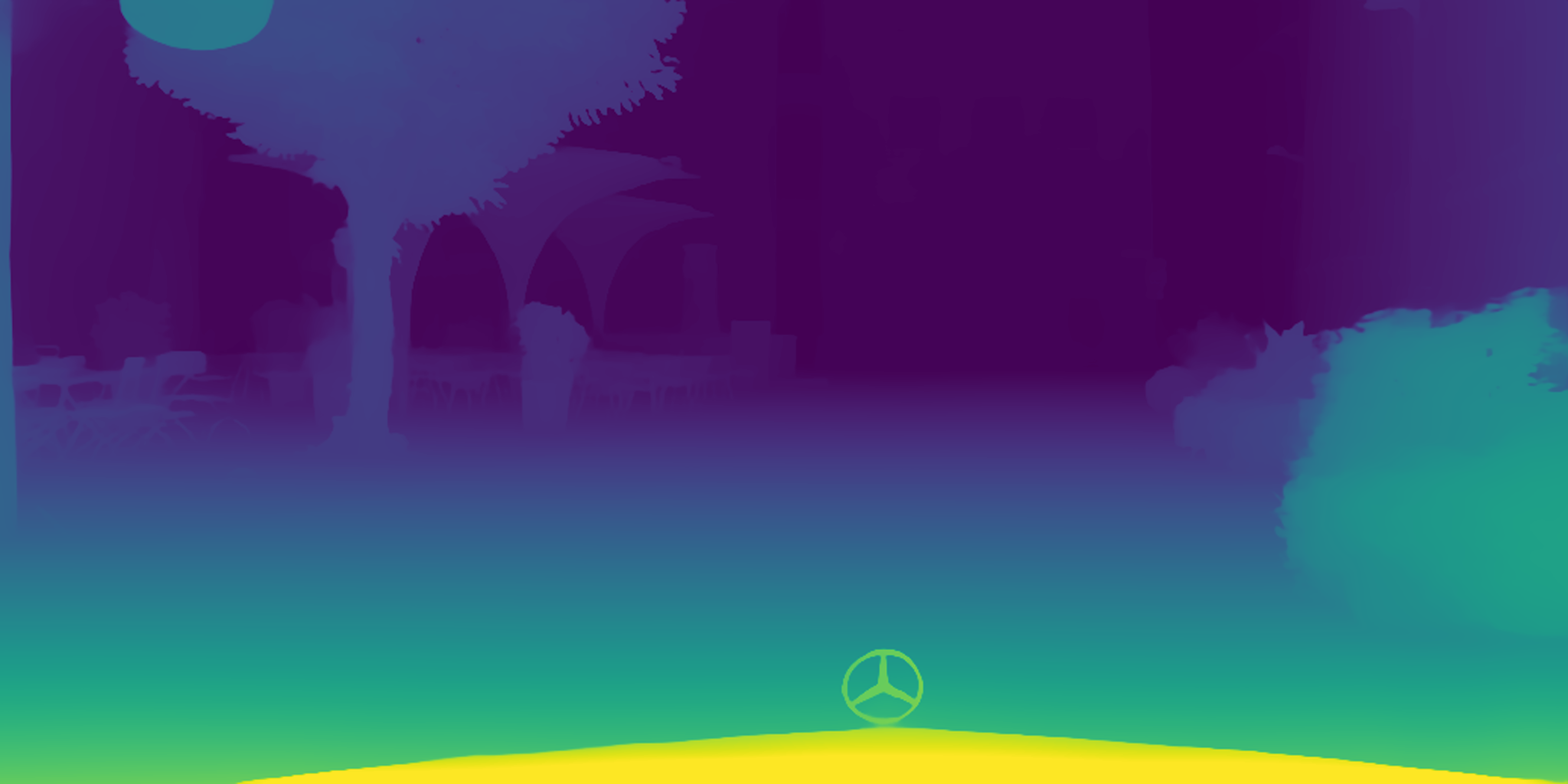} &
\includegraphics[width=0.19\textwidth]{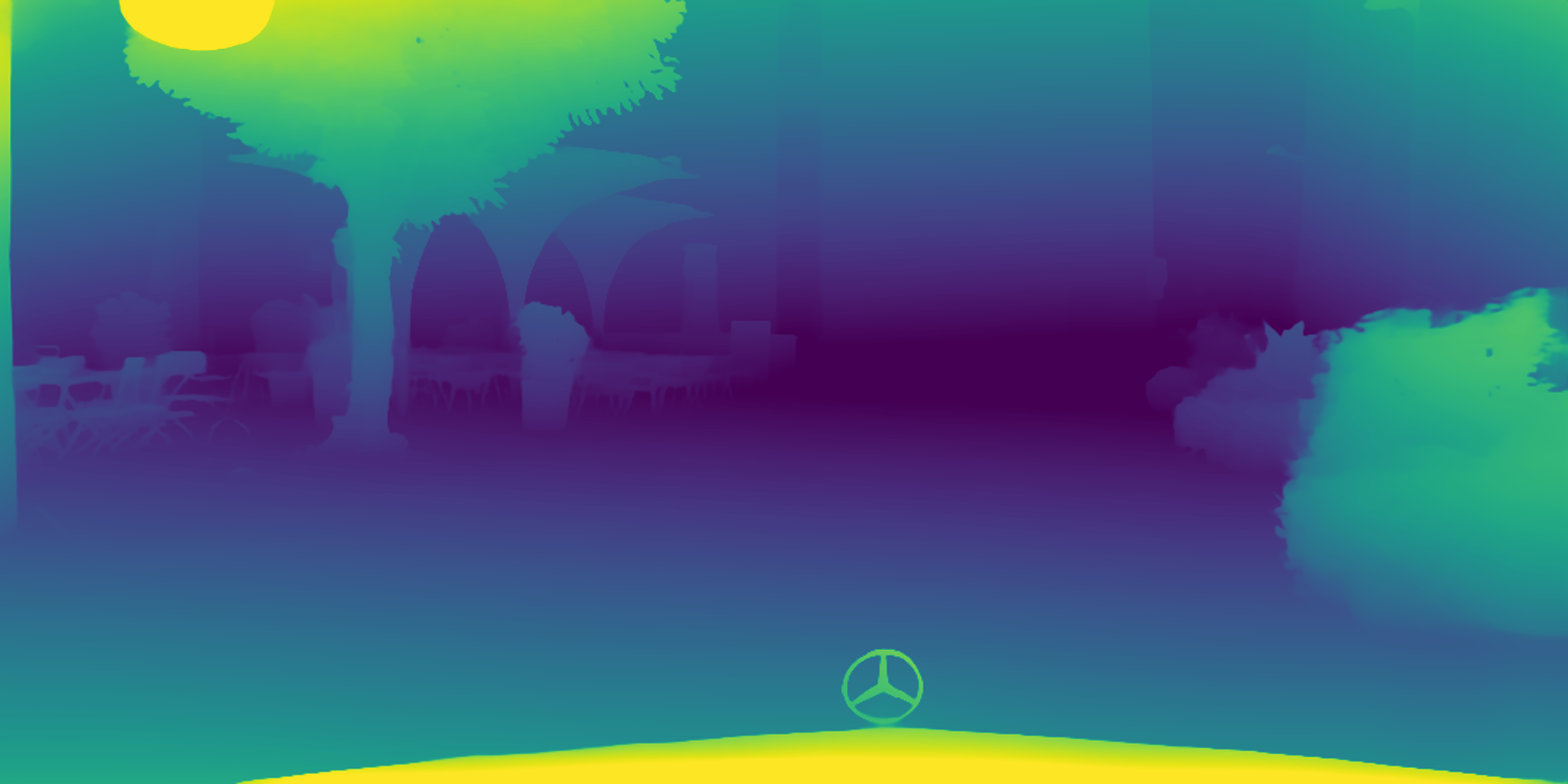} &
\includegraphics[width=0.19\textwidth]{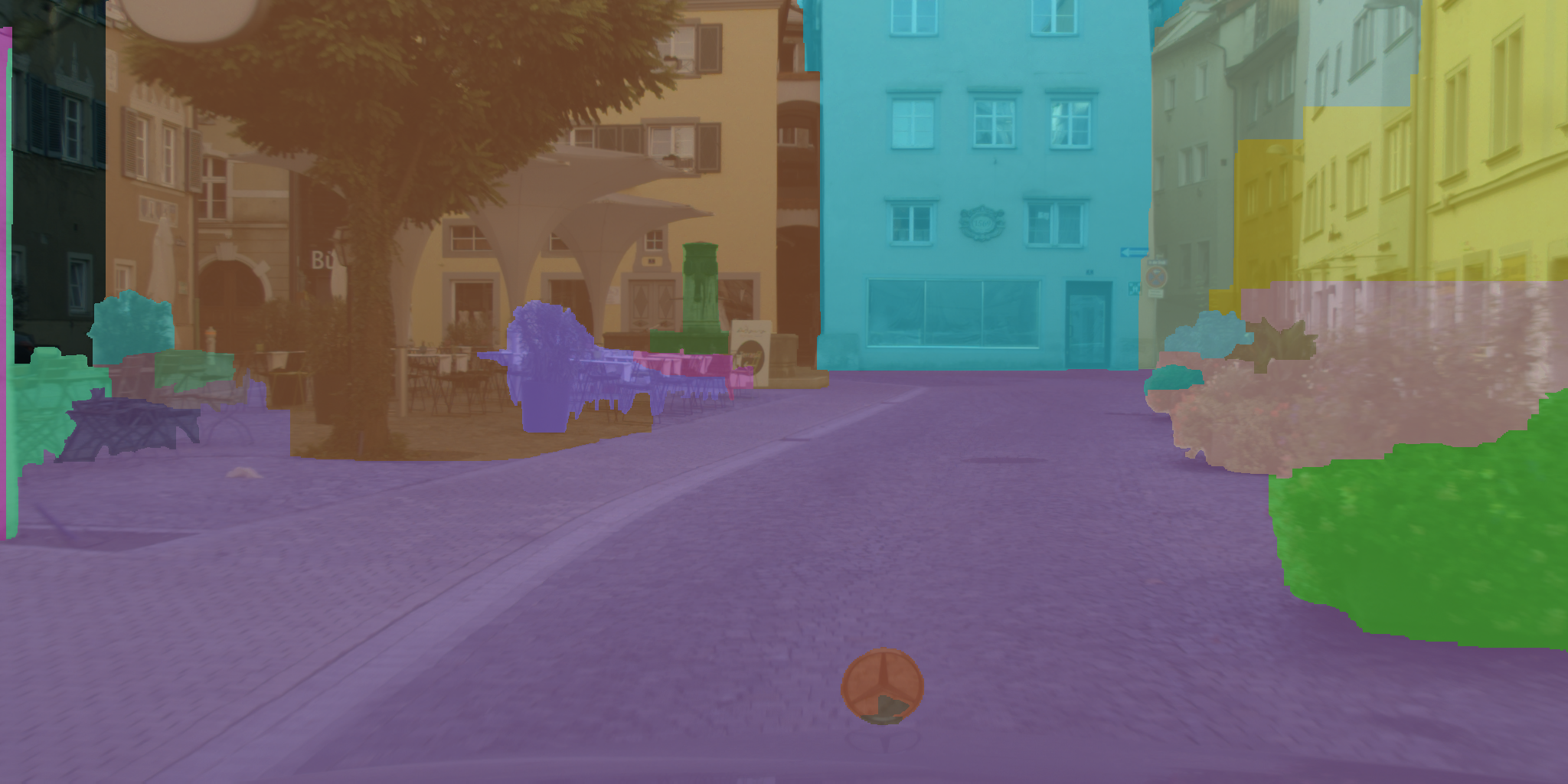} &
\includegraphics[width=0.19\textwidth]{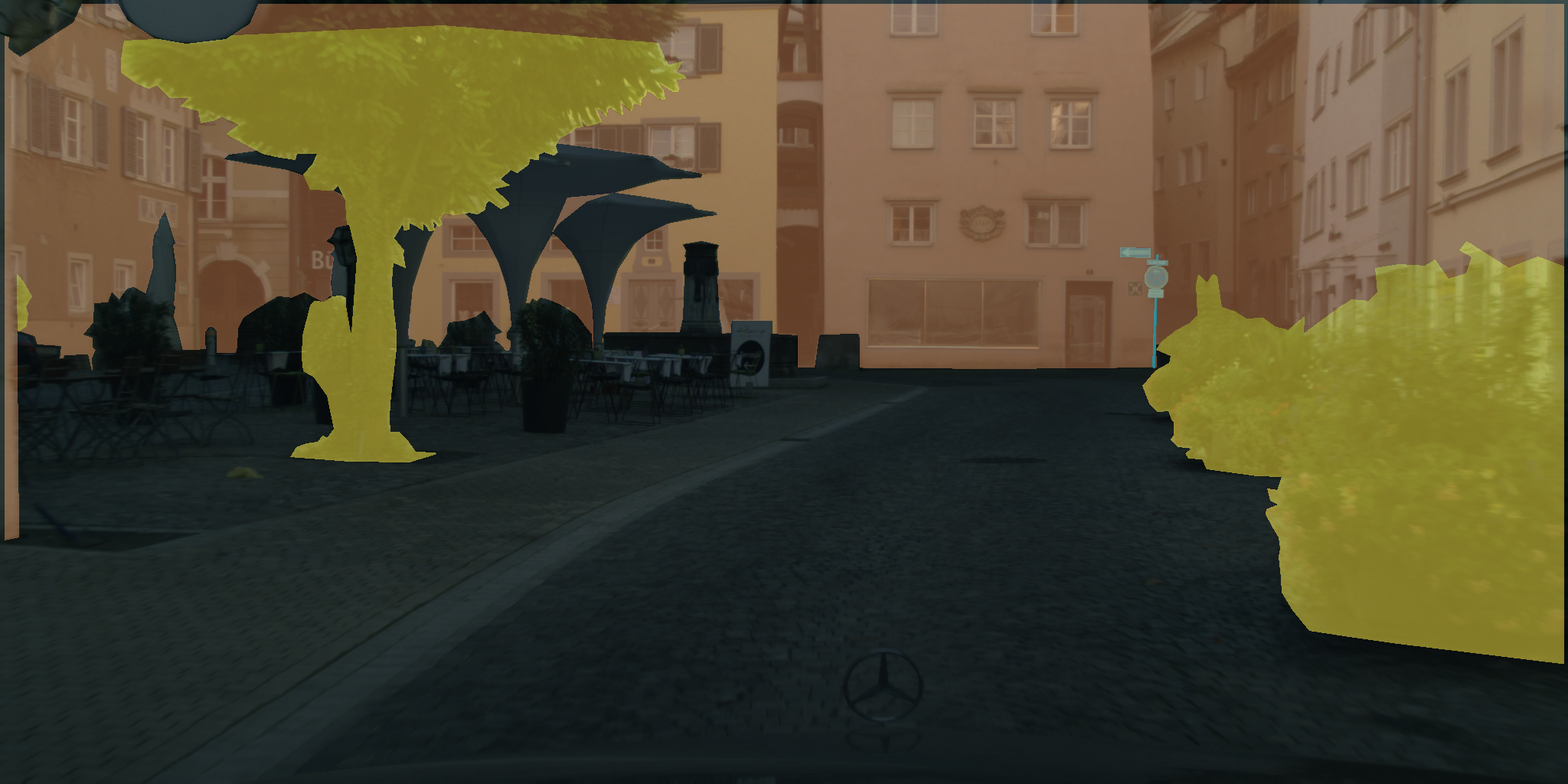} \\
\includegraphics[width=0.19\textwidth]{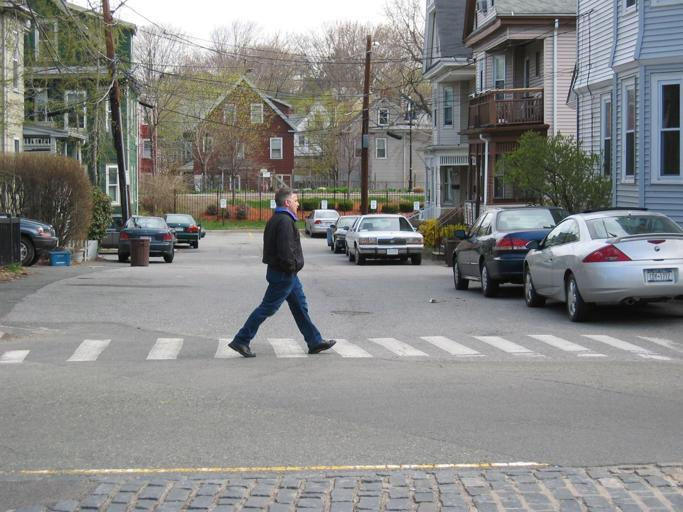} &
\includegraphics[width=0.19\textwidth]{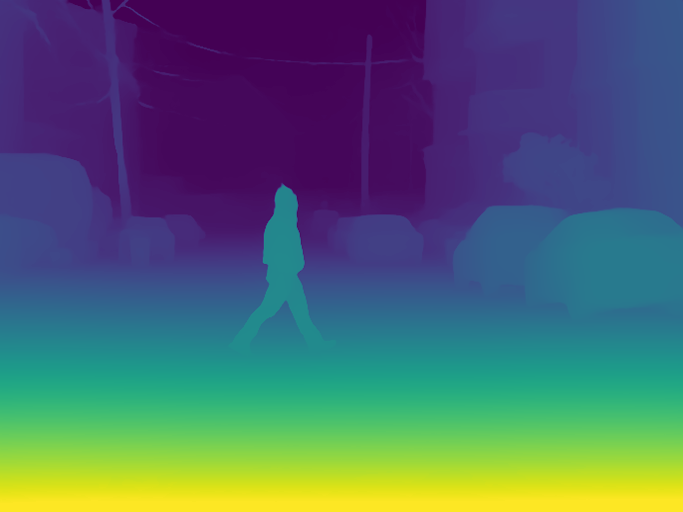} &
\includegraphics[width=0.19\textwidth]{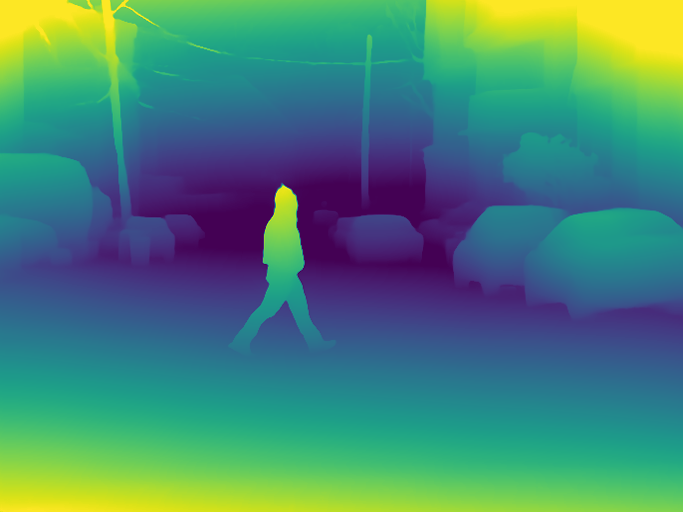} &
\includegraphics[width=0.19\textwidth]{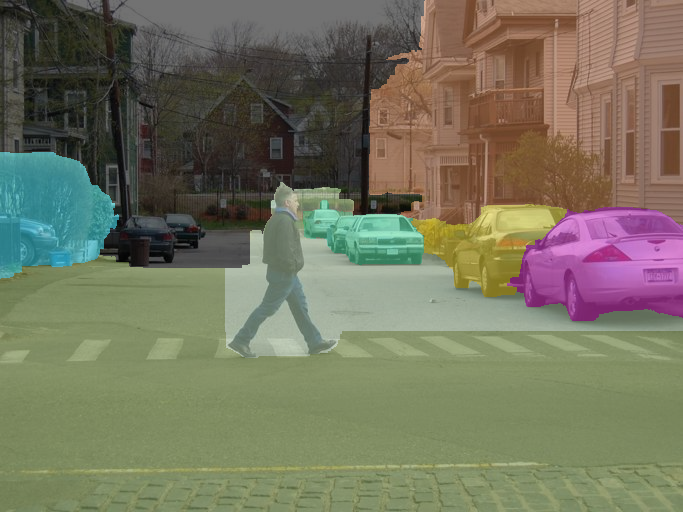} &
\includegraphics[width=0.19\textwidth]{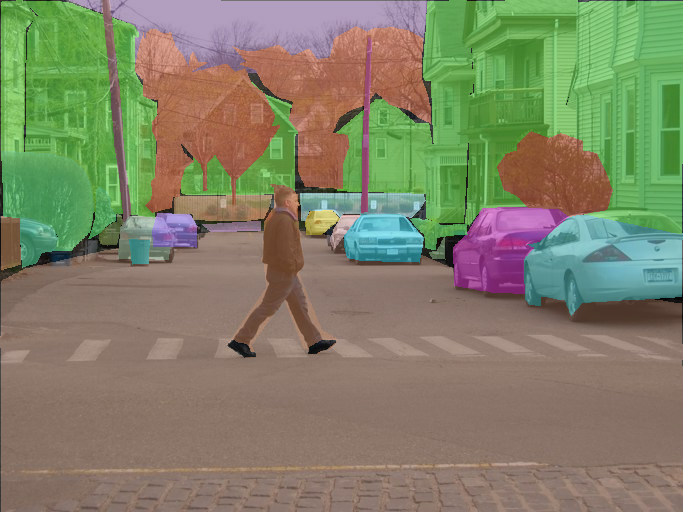} \\
\end{tabular}
\caption{Additional qualitative results, arranged by row: (1) COCO \texttt{000000145781}, (2) COCO \texttt{000000070048}, (3) Cityscapes \texttt{lindau\_000038\_000019}, (4) Cityscapes \texttt{lindau\_000018\_000019}, and (5) ADE20K \texttt{ADE\_val\_00000783}. Predictions and ground truth are shown as colored overlays on RGB images. These examples include successful foreground separation, partial merging, and a case where the predicted region granularity exceeds that of the ground-truth annotation.}
\label{fig:supp_qualitative}
\end{figure*}

We provide additional examples illustrating the range of segmentation behavior across COCO Panoptic, Cityscapes, and ADE20K. Figure~\ref{fig:supp_qualitative} presents five examples using the same depth estimators and evaluation settings as the main experiments. The Cityscapes gallery images were rendered with $m=16$, which produced identical results to the adopted $m=3$ in their jointly evaluated configurations. Each example includes the original image, estimated depth, detrended depth, the final predicted segmentation overlaid on RGB, and the ground-truth partition overlaid on RGB. The two overlays visualize regions for comparison; the ground truth is not supplied to our segmentation algorithm.

On COCO, \texttt{000000145781} and \texttt{000000070048} illustrate meaningful separation of foreground regions, although not all object boundaries are recovered. Their H-mIoU scores are 0.6236 and 0.6089, respectively. Cityscapes \texttt{lindau\_000038\_000019} provides a comparatively strong urban-scene result (H-mIoU 0.3991), with 22 predicted regions and 23 ground-truth segments. In contrast, \texttt{lindau\_000018\_000019} yields 23 predicted regions against four ground-truth segments (H-mIoU 0.0251). This case illustrates sensitivity to differences in annotation granularity rather than establishing that all predicted boundaries are geometrically incorrect. On ADE20K, \texttt{ADE\_val\_00000783} exhibits a mixed outcome (H-mIoU 0.2067): a prominent foreground person is separated, while some neighboring vehicles remain merged.

The selected images are illustrative rather than a random sample of the benchmark splits. In particular, the examples indicate that geometric structure can produce meaningful partitions even without semantic predictions, but that the specified target region count and the input depth geometry may prevent separation of objects that appear distinct in the ground truth.

\section{Additional Merge-Tree Visualizations}
\label{app:additional_merge_tree_visualizations}

We further illustrate the depth-ordered sweep and the sequence of state-dependent geometric merge decisions. These visualizations are obtained from instrumented runs of the adopted segmentation algorithm, rather than from a threshold-independent hierarchy constructed separately. In particular, rejected component pairs remain eligible for reconsideration at subsequent bridging pixels, and the resulting accepted-merge structure depends on the operating threshold.

\subsection{Component Evolution During the Depth Sweep}
\label{app:supp_sweep}

\begin{figure*}[t]
\centering
\setlength{\tabcolsep}{3pt}
\begin{tabular}{@{}ccc@{}}
{\scriptsize 10\% (35 components)} & {\scriptsize 25\% (34 components)} & {\scriptsize 50\% (28 components)} \\
\includegraphics[width=0.29\textwidth]{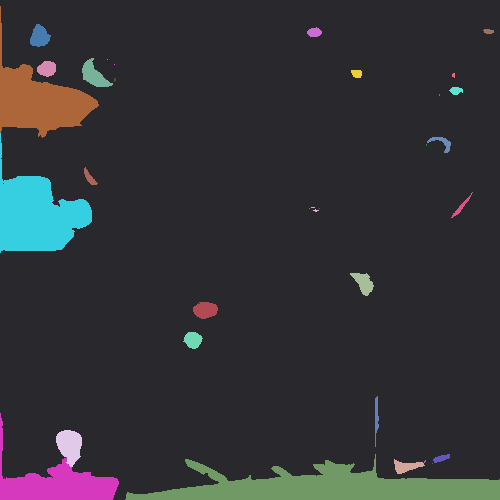} &
\includegraphics[width=0.29\textwidth]{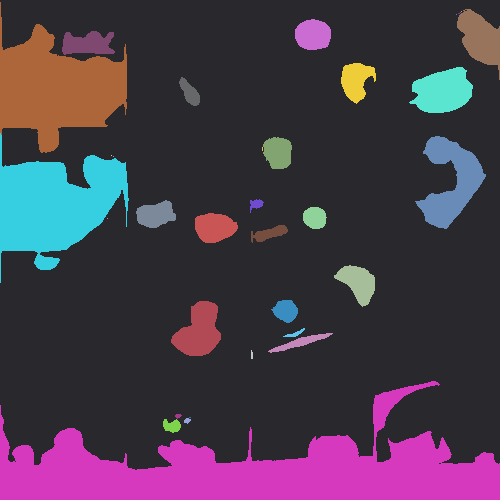} &
\includegraphics[width=0.29\textwidth]{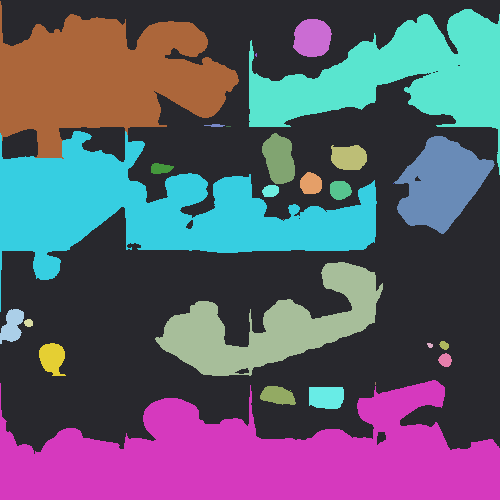} \\[3pt]
{\scriptsize 75\% (9 components)} & {\scriptsize 90\% (7 components)} & {\scriptsize 100\% (6 components)} \\
\includegraphics[width=0.29\textwidth]{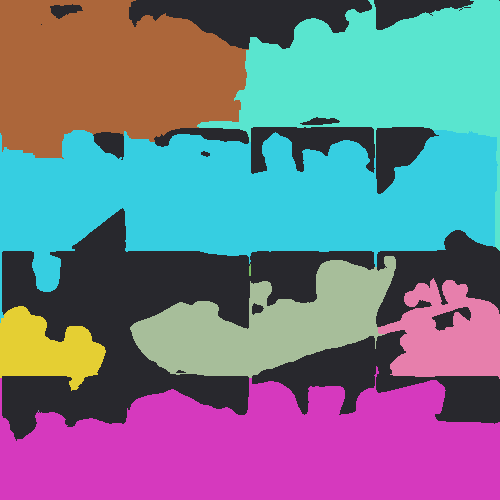} &
\includegraphics[width=0.29\textwidth]{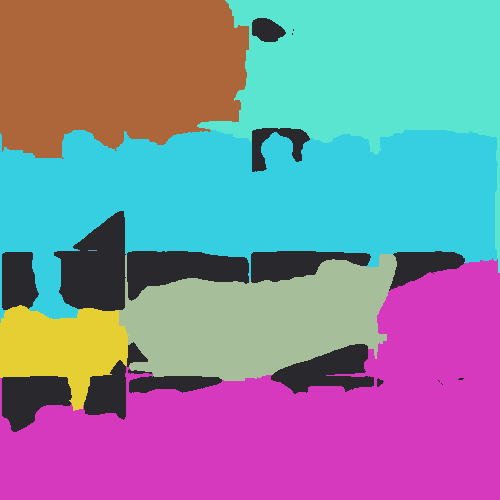} &
\includegraphics[width=0.29\textwidth]{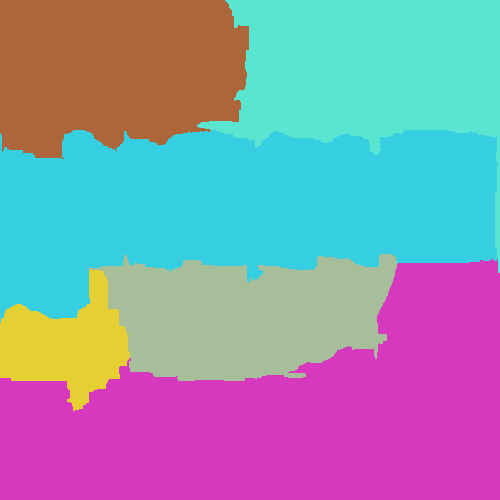} \\
\end{tabular}
\caption{Evolution of connected components during one depth-ordered sweep on COCO image \texttt{000000435081}. Each panel shows the active components at the indicated fraction of pixels processed in descending detrended-depth order; dark areas are not yet activated. Colors distinguish components within a snapshot. The last panel shows the partition before cleanup; for this particular image, cleanup does not change the partition, leaving six regions.}
\label{fig:supp_merge_sweep}
\end{figure*}

Figure~\ref{fig:supp_merge_sweep} shows six snapshots from the same fixed-threshold sweep of COCO image \texttt{000000435081}. The fraction of activated pixels increases from 10\% to 100\%. The number of live components in the shown snapshots is 35, 34, 28, 9, 7, and 6, respectively. New components may be born as the sweep reaches previously inactive depth maxima, while permitted merges consolidate components encountered at bridging pixels. The displayed component count describes the current sweep state, not the number of regions returned by a separate threshold search.

\subsection{Accepted Merges and Rejected Evaluations}
\label{app:supp_merge_forest}

\begin{figure*}[t]
\centering
{\scriptsize (a) Approved merges and surviving components}\par\vspace{2pt}
\includegraphics[width=0.84\textwidth]{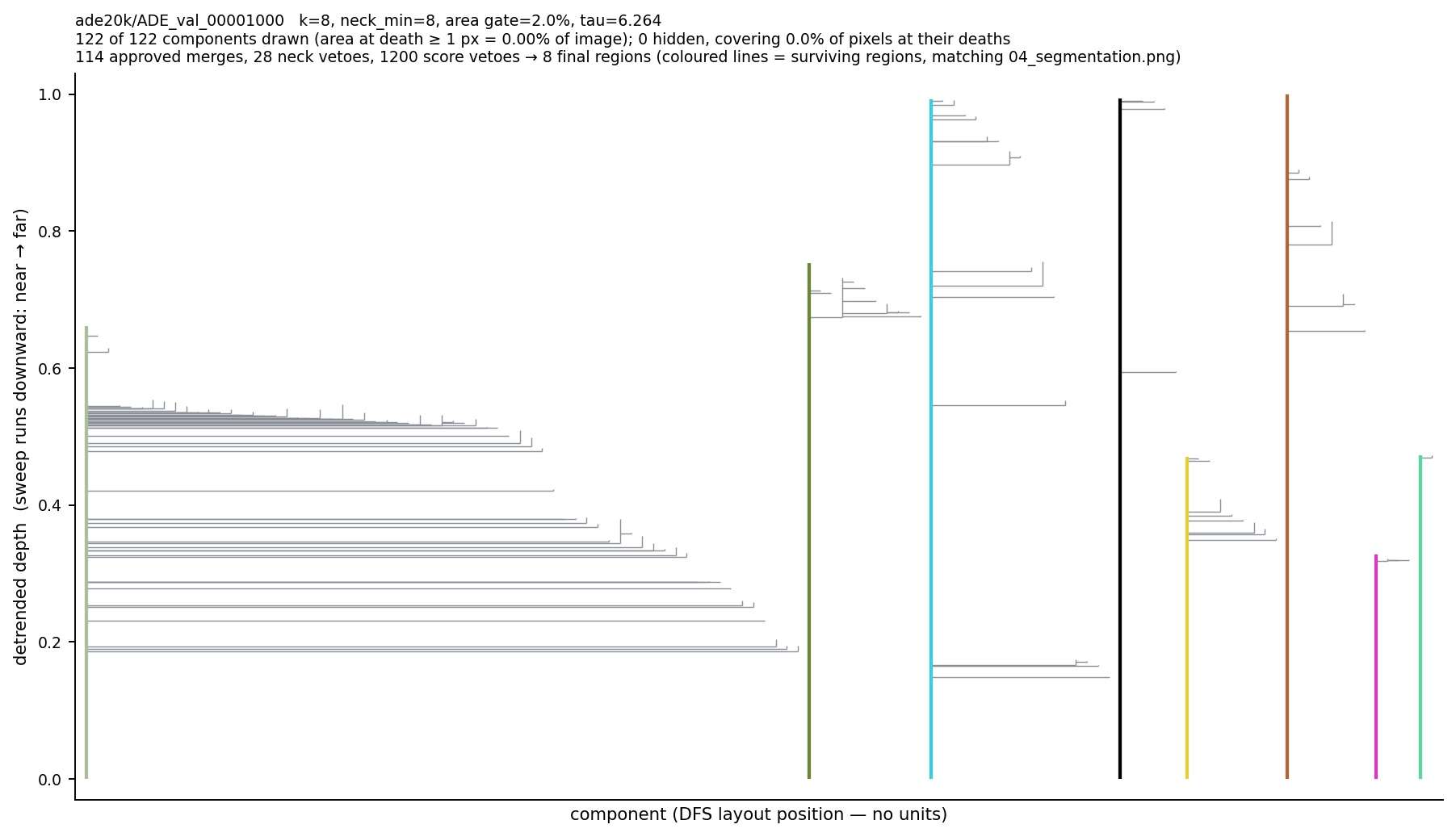}\par\vspace{5pt}
{\scriptsize (b) Approved merges with contact- and score-veto evaluations}\par\vspace{2pt}
\includegraphics[width=0.84\textwidth]{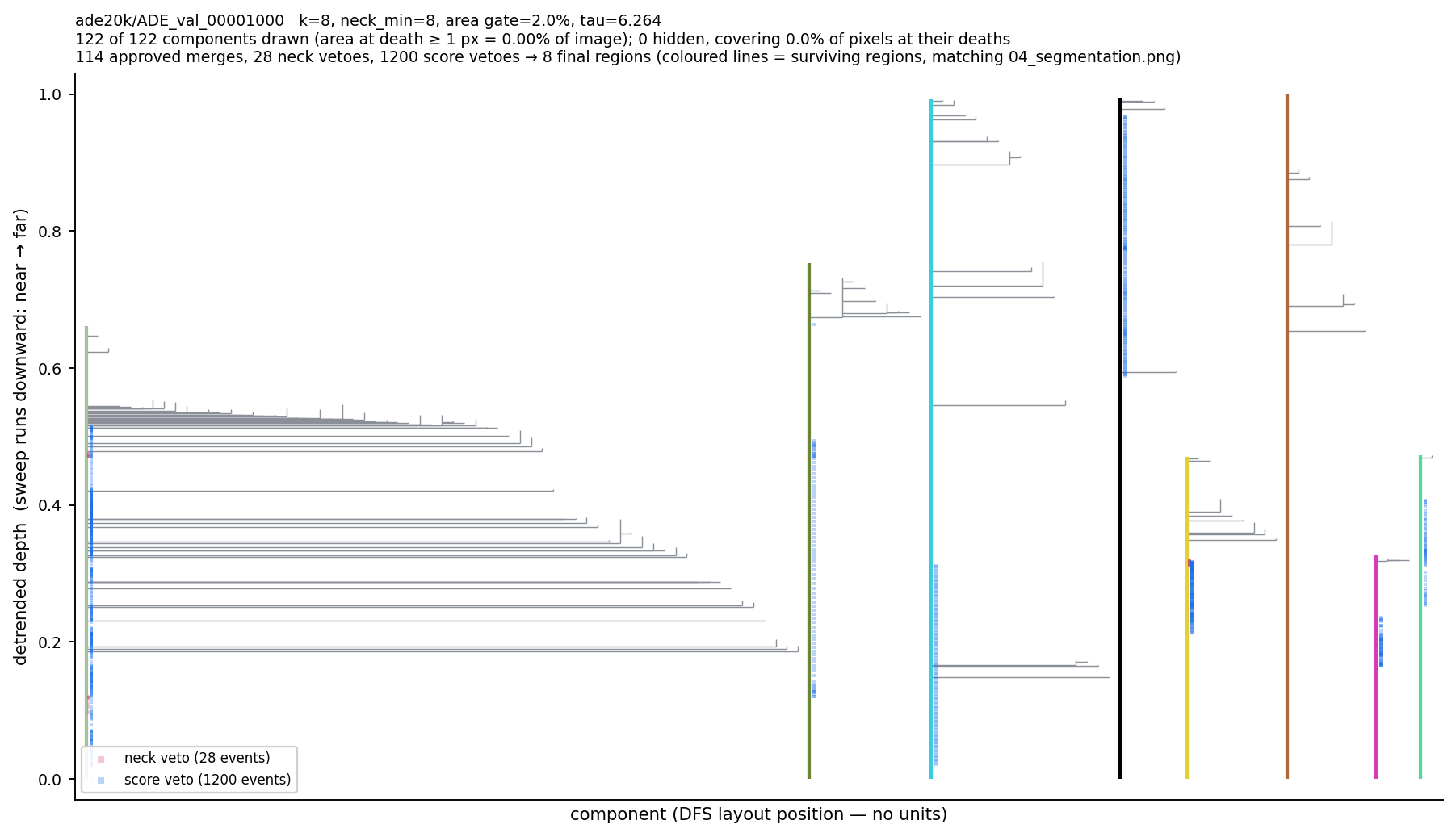}
\caption{Threshold-dependent accepted-merge forest for ADE20K image \texttt{ADE\_val\_00001000}. (a) Component lifetimes and approved merges during the sweep. (b) The same trace with markers for rejected evaluations under the gated spatial-contact constraint and area--persistence score. Rejected evaluations do not permanently prohibit a merge; the corresponding components can be reconsidered upon later contact. Because this is a threshold-dependent forest with multiple surviving roots, the plot should not be interpreted as a conventional threshold-independent merge tree.}
\label{fig:supp_merge_forest}
\end{figure*}

Figure~\ref{fig:supp_merge_forest} shows two views of the instrumented merge forest for ADE20K image \texttt{ADE\_val\_00001000}. The first displays component lifetimes and approved merges, whereas the second overlays the rejected merge evaluations produced by the contact and area--persistence criteria. The trace contains 122 components, 114 approved merges, 28 contact-veto evaluations, and 1,200 score-veto evaluations, leaving eight surviving regions. These veto counts refer to \emph{evaluations}, not distinct component pairs: a pair may be rejected at one bridging pixel and reconsidered at later encounters with updated area, contact, and persistence values.

The two views emphasize complementary aspects of the algorithm. Approved merges describe the evolving partition at the selected threshold, while the veto markers reveal that many potential merges are reconsidered as the sweep progresses. Their presence also explains why recording only component births and accepted merges is insufficient to reconstruct the full sequence of merge decisions. The diagrams do not separately visualize the result of toggling the contact constraint, and no such visual comparison is claimed here.

\end{document}